\documentclass{article}

 \usepackage[main, final]{neurips_2026}

\usepackage[utf8]{inputenc}
\usepackage[T1]{fontenc}
\usepackage{microtype}
\usepackage{graphicx}
\usepackage{subcaption}
\usepackage{wrapfig}
\usepackage{booktabs}
\usepackage[inline]{enumitem}
\usepackage{adjustbox}
\usepackage{float}
\usepackage{hyperref}
\usepackage{etoc} 
\usepackage{url}
\usepackage{xcolor}
\usepackage{amsmath}
\usepackage{amsfonts}
\usepackage{amssymb}
\usepackage{mathtools}
\usepackage{amsthm}
\usepackage{xspace}
\usepackage{ulem}
\usepackage{soul}

\usepackage{algorithm}
\usepackage[noEnd=true,indLines=true,commentColor=black,beginLComment=//~,endLComment=]{algpseudocodex}

\usepackage[square,numbers]{natbib}
\usepackage[capitalize,noabbrev]{cleveref}

\newcommand{\Description}[1]{}

\setstcolor{red}

\usepackage{bm,mathtools}

\def\dt{\mathrm{t}}
\def\dx{\mathrm{x}}
\def\dc{\mathrm{c}}

\def\Nx{N_{\dx}}
\def\Nt{N_{\dt}}
\def\Nc{N_{\dc}}

\def\la{\lambda}
\def\vla{\vlambda}
\def\vlaD{\sbs{\vla}{D}}
\def\vlaIC{\sbs{\vla}{IC}}

\def\NN{S_{\theta}}
\def\xtla{\mX^{\la}_{[t\sminus r]}}
\newcommand{\xtlapre}[1]{\mX^{\la}_{t\sminus #1}}

\def\eps{\varepsilon}
\def\teps{\tilde{\varepsilon}}

\def\1{\bm{1}}

\def\eps{{\varepsilon}}

\newcommand\sbs[2]{{#1}_{\text{#2}}}

\newcommand{\ldotscomp}{%
  \mathinner{{\ldotp}{\ldotp}{\ldotp}}%
}

\def\eq{\mathop{=}}

\newcommand\myin{\mathop{\in}}
\newcommand\mysubset{\mathop{\subset}}

\newcommand\mycolon{\mkern2mu{:}\mkern4mu}

\newcommand\mytimes{\mathop{\times}}

\newcommand\mymid{\mathop{\mid}}
\newcommand\myapprox{\mathop{\approx}}

\DeclareMathSymbol{\shortminus}{\mathbin}{AMSa}{"39}
\newcommand{\medminus}{\scalebox{0.6}[0.7]{\(-\)}}
\newcommand{\sminus}{\mathchoice{-}{-}{\medminus}{\shortminus}}

\newcommand{\splus}{\scalebox{0.6}[0.7]{\(+\)}}
\newcommand{\sequal}{\scalebox{0.6}[0.7]{\(=\)}}
\newcommand{\stimes}{\scalebox{0.6}[0.7]{\(\times\)}}

\def\vphi{{\bm{\phi}}}

\def\vtheta{{\bm{\theta}}}
\def\vlambda{{\bm{\lambda}}}
\def\vmu{{\bm{\mu}}}

\def\vk{{\bm{k}}}

\def\vx{{\bm{x}}}

\def\mX{{\bm{X}}}

\DeclareMathAlphabet{\mathsfit}{\encodingdefault}{\sfdefault}{m}{sl}
\SetMathAlphabet{\mathsfit}{bold}{\encodingdefault}{\sfdefault}{bx}{n}

\def\sB{{\mathcal{B}}}

\def\sN{{\mathcal{N}}}

\def\sR{{\mathcal{R}}}

\def\sX{{\mathcal{X}}}

\newcommand{\R}{\mathbb{R}}

\newcommand\Var[2]{\mathrm{Var}_{#1}\left[#2\right]}

\theoremstyle{plain}
\newtheorem{theorem}{Theorem}[section]

\theoremstyle{definition}

\newtheorem{assumption}[theorem]{Assumption}
\theoremstyle{remark}
\newtheorem{remark}[theorem]{Remark}

\long\def\hide#1{}

\title{Learning Where to Simulate: Generative Active Sampling for Online PDE Surrogate Training}

\author{%
  Pierre Cesar \\
  Univ. Grenoble Alpes, Inria, CNRS, Grenoble INP, LIG, Grenoble, France \\
  \texttt{pierre.cesar@inria.fr}
  \And
  Sofya Dymchenko \\
  Univ. Grenoble Alpes, Inria, CNRS, Grenoble INP, LIG, Grenoble, France \\
  \texttt{sofya.dymchenko@inria.fr}
  \And
  Abhishek Purandare \\
  Univ. Grenoble Alpes, Inria, CNRS, Grenoble INP, LIG, Grenoble, France \\
  \texttt{abhishek.purandare@inria.fr}
  \And
  Bruno Raffin \\
  Univ. Grenoble Alpes, Inria, CNRS, Grenoble INP, LIG, Grenoble, France \\
  \texttt{bruno.raffin@inria.fr}
}

\begin{document}

\maketitle

\begin{abstract}
Paper published at NeurIPS 2026.  For data-driven PDE surrogates, generalizing across a wide range of initial conditions and physical coefficients requires a training set of numerical simulations that consistently covers the resulting dynamics. Yet uniform sampling of these configuration parameters can leave challenging regimes insufficiently covered, leading to high prediction errors and large error variance in the trained surrogate. To address this sampling imbalance, we introduce the Online Generative Active Sampling  (OGAS), which continuously adapts configuration sampling during surrogate training, as an alternative to classical round-based active learning. As the surrogate learns from a stream of simulation data generated online, OGAS trains a fast diffusion model in parallel to sample configuration parameters conditioned on a training difficulty signal, such as loss or uncertainty. This conditional sampler enables OGAS to parametrize subsequent simulations to generate challenging training data by prioritizing configurations with high difficulty values. We evaluate OGAS on 2D PDEs with distinct challenging dynamics—Kuramoto–Sivashinsky, Navier–Stokes, and Gray–Scott—using multiple surrogate architectures and up to 308 configuration parameters. Across the tested PDEs and architectures, OGAS consistently reduces the 99th-percentile prediction error and overall error dispersion compared with uniform sampling, while wall-clock overhead is negligible. Code and experiments are available at \url{https://github.com/melissa-sa/ogas}.
\end{abstract}

\etocdepthtag.toc{mtmain}
\section{Introduction}
\label{sec:intro}

\begin{figure}[t]
    \centering
    \includegraphics[width=1\linewidth]{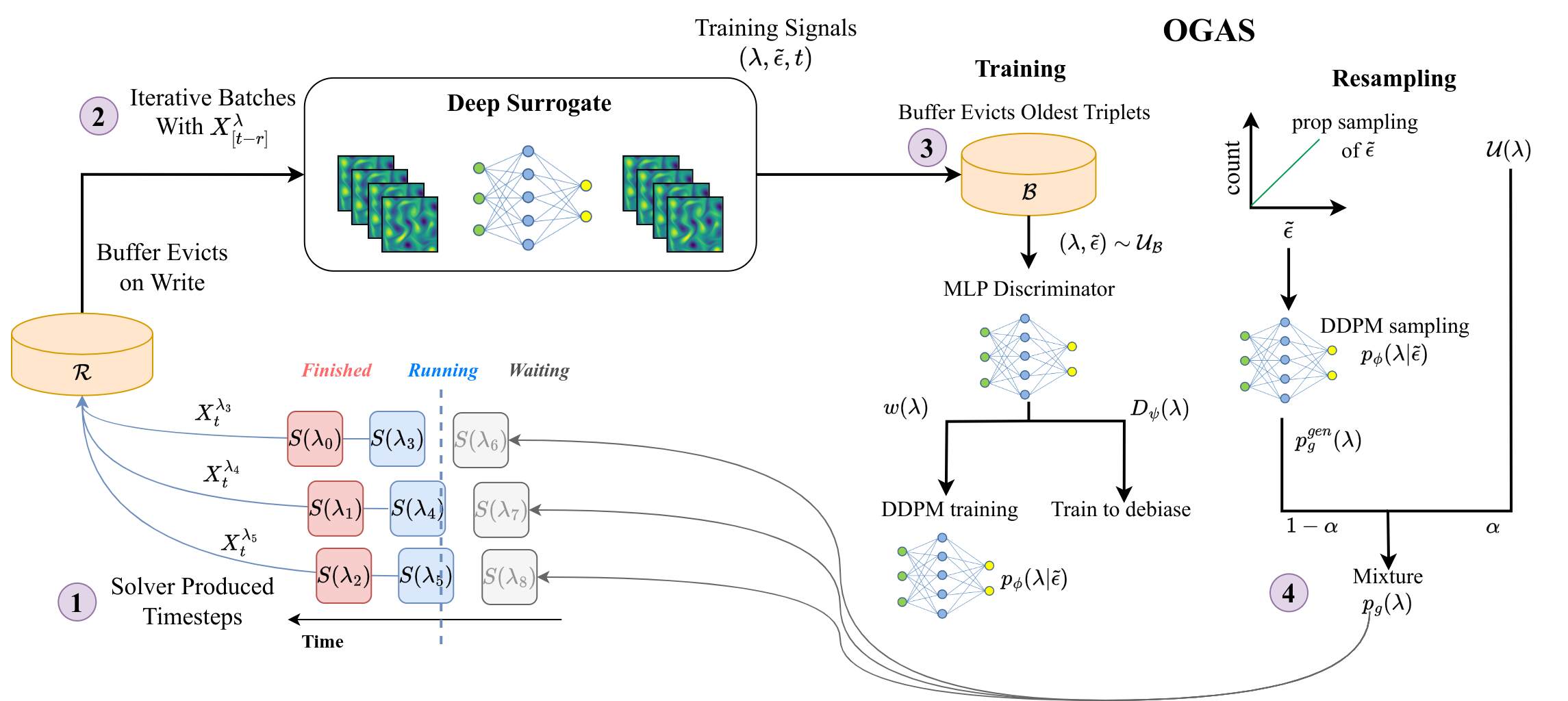}

    \ifdefined\Description
    \Description{Schematic of the online surrogate-training workflow, showing the numerical solvers, data buffer, surrogate ensemble, active-learning signal, and OGAS generator operating in parallel.}
    \fi

    \caption{\textbf{OGAS workflow.} \textbf{(1) Trajectory Streaming:} Solvers, parametrized with different $\boldsymbol{\lambda}$, run concurrently and stream states $X_t^{\boldsymbol{\lambda}}$ into an in-memory buffer $\sR$. \textbf{(2) Surrogate Update:} The surrogate model trains on batches sampled from $\sR$ and computes per-sample difficulty signals $\tilde{\epsilon}$. \textbf{(3) Generative Training:} Pairs $(\boldsymbol{\lambda}, \tilde{\epsilon})$ are buffered in $\sB$ to train a conditional generative model in parallel with the surrogate. \textbf{(4) Adaptive Sampling:} The generative sampler draws new $\boldsymbol{\lambda}$ conditioned on high difficulty values $\tilde{\epsilon}$ to parametrize subsequent solver executions.}
    \label{fig:al-workflow}
  \end{figure}
    
Neural networks that approximate solutions to partial differential equations (PDEs), known as \textit{deep PDE surrogates}, provide learned alternatives to traditional numerical solvers~\cite{pfaff2020learning,li2020fourier,brandstetter2021message,koupai-EfficientGenerativeTransformer-2025,mousavi-RIGNOGraphbasedFramework-2025}.
Data-driven models are typically trained on synthetic data generated by numerical solvers under varying configurations, such as initial conditions, physical coefficients, or geometries.
Their scope spans from compact models trained on a single PDE over restricted configurations to foundation models trained across multiple PDEs and broad parameter spaces~\cite{herde-PoseidonEfficientFoundation-2024,holzschuh-PDEtransformerEfficientVersatile-2025}.

Generalization across these configurations requires training data that adequately cover the resulting dynamics, yet identifying the configurations needed to  generate the required data  is difficult in high-dimensional parameter spaces.
Under a finite simulation budget, uniform sampling may over-represent trivial dynamics (e.g., near-equilibrium states) while under-representing challenging regimes, resulting in surrogates that perform well on average but fail on the hardest inputs~\citep{musekamp-ActiveLearningNeural-2025}.
Moreover, which configurations are difficult depends on the surrogate’s current state and can change during training.
The challenge is therefore not only to generate comprehensive  data, but to allocate simulations according to the surrogate's evolving needs.

Active learning (AL) offers a way to adapt this allocation using feedback from the surrogate~\cite{settles2009active}. AL methods are commonly classified as either pool-based, which select from a finite candidate set, or query-synthesis, which directly generate new samples. Existing AL methods for PDEs take several forms: pool-based approaches~\cite{pestourieActiveLearningDeep2020,musekamp-ActiveLearningNeural-2025}, pool-based approaches complemented with query-synthesis~\cite{wang-PlugandPlayQuerySynthesis-2025}, query-synthesis alone~\cite{dymchenko-SC2024}, or a selector that chooses whether the surrogate or the solver generates a given timestep~\cite{kim-ActiveLearningSelective-2025}.

In this paper we introduce the \textit{Online Generative Active Sampling} (OGAS), a query-synthesis algorithm that departs from previous approaches by relying on a generative model to learn the distribution of configurations conditioned on a difficulty signal. The model is continuously updated as surrogate training produces new training signals, and queried when new configurations are needed to steer data generation. By conditioning the generator on high difficulty levels, OGAS proposes new configurations directly in parameter space. These configurations parametrize  the solvers to generate hard-to-learn data.
This makes OGAS well suited to an online surrogate training workflow~\citep{meyer-SC23}, the context adopted for experiments. The workflow is detailled in Figure~\ref{fig:al-workflow}.
We consider two difficulty signals: per-sample training loss (\textit{OGAS-Loss}) and ensemble disagreement (\textit{OGAS-Unc}). Both reuse predictions from recent training batches, avoiding additional surrogate forward passes for signal acquisitions. To mitigate sampling bias and maintain full space coverage, we correct the diffusion model updates using a density-ratio estimator and mix conditional samples with uniform sampling.


We evaluate OGAS for inferring configuration spaces with up to 308 parameters, for  three 2D PDEs capturing distinct dynamics (Kuramoto--Sivashinsky, Navier--Stokes Kolmogorov flow, and Gray--Scott reaction--diffusion), and  three  surrogate models from distinct families and sizes (U-Net, FNO~\cite{li2020fourier}, and ScOT~\cite{herde-PoseidonEfficientFoundation-2024}).
Under matched budgets of solver runs and surrogate updates, OGAS is compared against uniform and Sobol sampling, online adaptations of pool-based SBAL and Top-$K$~\citep{musekamp-ActiveLearningNeural-2025}, and online adaptive importance sampling~\cite{dymchenko-SC2024}.
Complementary 1D studies examine history-bias correction and sensitivity to key design choices. 

Across the tested 2D settings, OGAS substantially reduces worst-case surrogate errors (up to 2.13$\times$),  99th percentile errors (up to 1.93$\times$) and error standard deviation (up to 2.3$\times$) across settings, at a variable cost to average error (0.84--1.12$\times$).  With the generator running asynchronously on an auxiliary GPU, OGAS adds approximately 0.2\% to training wall-clock time in our implementation.


\section{Related Work} \label{sec:related}

The original goal of AL is to decrease the cost of sample labeling by having the learning algorithm interactively query a teacher for the most relevant samples to label~\citep{settles2009active,actdl2020,cacciarelli-ActiveLearningData-2024}. 
For PDE surrogates, this translates into interactively choosing solver parameters (e.g., ICs, physical parameters) to produce training trajectories that improve the surrogate's performance compared to sampling with a fixed prior. A classical AL approach is to gather a pool of candidate samples to select from. 
Prior works for PDEs build on such \textit{pool-based AL}.
These operate in rounds: surrogate training, parameter selection over a pre-sampled discrete pool based on an acquisition metric, and generation of new data to update the training set. 
\citet{pestourieActiveLearningDeep2020,pestouriePhysicsenhancedDeepSurrogates2023} use ensemble uncertainty estimates for selection \cite{lakshminarayananSimpleScalablePredictive2017}, \citet{bajracharyaFeasibilityStudyActive2024} compare uncertainty \cite{joshiMulticlassActiveLearning2009}, loss \cite{huangSemiSupervisedActiveLearning2021}, and diversity \cite{senerActiveLearningConvolutional2018} metrics, while \citet{vardhanDataEfficientSurrogate2022} use auxiliary acquisition score predictor.
\citet{musekamp-ActiveLearningNeural-2025} (AL4PDE) benchmark several pool-based methods, including SBAL \cite{kirschStochasticBatchAcquisition2023} and LCMD \cite{holzmullerFrameworkBenchmarkDeep2023}. 
Instead of selecting solver parameters, other works select data resolution \citep{liMultiResolutionActiveLearning2024} or time steps \citep{kim-ActiveLearningSelective-2025}.

An alternative to finite pools is \textit{query-synthesis AL}, which directly synthesizes the most informative configurations.
 This fits PDE surrogate training well, where samples can be generated on demand through solver calls rather than selected from a pre-existing dataset. 
Query-synthesis bypasses the pool-filling step and its associated  compute cost. 
It also avoids the discretization induced by the pool, while the solver's parameter space is  continuous and high-dimensional. 
So far, however, query-synthesis for PDE surrogates has received only moderate attention.
\citet{wang-PlugandPlayQuerySynthesis-2025} optimize acquisition scores via gradient ascent in continuous space, but only as an add-on to offline pool-based strategies, showing modest improvements.
 \citet{dymchenko-SC2024} use adaptive multiple importance sampling~\cite{cappe-PMC} to continuously learn a loss-proportional proposal distribution of solver parameters, but with limited scalability and only a local view of the parameter space, as it relies on Gaussian mixture proposals centered around past samples.

Query-synthesis has also been considered for Physics-informed Neural Networks (PINNs),
 guided by loss~\citep{wuComprehensiveStudyNonadaptive2022,tang2023daspinn,dymchenkoLossdrivenSamplingHardtolearn2023,daw_mitigating_2023}, gradient information~\citep{lau-PINNACLEPINNAdaptive-2023}, or uncertainty~\citep{zanisiEfficientTrainingSets2023,aikawaImprovingEfficiencyTraining2024}.  But there training points are sampled within the spatio-temporal domain and evaluated through differentiable surrogate forward passes, eliminating the parametrization and computational costs of solver runs. It is also related to  simulation-based inference (SBI)~\citep{cranmer-FrontierSimulationBasedInference-2020,lyu-DynamicSBIRoundfree-2025}, which uses simulations to train density estimators for posterior inference of system parameters, but SBI models approximate posteriors for a fixed observed system, in contrast to evolving surrogate training dynamics and informative parameters in our setting.
To our knowledge, no existing method provides a scalable active learning approach that operates in continuous parameter spaces, reactively adapts online to evolving surrogate dynamics, and does so without stalling the training workflow (Appendix~\ref{app:online_vs_offline_AL}).

\section{Problem Setting}
\label{sec:preliminaries}
\paragraph{Parametrized PDE and numerical solver.}
Consider a parametrized PDE defined on a periodic bounded spatial domain $\sX\mysubset\R^d$ over a time interval $[0,T]$:
\begin{equation}
    \partial_t u = \sN[u](t,\vx;\vlaD), \quad (t,\vx) \in [0, T]\times \sX; \quad  u(0,\vx) = u_0(\vx; \vlaIC), \quad \vx \in \sX;
\end{equation}
where $u(t,\vx)\mycolon[0,T]\mytimes\sX\to\R^{\Nc}$ is the solution with $\Nc$ channels, $\sN[\cdot]$ is a differential operator parametrized by physical coefficients $\vlaD$, and $u_0(\cdot;\vlaIC)$ is a parametric initial condition (IC).
We group physical and IC parameters into a single solver parameters vector $\vla\eq(\vlaD,\vlaIC) \myin\Lambda\mysubset\R^{\sbs{N}{D}\splus\sbs{N}{IC}}$.

For given $\vla$, a spatial grid of size $\Nx$, and a timestep size $\Delta_{\dt}$, a numerical solver $S$ produces an approximate solution discretized in space and time in an autoregressive manner. Let $\mX^{\la}_{t} \myin \R^{\Nx\stimes\Nc}$ denote the discrete solution state at timestep $t$. A sequence of $\Nt$ states over time, i.e., a trajectory, is produced by evaluating $u_0(\cdot; \vlaIC)$ at grid points to obtain $\mX^{\la}_0$. and then applying the solver:
\begin{equation}
    \mX^{\la}_{t} = S(\mX^{\la}_{t\sminus 1}, \vlaD) \quad \text{for } t \eq 1,\ldotscomp,\Nt, \quad \text{where } \, \mX^{\la}_{t} \approx u(t\cdot\Delta_{\dt}, \cdot).
\end{equation}


\paragraph{Surrogate training.}
We aim to approximate the solver $S$ with a $\vlaD$ conditioned autoregressive surrogate $\NN$ parametrized by $\vtheta$. We adopt a $r$-step training strategy, following \cite{musekamp-ActiveLearningNeural-2025}.
Let $\mX^{\la}_{[t\sminus r]}$ denote a training sample consisting of states $\{\mX^{\la}_{t\sminus r}, \ldotscomp, \mX^{\la}_{t}\}$ and  vector $\vlaD$, $\NN^{(i)}$ be the $i$-th autoregressive application, then the $i$-th step MSE loss is:
\begin{equation}\label{eq:i_mse_loss}
\ell_i(\xtlapre{r}, \xtlapre{r \splus i}; \vtheta) = \frac{1}{\Nx\Nc}\big\| \NN^{(i)}(\xtlapre{r}, \vlaD) - \xtlapre{r \splus i}\big\|_{2}^2.
\end{equation}
The average of the $r$ losses, i.e., $\ell(\xtla; \vtheta) \eq\frac{1}{r}\sum_{i\sequal 1}^{r} \ell_i(\xtlapre{r}, \xtlapre{r \splus i}; \vtheta)$, is used for training updates.




Training is performed online~\cite{meyer-SC23}, as detailled in figure~\ref{fig:al-workflow},  with a  simulation budget of $N$ solver executions. During training, $N_S\ll N$ solver instances run concurrently and stream the produced timesteps to the surrogate model for training. By default, the parameters $\lambda$ of the solver instances are sampled uniformly. Using AL, the sampling distribution $p(\boldsymbol{\lambda})$ is dynamically adapted to generate more relevant samples.



\section{Online Generative Active Sampling}\label{sec:method}
We propose \textit{Online Generative Active Sampling} (OGAS), a query-synthesis AL method that learns a conditional distribution over solver parameters $\vla$ given a sample informativeness (training signal). 
At its core, a fast generative model (generator) is trained asynchronously alongside the surrogate to reactively capture this evolving distribution.
Future solver parameters are sampled from a mixture combining a uniform component, ensuring global coverage of $\Lambda$, and a generator-guided component, using a user-defined proposal distribution over the signal. 
The generator objective is corrected to mitigate sampling bias. Implementation and theoretical details are provided in Appendices~\ref{app:theory},~\ref{app:al_training}. We summarize the method in Alg.~\ref{alg:ogas_compact}
and illustrate the workflow in Fig.~\ref{fig:al-workflow}.

\paragraph{Training signal.}\label{sec:signal_conditional}
For each sample $\xtla$ in a training batch, we compute a \textit{training signal} with a function $\eps = A(\xtla;\vtheta)\in\R_{+}$, which is designed to quantify informativeness of the sample at this training step $\vtheta$.
Our main approach employs a \textit{loss-based} signal, treating surrogate loss as a measure of trajectory difficulty and thus informativeness; specifically, it is a scaled last-step MSE loss divided by the $L^2$ norm of the target state:
\begin{equation}\label{eq:normalized_loss}
  A_{\ell}(\xtla;\vtheta) = \frac{\ell(\xtla;\vtheta)}{\|\mX^\la_t\|_2+\delta},
\end{equation}
where $\delta>0$ is added to avoid division by near-zero values. Scaling prevents excessive prioritization of high-amplitude errors driven by the magnitude of the target state.
Alternatively, we consider an \textit{uncertainty-based} signal, following prior AL work \cite{musekamp-ActiveLearningNeural-2025}, which relies on an ensemble of $M$ identical models with independently initialized and trained weights \cite{seungQueryCommittee1992}. Let $\left\{\vtheta_{m}\right\}_{m\sequal1}^M$ denote the parameters of the $M$ surrogates of the ensemble.  Then, $A$ is defined as the total variance of last-step predictions across the ensemble, averaged over all channels and spatial locations:
\begin{equation}\label{eq:uncertainty}
  A_{u}(\xtla;\vtheta) = \Var{m}{S^{(r)}_{\theta_m}(\xtlapre{r},\vlaD)}.
\end{equation}
As raw loss magnitudes decay as the surrogate converges, we z-standardize training signals using running batch statistics.
At each surrogate update, we compute the current batch statistics to update a moving mean $\mu$ and standard deviation $\sigma$, yielding the normalized signal $\tilde\eps \eq (\eps-\mu)/{\sigma}$. 

\textbf{Buffer.} We log pairs $(\vlambda,\tilde\eps)$ into a dedicated fixed-capacity \textit{buffer} $\sB$ (Fig.~\ref{fig:al-workflow}), where incoming pairs overwrite the oldest entries to prioritize recent data. Since $\tilde\eps$ is observed for particular timestep but OGAS samples only solver parameters, we marginalize over time when logging to $\sB$. Under the assumption that the reservoir $\sR$ samples are drawn approximately uniformly over timesteps~\cite{meyer-ICML23}, the buffer $\sB$, represents a timestep-marginal joint distribution over $(\vlambda,\tilde\eps)$. We denote by $p_{\sB}(\vlambda, \tilde\eps)$ the distribution obtained by sampling a pair uniformly from \textit{the current contents} of $\mathcal{B}$, with marginal $p_{\sB}(\vlambda)$, and conditional $p_{\sB}(\vlambda \mid \tilde\eps)$ capturing which $\vla$ tend to produce trajectory samples with difficulty level $\tilde\eps$.

\paragraph{Signal-conditional generator.}\label{sec:generator}
We train a conditional generative model, the \textit{generator}, to approximate this buffer conditional by minimizing
\begin{equation}
\label{eq:generative_objective_def}
\mathbb{E}_{(\vlambda,\tilde\eps)\sim p_{\sB}}
\big[-\log p_\phi(\vlambda \mid \tilde\eps)\big].
\end{equation}
This objective is model-agnostic: it defines the conditional target of the active sampler, not the specific training loss of some generative model.
In our implementation, $p_\phi$ is a fast conditional DDPM~\cite{ho_denoising_2020} that operates with the solver parameters vector $\vlambda$ of size $N_{\text{D}}+N_{\text{IC}}$; it is trained with the standard conditional noise-prediction objective detailed in Appendix~\ref{app:diffusion}.

Generator training proceeds in short cycles, each triggered as new $(\vlambda, \tilde\eps)$ pairs arrive from surrogate training. These inserted pairs in $\sB$ define the empirical training distribution.
At each cycle, OGAS executes $K$ optimization steps on mini-batches sampled uniformly from $\sB$ before the next buffer update occurs.
By warm-starting the generator from its previous state, the model tracks a sequence of evolving objectives rather than a stationary distribution. In practice, setting $K$ sufficiently large ensures the generator adapts quickly relative to buffer drift.
Because historical buffer entries reflect previous sampling policies, we correct the induced history bias using the density-ratio reweighting introduced below in Eq.~\eqref{eq:iw_objective_uniform_compact}.

\paragraph{Sampling next parameters.}\label{sec:ic_sampling}
OGAS is regularly called to sample parameters for the upcoming solver instances.
We denote by $p_g(\vla)$ the $g^\text{th}$ \textit{generation} of the distribution  used  for sampling  these parameters. 
The mixture policy $p_g(\vla)$  combines a global uniform distribution $\mathcal{U}_\Lambda$ with the generator-guided policy $p^{\mathrm{gen}}_g(\vla)$:
\begin{equation}
  p_g(\vlambda) = \alpha\,\mathcal{U}_\Lambda + (1-\alpha)\,p^{\mathrm{gen}}_g(\vlambda)  \quad \text{ where }  \alpha \in (0,1).
\label{eq:mixture_policy}
\end{equation}
The uniform component preserves global coverage of $\Lambda$, while the generator-guided component concentrates solver runs in regimes that are currently difficult for the surrogate.

The policy $p^{\mathrm{gen}}_g(\vlambda) = \int k(\tilde\eps)\,p_{\phi_g}(\vlambda \mid \tilde\eps)\,d\tilde\eps$ marginalizes over a target signal distribution $k(\tilde\eps)$. 
To draw from $p^{\mathrm{gen}}_g$, we first sample a target signal level $\tilde\eps \sim k(\tilde\eps)$ and then generate parameters $\vlambda \sim p_{\phi_g}(\vlambda \mid \tilde\eps)$. 
We design $k(\tilde\eps) \propto \tilde\eps\,\mathbf{1}_{\mathcal{A}}(\tilde\eps)$ to emphasize high error regimes and reduce tail surrogate error.
 Here, $\mathbf{1}_{\mathcal{A}}$ restricts target signals to an admissible range $\mathcal{A} = [0, 5]$.
  Because $\tilde\eps$ is z-normalized, this range discards below average signals ($\tilde\eps < 0$) and concentrates sampling on challenging trajectories up to 5 standard deviations above the mean, while avoiding extreme numerical outliers.

\begin{algorithm}[t]
  \caption{OGAS: Online Generative Active Sampling}
  \label{alg:ogas_compact}
  \begin{algorithmic}[1]
  \Require mixture $\alpha\in(0,1)$; admissible signal set $\mathcal{A}=[\tilde\eps_{\min},\tilde\eps_{\max}]$;
  proposal~$k(\tilde\eps)\propto \tilde\eps\,\mathbf{1}_{\mathcal{A}}(\tilde\eps)$; EMA decay $\gamma$; generator updates per cycle $K$; weight clip $[w_{l},w_{r}]$.
  \Ensure generator $p_\phi(\lambda\mid\tilde\eps)$; discriminator $D_\psi(\lambda)\in(0,1)$;
  buffer $\mathcal{B}$ storing $(\lambda,\tilde\eps)$; $\mu\gets 0,\sigma\gets 1$.
  \While{training}
    \State \textbf{(1) Log standardized signals:}
    \State Receive batch $\{(\lambda_i,\eps_i)\}_{i=1}^n$.
    \State Compute minibatch $\{\eps_i\}$ statistics $(\mu_{\text{mb}},\sigma_{\text{mb}})$.
    \State Update moving statistics: $\mu\gets \gamma\mu+(1-\gamma)\mu_{\text{mb}}$ and $\sigma\gets \gamma\sigma+(1-\gamma)\sigma_{\text{mb}}$.
    \State Standardize $\tilde\eps_i\gets (\eps_i-\mu)/\sigma$ and push $\{(\lambda_i,\tilde\eps_i)\}_{i=1}^n$ to $\mathcal{B}$.
    \State \textbf{(2) Train the generator:}
    \For{$s=1,\dots,K$}
      \State Sample $\{(\lambda_j,\tilde\eps_j)\}_{j=1}^m \sim \mathcal{B}$. \Comment{$y{=}1$ (buffer) class}
      \State Sample $\{\bar\lambda_j\}_{j=1}^m \sim \mathcal{U}_\Lambda$. \Comment{$y{=}0$ (uniform) class}
      \State Update $D_\psi$ on $\{(\bar\lambda_j,0)\}_{j=1}^m \cup \{(\lambda_j,1)\}_{j=1}^m$,
      where $D_\psi(\lambda)=P_\psi(y=0\mid\lambda)$.
      \State Set $\widehat w_{\mathcal{B}}(\lambda_j)\gets \frac{D_\psi(\lambda_j)}{1-D_\psi(\lambda_j)}$ and clamp to $[w_{l},w_{r}]$.
      \State Update $\phi$ with weighted loss Eq.~\eqref{eq:iw_objective_uniform_compact}.
    \EndFor
    \State \textbf{(3) If resampling request:}
    \For{each new solver run}
      \State $\lambda \sim \mathcal{U}_\Lambda$ with probability $\alpha$, else $\tilde\eps \sim k(\tilde\eps)$ and $\lambda \sim p_\phi(\lambda\mid \tilde\eps)$.
    \EndFor
  \EndWhile
  \end{algorithmic}
  \end{algorithm}

\paragraph{History-bias correction and uniform-prior training.}
\label{sec:history_bias_correction}
\hide{We denote by $p_{\mathcal{B}}(\vlambda,\tilde\eps)$ the time marginal joint
distribution obtained by sampling a pair uniformly from the \textit{current contents} of $\mathcal{B}$
(i.e., within the time-local window represented by the fixed capacity buffer). Its marginal and conditional
are $p_{\mathcal{B}}(\vlambda)$ and $p_{\mathcal{B}}(\tilde\eps\mymid\vlambda)$.}
Given that minimization of Eq.~\eqref{eq:generative_objective_def} learns
$p_{\phi}(\vlambda\mymid\tilde\eps)\approx p_{\mathcal{B}}(\vlambda\mymid\tilde\eps)$, and applying Bayes' rule,
\begin{equation}
p_{\mathcal{B}}(\vlambda\mid\tilde\eps)\propto p_{\mathcal{B}}(\vlambda)\,p_{\mathcal{B}}(\tilde\eps\mid\vlambda),
\label{eq:buffer_bayes_compact}
\end{equation}
the learned conditional $p_{\phi}(\vlambda\mymid\tilde\eps)$ inherits the generally non-uniform buffer prior $p_{\mathcal{B}}(\vlambda)$ from previous sampling generations.
Consequently, reusing $p_{\phi}(\vlambda\mymid\tilde\eps)$ to propose future parameters can induce a \textit{history bias}
that persistently favors historically over-sampled regions of $\Lambda$ and potentially keeps the generator focused on regions that are no longer relevant as the surrogate evolves (Appendix~\ref{app:theory_bias}).
To debias the generator, we train it under a uniform reference prior by targeting the joint distribution
\begin{equation}
p_{\sB}^\star(\vlambda,\tilde\eps)
=\mathcal{U}_\Lambda(\vlambda)\,p_{\sB}(\tilde\eps\mid\vlambda).
\label{eq:target_joint_unif_main}
\end{equation}
We minimize the corresponding importance-weighted log-likelihood.
Using samples from $\mathcal{B}$, we implement this objective by importance weighting \cite{bickel_discriminative_2009}:
\begin{equation}
\label{eq:iw_objective_uniform_compact}
\mathcal{L}_{\mathrm{unif}}(\vphi)
=\mathbb{E}_{(\vlambda,\tilde\eps)\sim p_{\mathcal{B}}}\!\left[-w_\mathcal{B}(\vlambda)\,\log p_\phi(\vlambda\mid\tilde\eps)\right],
\end{equation}
with $w_{\mathcal{B}}(\vlambda) = \mathcal{U}_\Lambda(\vlambda)/p_{\mathcal{B}}(\vlambda)$ the corresponding density ratio.

\textbf{Density ratio estimate.}
\label{sec:density_ratio}
The weight \(w_{\mathcal{B}}(\vlambda)\) is not available in closed form since the buffer marginal \(p_{\mathcal{B}}(\vlambda)\) is induced by a history-dependent sampling policy. We therefore estimate the density ratio via binary classification. We train a discriminator \(D_\psi(\vlambda)\in(0,1)\) to distinguish samples from the uniform reference \(\mathcal{U}_\Lambda\) (\(y{=}0\)) and from the buffer marginal \(p_{\mathcal{B}}\) (\(y{=}1\)), i.e., \(D_\psi(\vlambda)=P(y{=}0\mid \vlambda)\). The implied density-ratio estimator is
\begin{equation}
  \label{eq:ratio_from_disc_compact}
\widehat w_{\mathcal{B}}(\vlambda)=\frac{\pi_1}{\pi_0}\cdot\frac{D_\psi(\vlambda)}{1-D_\psi(\vlambda)}.
\end{equation}
where \(\pi_c=P(y{=}c)\) are the class priors \cite{goodfellow_generative_2014}.
In practice we use balanced classes (\(\pi_0=\pi_1\)), so the prefactor cancels. We clip \(\widehat w_{\mathcal{B}}(\vlambda)\) to \([w_l,w_r]\) for numerical stability.

\section{Experimental Setup}\label{sec:exp}

\subsection{Data Generation}\label{sec:exp_benchmarks}
\textbf{2D PDEs.} We use the pseudo-spectral solvers from  APEBench~\cite{koehler-APEBenchBenchmarkAutoregressive-2024} for three high-order nonlinear PDEs: Kuramoto-Sivashinsky (KS), Navier-Stokes Kolmogorov Flow (NS), and unsteady Gray-Scott-$\beta$ (GS). The first two model chaotic and turbulent fluid dynamics, while the third captures continually evolving reaction--diffusion pattern formation in a chemical system ($N_c\eq2$).
Each PDE is defined on a 2D periodic domain $\sX \eq [0,L]^2$ discretized into $\Nx \eq 256^2$ grid points. To ensure numerically stable and accurate trajectories, solvers run at a fine time resolution $\Delta t_{\text{eff}}$, but we expose only one state every $\Delta t$ timesteps for training or validation. 
We also discard the first $N_{t_w}$ states to avoid initial transients.Training uses trajectories of $N_t\eq32$ timesteps, while validation uses $N_t\eq51$ timesteps to assess generalization beyond the training horizon. These choices closely follow the established setup of \citet{holzschuh-PDEtransformerEfficientVersatile-2025} (Appendix~\ref{app:pde_solver_config}).

\textbf{Solver parameters.}
Each solver takes an input parameter $\vlambda\eq(\vlambda_{\text{D}}, \vlambda_{\text{IC}})$ as described in Sec.~\ref{sec:preliminaries}. 
ICs for KS and NS are \textit{truncated Fourier series} with up to $8$ modes and varied amplitudes and phases per mode, yielding $N_{\text{IC}}\eq 308$ parameters. ICs for GS are superpositions of $8$ \textit{Gaussian blobs} with varied locations, variances, and intensities, accounting for $N_{\text{IC}}\eq48$ parameters (Appendix~\ref{app:ic_families}).
To capture diverse physical regimes, KS and GS domain sizes $\vlambda_{\text{D}}\eq L$ vary within the $[10.0, 130.0]$ and $[0.5, 10.0]$ intervals, respectively; NS viscosity coefficient $\vlambda_{\text{D}}\eq\nu$ is defined on $[0.003, 0.008]$, corresponding to Reynolds numbers $Re \in [125, 330]$. Appendix~\ref{app:impl} shows example trajectories across the parameter space in Figures~\ref{fig:pde_gray_scott},~\ref{fig:pde_navier_stokes}, and~\ref{fig:pde_ks}; the GS and KS examples illustrate the wide range of dynamics as $L$ varies, while the NS examples show narrower variation across $\nu$.

\subsection{Compared Methods} \label{sec:exp_baselines}

We evaluate two variants of the proposed \textit{OGAS} method: \textit{OGAS-Loss} uses the training signal $A_{\ell}$ from Eq.~\eqref{eq:normalized_loss}, while  \textit{OGAS-Unc} uses $A_u$ from Eq.~\eqref{eq:uncertainty} (Appendix~\ref{app:ogas_implementation}). We compare them against the following baselines and state-of-the-art AL methods. \textbf{Standard baselines:} We use \textit{Uniform} random sampling and quasi-uniform sampling with \textit{Sobol} sequences, providing a low-discrepancy space-filling coverage of $\Lambda$.
\textbf{Pool-based AL:} We take two pool-based approaches, \textit{SBAL} and \textit{Top-$K$}, with ensemble-based uncertainty estimates from AL4PDE~\cite{musekamp-ActiveLearningNeural-2025}, and adapt them to the online setting. Instead of maintaining a fixed pool with eviction as in the original works, we uniformly resample a new candidate pool from $\Lambda$ for each selection phase.
 These methods estimate uncertainty using the variance over full-rollout trajectory predictions from the ensemble (Appendix~\ref{app:sbal}).
  Specifically, every 1{,}000 completed simulations, we pause data generation and surrogate training to score 5{,}000 newly sampled candidate parameters and select 1{,}000 for the next solver runs. We choose this pool size to balance selection quality and compute efficiency, as most compute resources are left idle during the AL procedure (overheads are profiled in Sec.~\ref{sec:exp_results}).
  \textbf{Online AL:} We use \textit{Breed} from \citet{dymchenko-SC2024} where a loss-based posterior is modeled with a Gaussian mixture proposal using an adaptive importance sampling approach; the mixture is periodically updated and combined with uniform sampling. To isolate the effect of the sampling strategy, i.e., generative model vs. Gaussian mixture, we use the same update period, mixture rate $\alpha$, and loss-based training signal as in \textit{OGAS-Loss} (Appendix~\ref{app:breed_impl}).

\begin{figure}[t]
  \centering
  \includegraphics[width=\linewidth]{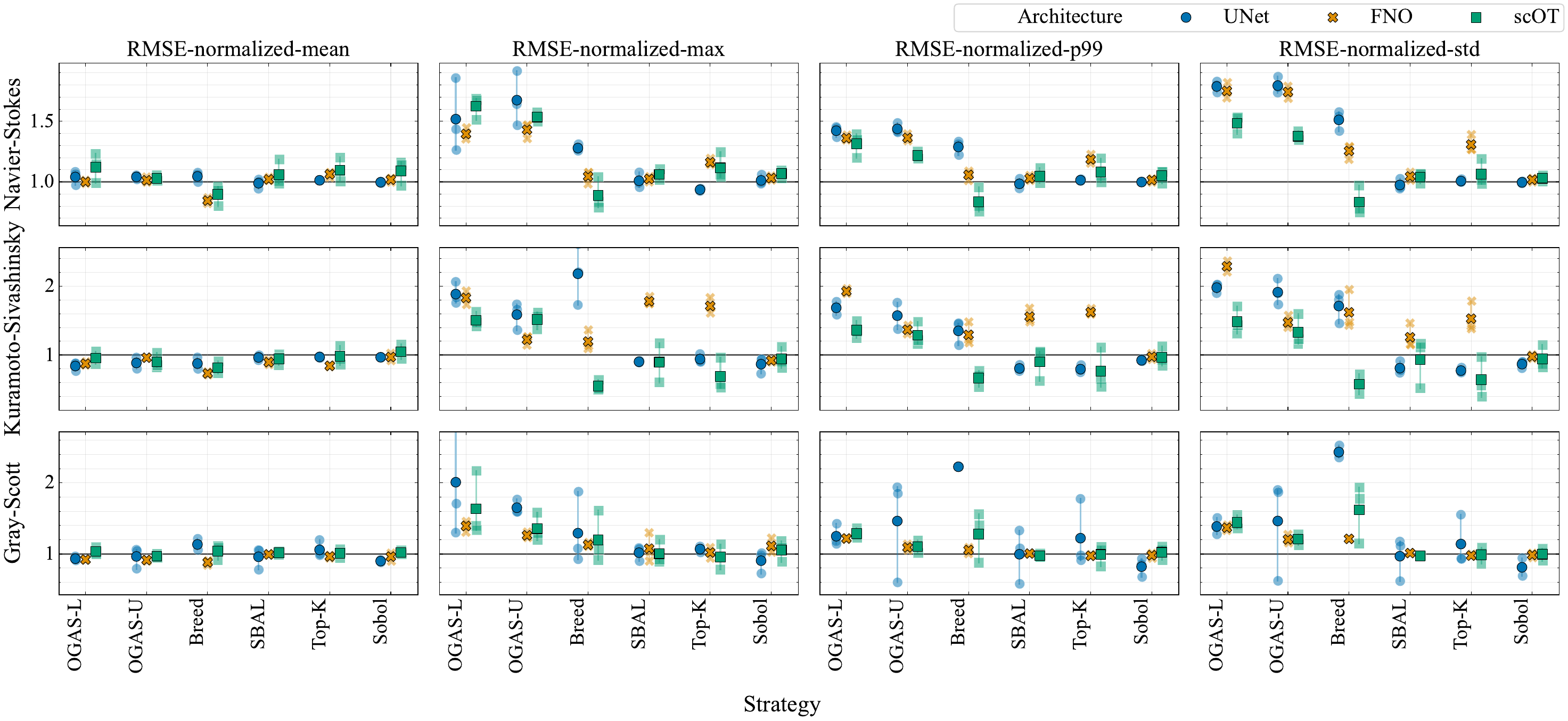}
  \Description{Point plots of improvement ratios over uniform sampling for mean, maximum, 99th-percentile, and standard-deviation normalized RMSE across the tested PDEs, surrogate architectures, and active-learning methods.}
  \caption[Improvement ratios over Uniform for normalized RMSE statistics.]{Improvement ratios over Uniform for normalized RMSE statistics ($\mathrm{metric}(\textit{Uniform})/\mathrm{metric}(\textit{method})$); values above one indicate lower error than Uniform. Columns report mean, maximum, p99, and standard deviation; rows report PDEs; marker color/shape denotes the surrogate architecture. Each connected triplet shows the three seeds for one PDE--architecture--method setting, and bold points  their average.}
  \label{fig:metrics}
\end{figure}

\subsection{Surrogates Training and Validation}\label{sec:exp_protocol}

For each experiment, we train an ensemble of two surrogates ($M\eq2$) with different initializations, shared batches from $\sR$, and a 2-step MSE loss ($r\eq2$, Eq.~\ref{eq:i_mse_loss}). The ensemble supports uncertainty methods (\textit{OGAS-Unc}, \textit{SBAL}, \textit{Top-$K$}); loss-based methods (\textit{OGAS-Loss}, \textit{Breed}) use the average member loss for comparable training conditions. Single-surrogate \textit{OGAS-Loss} results are also given in Appendix~\ref{app:single_model_eval}.
For surrogates, we use three distinct backbone architectures of increasing parameter counts: UNet (2.3M) and FNO (29.5M)~\cite{li2020fourier} from AL4PDE are adapted to our settings, and scOT (184.3M) is adapted from \cite{herde-PoseidonEfficientFoundation-2024} for conditional autoregressive prediction (Appendix~\ref{app:model_architecture}). Optimization and hyperparameters follow AL4PDE (Appendix~\ref{app:model_training_details}).

For validation, we use ensemble-average predictions, $S_{\theta}(\cdot) \eq \frac{1}{M}\sum_{m\sequal1}^M S_{\theta_m}(\cdot)$, computed after the final training step over a held-out validation set $\mathcal{D}_v$ of 750 trajectories whose parameters $\vlambda$ are drawn from a Halton sequence, evaluated at $t\myin\{1,\ldotscomp,51\}$ (total of $37{,}500$ samples). Performance is measured with the normalized 1-step RMSE
\begin{equation}\label{eq:normalized_error}
\text{RMSE}(\vla, t;\vtheta) = \sqrt{\ell_1(\mX^{\la}_{t\sminus1}, \mX^{\la}_{t}; \vtheta)} \big/ \left(\|\mX^{\la}_{t}\|_2+\delta\right),
\end{equation}
summarized through its mean (\textit{RMSE-mean}), maximum and 99th percentile (\textit{RMSE-max}, \textit{RMSE-p99}, tail robustness), and standard deviation (\textit{RMSE-std}, consistency).  Validating AL methods against a uniformly sampled dataset, as done in this paper, tends to obfuscate the achieved error reduction. AL methods are designed to concentrate training on hard-to-learn regions. Because these regions typically occupy only a small fraction of the parameter space, they represent a similarly small portion of a uniformly sampled dataset. Consequently, AL performance gains usually materialize through \textit{RMSE-max}, \textit{RMSE-p99} and  \textit{RMSE-std} rather than \textit{RMSE-mean}.

Experiments ran on the Leonardo supercomputer. To keep the compute budget accessible, experiments use low compute-cost solvers (compared to some production ones) from APEBench~\cite{koehler-APEBenchBenchmarkAutoregressive-2024} and are evaluated over three random seeds, in line with standard practice in prior AL benchmarks for high-resolution PDE surrogates~\citep{musekamp-ActiveLearningNeural-2025,kim-ActiveLearningSelective-2025}. Accounting for all experiments required for this paper, including preparatory ones, leads to a total of about 15K GPU hours and 785K core hours.

We use the Melissa framework to enable online surrogate training~\citep{Schouler-JOSS23} and extended it to support AL. To ensure a sound basis for comparisons, we train the PDE surrogates with the same number of gradient updates, training batches, and total number of solver executions. To achieve this goal within the Melissa framework, which is designed to favor performance through asynchronism, we added a flow control to constrain the surrogate update rate to a fixed ratio of gradient updates per solver execution (Appendix~\ref{app:speed_limiter}).

The total simulation budget is set to $N=10{,}000$ solver executions with $N_S\eq56$ parallel solver instances running on 2 CPU cores each. Every surrogate completes $18$k gradient updates, requiring approximately 3-4.5 hours using 2 NVIDIA A100 GPUs (one per ensemble member), depending on the backbone architecture. \textit{OGAS} runs its AL process asynchronously on a separate NVIDIA A100 GPU. This standalone GPU for OGAS ensures minimal interference with the main GPUs dedicated to surrogate training. Extra experiments where OGAS instead shares one of the training GPUs show longer wall-clock training times, but no impact on surrogate quality (Appendix~\ref{app:gpu_time_sharing}).

\section{Experimental Results}\label{sec:exp_results}

Figure~\ref{fig:metrics} reports improvement ratios relative to \textit{Uniform}, computed as \nolinebreak{$\mathrm{metric}(\textit{Uniform})/\mathrm{metric}(\textit{method})$} for each PDE--architecture--seed setting. Full numerical results are reported in Appendix~\ref{app:2d_results}, and per-PDE example trajectory predictions for \textit{OGAS-Loss} and \textit{Uniform} appear in appendix Figures~\ref{fig:ns_samples},\ref{fig:ks_samples},\ref{fig:gs_samples}.

\begin{figure}[t]
  \centering
  \includegraphics[width=1.0\linewidth]{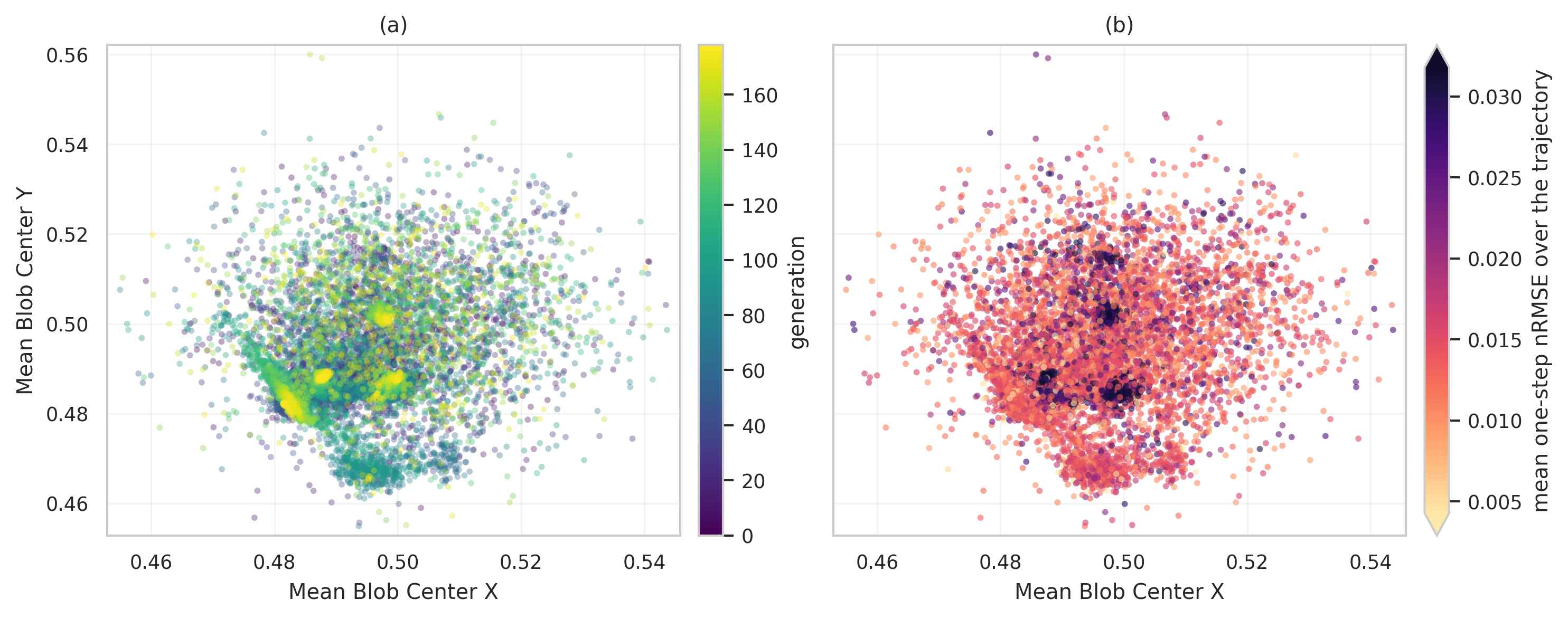}
  \caption{scOT trained with OGAS on Gray--Scott. Mean blob center coordinates of the parameter samples:
  evolution across generator updates (left) and  final surrogate's loss per trajectory for each parameter sampled during training (right). }
  \label{fig:gs_sampled_distribution_evolution}
\end{figure}

\textbf{Tail compression.}
The dominant effect of \textit{OGAS} is a strong compression of the high-error tail, directly improving robustness (\textit{RMSE-max}, \textit{RMSE-p99}) and consistency (\textit{RMSE-std}). The right three columns of Fig.~\ref{fig:metrics} show that the seed-averaged ratios sit above one in nearly every PDE--architecture cell for both variants, with cross-cell averages of $1.64$/$1.42$/$1.66$ (max/p99/std) for \textit{OGAS-Loss} and $1.47$/$1.32$/$1.50$ for \textit{OGAS-Unc}. The gain is consistent across PDEs and architectures, and largest on the more diverse parameter ranges (KS, GS).
This effect arises from a reallocation of simulation budget within the same parameter domain $\Lambda$ rather than any expansion of either of them. OGAS is able to reactively identify and focus on the most challenging regimes as training progresses, while the $\alpha$-weighted uniform component ensures that no region is completely neglected. The debiasing correction and fast training cycles further enhance this adaptivity, preventing overfitting to past signals. Figure~\ref{fig:gs_sampled_distribution_evolution} illustrates this for Gray--Scott/scOT, where generated parameters progressively concentrate on blob-location regions at different sampling stages. 

Compared to the Gaussian mixture proposal in \textit{Breed}, the generative model can capture complex, multimodal distributions globally across $\Lambda$ and adapt them flexibly over time, without being constrained to a fixed parametric family. \textit{Breed} achieves tail improvements but inconsistently, i.e., $1.17$/$1.23$/$1.42$ (max/p99/std) with high cross-cell variability. Pool-based methods show similarly mixed and smaller gains despite uncertainty-driven selection, and \textit{Sobol}'s low-discrepancy coverage alone is insufficient to consistently target the most challenging regimes.

\textbf{Mean trade-off.}
On \textit{RMSE-mean}, \textit{OGAS-Loss} and \textit{OGAS-Unc} reach $0.97$ and $0.96$ on average, a modest cost relative to their substantially larger tail and consistency gains. No competing method matches this trade-off: \textit{Breed}'s tail improvements come at a markedly larger mean cost ($0.92$), while \textit{SBAL}, \textit{Top-K}, and \textit{Sobol} stay close to \textit{Uniform} on the mean ($0.98$, $1.00$, $1.00$) without yielding the tail gains discussed above. Mean degradations concentrate in the lower-capacity UNet/FNO runs on KS and GS, whereas scOT and narrower-range 2D checks preserve the tail gains with improved means (Appendix~\ref{app:rel_improvments_pde_and_arch},~\ref{app:narrower_range}). Note that the validation set is itself sampled uniformly over $\Lambda$, which  gives \textit{Uniform} a slight advantage on \textit{RMSE-mean}. 
On a validation set sampled by the trained \textit{OGAS} generator, \textit{OGAS} significantly outperforms \textit{Uniform} on \textit{RMSE-mean} (Figure~\ref{fig:hard_set_improvment} in Appendix~\ref{app:hard_ddpm_eval}).

\textbf{Loss vs. uncertainty.}
\textit{OGAS-Loss} emerges as a more consistent variant than \textit{OGAS-Unc}. On Gray--Scott, ensemble disagreement with $M{=}2$ is noisy: for UNet, \textit{OGAS-Unc} shows much larger p99 variability ($4.89\pm3.55$) than \textit{OGAS-Loss} ($4.35\pm0.50$), with the same pattern carrying over to p95/p99 (Table~\ref{tab:gray-scott-beta-low-res}). On the smoother Navier--Stokes cases, \textit{OGAS-Unc} can match or slightly outperform \textit{OGAS-Loss}, suggesting that uncertainty guidance remains useful when estimated sufficiently accurately.  Conditioning on the energy CRPS of the ensemble instead of its disagreement brings \textit{OGAS-Unc} to the tail errors of \textit{OGAS-Loss} (Appendix~\ref{app:crps_ablation}). Further increasing the ensemble size showed limited gains on the uncertainty estimate  (Appendix~\ref{app:ensemble_size}). A larger ensemble would also increase the computational overhead, especially for large-scale surrogates, while loss-based signals are directly available from training with minimal extra cost.

\textbf{Wall-time overhead.} As \textit{OGAS} and \textit{Breed} perform AL asynchronously with surrogate training, the only wall-time overhead comes from sending training signal batches to the generator process: \textit{OGAS} adds 23s ($\myapprox$0.2\%) and \textit{Breed} adds 70s ($\myapprox$0.7\%), when compared to the wall-time of surrogate training without AL. In contrast, pool-based methods introduce a substantial inference bottleneck that scales with the pool size, resampling frequency, but more importantly the surrogate architecture size and PDE resolution that determine the inference time per candidate. In our setup with moderate pool size and resampling frequency, we observe a 17.6mn ($\myapprox$10\%) overhead for \textit{SBAL} (Appendix~\ref{app:gpu_overhead}).

\subsection{Small Scale Tests and Ablations}
\begin{wrapfigure}{r}{0.40\columnwidth}
  \centering
  \includegraphics[width=\linewidth]{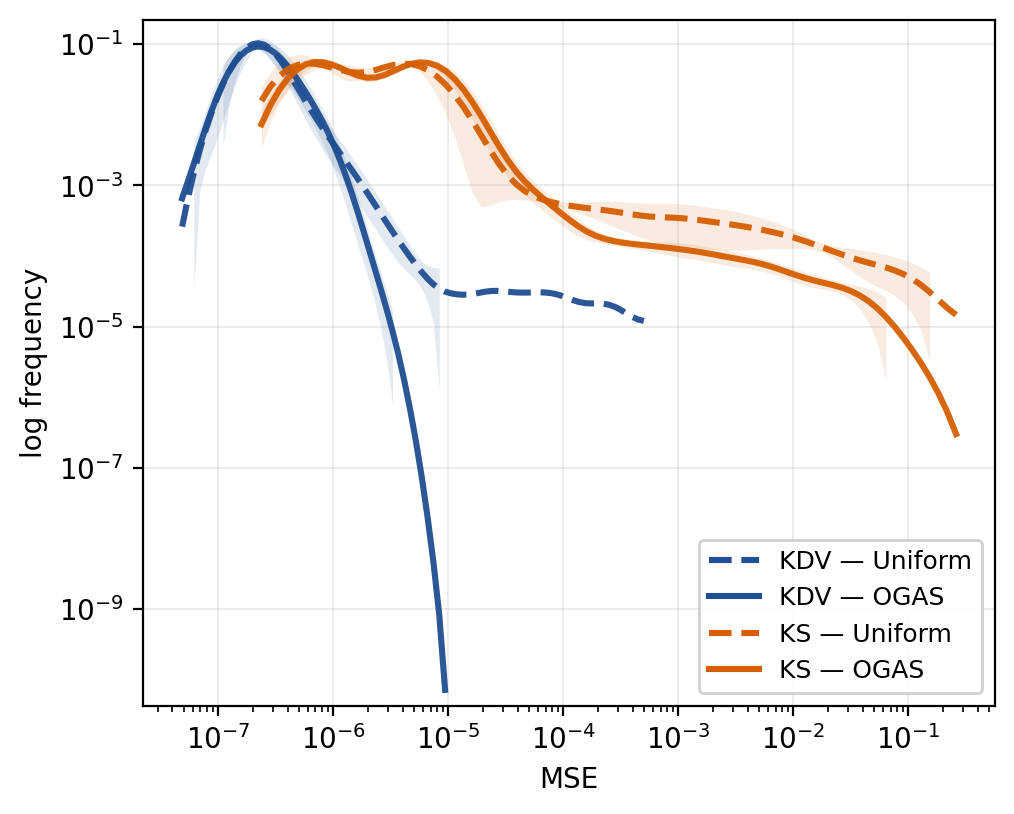}
  \caption{Loss distributions over seeds on 1D KdV and KS\label{fig:rmse_distributions_1d}}
\end{wrapfigure}

\textbf{1D PDEs.} On 1D PDEs (KS, conservative KS, and dispersion-diffusion Korteweg--de Vries (KdV)), the loss-based \textit{OGAS} variant (slightly different from \textit{OGAS-Loss}, see Appendix~\ref{app:1d_exp}) consistently improves all metrics across KS variants, with up to 4$\times$ and 2$\times$ average error reduction over \textit{Uniform} and \textit{Breed}. On KdV, \textit{OGAS} improves worst-case error and dispersion by an order of magnitude, while \textit{Breed} retains the best average error (Table~\ref{tab:all_pdes_one_table_boldbest}). In these simpler regimes (fewer than 10 parameters, smaller models without conditioning, 1D trajectories), \textit{OGAS} steering is especially effective.

\textbf{Ablations.} We probe three knobs of the loss-based \textit{OGAS} on 1D KS (Appendix~\ref{app:1d_exp}). Bias correction reduces error dispersion by 1.26$\times$, average by 1.22$\times$, and worst-case by 1.2$\times$ compared to no correction (Table~\ref{tab:all_pdes_one_table_boldbest}). For the reservoir $\sR$ and generation size, smaller reservoirs with more frequent resampling outperform larger configurations: a $10\times$ smaller setting reduces worst-case error by 1.82$\times$, dispersion by 2.4$\times$, and average error by 2.53$\times$ over the worst configuration (Table~\ref{tab:1d_buffer_ablation}), highlighting the value of rapid adaptation to evolving target conditionals. Finally, more aggressive mixture rates $\alpha$ improve worst-case error at the cost of stability and average performance (Table~\ref{tab:1d_tail_ablation}).

\textbf{Loss distribution.} Figure~\ref{fig:rmse_distributions_1d} shows the full loss distributions on 1D KdV and KS for loss-based \textit{OGAS} and \textit{Uniform}. The two methods nearly overlap on the central bulk of the distribution, yet the upper tails diverge by orders of magnitude in density: \textit{OGAS} reallocates simulation budget away from already-easy regimes into the rare hard ones, suppressing extreme errors without inflating the average, while \textit{Uniform} only improves the left tail with a long high-error tail remaining. 

\section{Limitations}
\label{sec:limitations}

All experiments are performed with online training, bringing extra hyperparameters and nondeterminism due to asynchronous and parallel computations. We could have conversely adapted and tested \textit{OGAS} for offline pool-based training, but that would have dampened \textit{OGAS}, which is designed on purpose for continuous updates and query-synthesis.

\textit{OGAS} is validated on a uniformly sampled dataset, the common practice in the field,  which is also \textit{Uniform}'s training distribution. On this dataset, \textit{OGAS} improves tail metrics while the mean is often unchanged or slightly degraded, so we cannot conclude that \textit{OGAS} reduces training time.  On a validation set sampled from the \textit{OGAS} generator, i.e., close to the training distribution of \textit{OGAS}, it significantly improves both mean and tail errors compared to \textit{Uniform} (Appendix~\ref{app:hard_ddpm_eval}). While such a dataset likely better samples difficult regimes, this validation is nonetheless biased toward  the  distribution that the generator managed to learn. This raises the question of designing a robust and fair evaluation of query-synthesis AL methods when no domain-informed and training-independent dataset or distribution is available.

Scaling is also important, as training surrogates is beneficial when the resulting model is several orders of magnitude faster than the original solver. \textit{OGAS} was not tested in that context to maintain an accessible compute budget. Scaling will likely require hyperparameters different from those validated in this paper, for instance, the number of concurrent solver instances.

PDE foundation models are a good target for online AL training at scale~\cite{ashton-FluidIntelligenceForward-2025,liu-TadpoleAutoencodersFoundation-2026}. Beyond scale, AL must handle heterogeneous solvers (different run times), parameter spaces (different PDEs, boundary types, IC dimensionalities), and cross-PDE signal normalization. This is left for future work.

Our study lacks a theoretical validation. Existing theory covers linear operators and offline training~\citep{subediBenefitsActiveData2025}; online training of nonlinear PDE surrogates remains an open problem.

\section{Conclusion}\label{sec:conclusion}

We introduced \textit{OGAS}, a query-synthesis AL method that learns a signal-conditional distribution over continuous solver parameters using a diffusion generator trained in parallel with the surrogate on the solver parameters paired with their surrogate difficulty signal. Compared with prior works~\citep{dymchenko-SC2024,musekamp-ActiveLearningNeural-2025,kim-ActiveLearningSelective-2025}, \textit{OGAS} avoids finite candidate pools and operates directly in the continuous parameter space. 
Scaling AL studies to higher-complexity PDEs and larger surrogates is essential for practical applicability. We advance beyond these works by targeting 2D PDEs at $256{\times}256$ resolution with up to $308$ parameters and a 184M-parameter transformer surrogate.
 Consistent tail improvements confirm effective steering toward difficult PDE regimes with negligible wall-time overhead in our asynchronous implementation.


\section*{Acknowledgements}
 	
 This work was supported by project Exa-DoST, NumPEx PEPR program, France 2030 state grant reference ANR-22-EXNU-0004.
 This project has received funding from the European Union’s Horizon  research and innovation program under grant agreement No 101144014 (EoCoe-III).
  This work was granted access to the HPC resources of IDRIS under the allocations AD010610366R4
   attributed by GENCI (Grand Equipement National de Calcul Intensif).   We acknowledge the EuroHPC Joint Undertaking for awarding this project access to the EuroHPC supercomputer LEONARDO,
   hosted by CINECA (Italy) and the LEONARDO consortium through an EuroHPC Development Access call.
   We benefited from access to the Grid'5000 testbed, supported by a scientific interest group hosted by Inria and including CNRS,
  RENATER and several Universities as well as other organizations.

\clearpage
\bibliography{main}
\bibliographystyle{plainnat}

\clearpage
\appendix
\onecolumn
\etocdepthtag.toc{mtappendix}
\etocsettagdepth{mtmain}{none}
\etocsettagdepth{mtappendix}{subsection}
\renewcommand{\contentsname}{Appendix Contents}
{\small\tableofcontents}
\clearpage
\section{History Bias Correction}
\label{app:theory}

This appendix complements Sec.~\ref{sec:method} with an analysis of (i) the distribution of logged pairs
$(\lambda,\teps)$ used to train the generator, (ii) the origin of history bias induced by
conditional resampling from a finite buffer $\mathcal{B}$, and (iii) the importance-weighted objective that enforces a uniform
reference prior over $\Lambda$.

We follow the notation of Sec.~\ref{sec:method}: $p_g(\lambda)$ is the \textit{behavior policy} used to launch solver runs at generation $g$ (Eq.~\eqref{eq:mixture_policy}), and $p_{\mathcal{B}}(\lambda,\teps)$ the distribution of the pairs currently stored in the buffer, with marginals $p_{\mathcal{B}}(\lambda)$, $p_{\mathcal{B}}(\teps)$ and conditionals $p_{\mathcal{B}}(\lambda\mid\teps)$, $p_{\mathcal{B}}(\teps\mid\lambda)$. Because $p_g$ changes over generations and $\mathcal{B}$ has finite capacity, $p_{\mathcal{B}}(\lambda)$ is generally non-uniform and captures the sampling history.

\subsection{Assumptions}\label{app:theory_assumptions}

\begin{assumption}[Near-uniform timestep mixing]\label{ass:uniform_timestep}
The timestep distribution induced by sampling segments uniformly from the surrogate buffer $\sR$ satisfies
$q_g(t)\approx \frac{1}{\Nt}$ for all $t\in\{0,\ldots,\Nt-1\}$.
\end{assumption}

\begin{assumption}[Generator optimality on $\mathcal{B}$]\label{ass:generator_optimal_buffer}
Within a generator-update window at generation $g$, maximum-likelihood training on $\mathcal{B}$ recovers the buffer conditional:
\[
p_{\phi_g}(\lambda\mid\tilde\eps)\approx p_{\mathcal{B}}(\lambda\mid\tilde\eps).
\]
\end{assumption}

Assumption~\ref{ass:uniform_timestep} concerns how timesteps enter the logged signal through the surrogate-training sampler.
Assumption~\ref{ass:generator_optimal_buffer} formalizes the idealized limit in which MLE training on $\mathcal{B}$ recovers
the conditional distribution present in the buffer.

\subsection{Logged Distribution and Timestep Marginalization}
\label{app:theory_logged}

Although the buffer stores only pairs $(\lambda,\tilde\eps)$, each $\tilde\eps$ is computed at some (latent) timestep $t$ during
surrogate updates. Let $q_g(\tilde\eps\mid\lambda,t)$ denote the signal distribution induced by the surrogate at generation $g$
when the signal is evaluated at timestep $t$ and filling the buffer $\mathcal{B}$. Under Assumption~\ref{ass:uniform_timestep}, the timestep-marginal (time-mixed)
signal conditional in the buffer is
\begin{equation}
p_{\mathcal{B}}(\tilde\eps\mid\lambda)
\;\approx\;
\sum_{t=0}^{\Nt-1} q_g(t)\,q_g(\tilde\eps\mid\lambda,t)
\;\approx\;\frac{1}{\Nt}\sum_{t=0}^{\Nt-1} q_g(\tilde\eps\mid\lambda,t).
\label{eq:app_time_marginal}
\end{equation}
Time-local sampling from the finite buffer can therefore be represented as
\begin{equation}
(\lambda,\tilde\eps)\sim p_{\mathcal{B}}(\lambda,\tilde\eps)
= p_{\mathcal{B}}(\lambda)\,p_{\mathcal{B}}(\tilde\eps\mid\lambda),
\label{eq:app_logged_joint_buffer}
\end{equation}
with induced signal marginal
\begin{equation}
p_{\mathcal{B}}(\tilde\eps)=\int_\Lambda p_{\mathcal{B}}(\lambda)\,p_{\mathcal{B}}(\tilde\eps\mid \lambda)\,d\lambda,
\label{eq:app_signal_marginal_buffer}
\end{equation}
and Bayes' rule giving the buffer conditional
\begin{equation}
p_{\mathcal{B}}(\lambda\mid\tilde\eps)=\frac{p_{\mathcal{B}}(\lambda)\,p_{\mathcal{B}}(\tilde\eps\mid \lambda)}{p_{\mathcal{B}}(\tilde\eps)}.
\label{eq:app_bayes_post_buffer}
\end{equation}

\subsection{Origin of History Bias}
\label{app:theory_bias}

Recall that the generator-induced proposal at generation $g{+}1$ is obtained by sampling $\tilde\eps\sim k(\tilde\eps)$ and then
$\lambda\sim p_{\phi_g}(\lambda\mid\tilde\eps)$:
\begin{equation}
p^{\mathrm{gen}}_{g+1}(\lambda)
=\int k(\tilde\eps)\,p_{\phi_g}(\lambda\mid\tilde\eps)\,d\tilde\eps.
\label{eq:app_gen_component}
\end{equation}
Under Assumption~\ref{ass:generator_optimal_buffer}, $p_{\phi_g}(\lambda\mid\tilde\eps)\approx p_{\mathcal{B}}(\lambda\mid\tilde\eps)$.
Substituting the Bayes form \eqref{eq:app_bayes_post_buffer} yields the (unnormalized) factorization
\begin{equation}
p^{\mathrm{gen}}_{g+1}(\lambda)
\;\propto\;
p_{\mathcal{B}}(\lambda)\int k(\tilde\eps)\,\frac{p_{\mathcal{B}}(\tilde\eps\mid \lambda)}{p_{\mathcal{B}}(\tilde\eps)}\,d\tilde\eps.
\label{eq:app_bias_factor_buffer}
\end{equation}

\paragraph{Interpretation and link to the behavior policies.}
Eq.~\eqref{eq:app_bias_factor_buffer} makes explicit a multiplicative \textit{history} term $p_{\mathcal{B}}(\lambda)$ induced by the
finite buffer. Since $\mathcal{B}$ stores pairs generated by solver runs launched under the sequence of behavior policies
$\{p_{g'}(\lambda)\}_{g'\le g}$, the buffer marginal $p_{\mathcal{B}}(\lambda)$ can be viewed as a time-local mixture of these policies,
weighted by how much data from each generation remains in the surrogate buffer.
Consequently, naive conditional resampling can inherit over-represented regions of $\Lambda$ through $p_{\mathcal{B}}(\lambda)$
(\textit{history bias}), even when the current surrogate signal favors other regions via the tilt term
$p_{\mathcal{B}}(\tilde\eps\mid\lambda)$.

\subsection{Uniform-prior Generator Training via Importance Weighting}
\label{app:theory_uniform_obj}

The objective in Sec.~\ref{sec:history_bias_correction} corresponds to training the generator under the target joint
\begin{equation}
p_{\mathcal{B}}^\star(\lambda,\tilde\eps)
=\mathcal{U}_\Lambda(\lambda)\,p_{\mathcal{B}}(\tilde\eps\mid\lambda),
\label{eq:app_target_joint}
\end{equation}
which enforces a uniform reference prior on $\Lambda$ while preserving the (time-marginal) signal mechanism encoded by
$p_{\mathcal{B}}(\tilde\eps\mid\lambda)$.
Since $\mathcal{B}$ provides samples from \eqref{eq:app_logged_joint_buffer}, importance sampling yields
\begin{equation}
\mathbb{E}_{(\lambda,\tilde\eps)\sim p_{\mathcal{B}}^\star}\!\big[L_{\mathrm{gen}}(\lambda,\tilde\eps;\phi)\big]
=
\mathbb{E}_{(\lambda,\tilde\eps)\sim p_{\mathcal{B}}}\!\left[
\frac{\mathcal{U}_\Lambda(\lambda)}{p_{\mathcal{B}}(\lambda)}\,L_{\mathrm{gen}}(\lambda,\tilde\eps;\phi)\right],
\label{eq:app_uniform_is}
\end{equation}
justifying the density ratio $w_{\mathcal{B}}(\lambda)=\mathcal{U}_\Lambda(\lambda)/p_{\mathcal{B}}(\lambda)$ used in the main text.
This is the standard covariate-shift correction \cite{bickel_discriminative_2009}.

\subsection{Classifier-based Density-ratio Estimation}
\label{app:theory_classifier}

Eq.~\eqref{eq:app_uniform_is} requires the marginal ratio $\mathcal{U}_\Lambda(\lambda)/p_{\mathcal{B}}(\lambda)$, but
$p_{\mathcal{B}}(\lambda)$ is unknown. We estimate this ratio with a probabilistic classifier.
We assign label $y{=}0$ to uniform reference samples $\lambda\sim\mathcal{U}_\Lambda$ and label $y{=}1$ to buffer samples obtained
by drawing $(\lambda,\tilde\eps)\sim p_{\mathcal{B}}$ and retaining $\lambda$.
We train a discriminator $D_\psi(\lambda)\in(0,1)$ to predict
$D_\psi(\lambda)=P_\psi(y{=}0\mid\lambda)$.
With class priors $\pi_0=P(y{=}0)$ and $\pi_1=P(y{=}1)$ estimated from the discriminator minibatch, Bayes' rule yields
\cite{goodfellow_generative_2014}
\begin{equation}
\widehat w_{\mathcal{B}}(\lambda)
=\frac{\pi_1}{\pi_0}\cdot\frac{D_\psi(\lambda)}{1-D_\psi(\lambda)}.
\label{eq:app_ratio_classifier_weight}
\end{equation}
In practice, we use balanced minibatches with an equal number of uniform and buffer samples, so the class-prior factor cancels.
We clip $\widehat w_{\mathcal{B}}(\lambda)$ to $[w_{\min},w_{\max}]$ for numerical stability.

\begin{remark}[Alternative: uniform-only generator updates]
\label{rem:uniform_only}
A simpler alternative is to train the generator only on pairs $(\lambda,\tilde\eps)$ originating from the uniform branch of the
behavior policy, removing the need for ratio estimation since the effective base measure is already uniform.
However, when the uniform fraction $\alpha$ is small, this approach reduces the number of available training pairs and we
found empirically that it adapts more slowly than ratio-weighted updates.
\end{remark}

\section{OGAS Implementation Details}
\label{app:al_training}

\subsection{Online vs.\ Offline AL}
\label{app:online_vs_offline_AL}

Traditional offline AL~\cite{musekamp-ActiveLearningNeural-2025} operates in sequential rounds: a batch of parameter candidates is selected, numerical solvers produce samples to populate a static dataset stored on disk, and the surrogate model is then trained on this dataset. In contrast, our framework adopts a fully online streaming paradigm where data generation, memory management, surrogate gradient updates, and AL occur concurrently within a unified asynchronous workflow (Fig.~\ref{fig:al-workflow}). While offline AL updates its sampling distribution sequentially with coarse steps, online AL updates the conditional sampling policy $p_g(\vla)$ continuously as new surrogate training signals are observed. This keeps the generator's inferred parameters up to date, redirecting solver work toward high-error parameter without waiting for the next round. 

A fully online workflow also brings additional benefits. Running solver instances and training in parallel on distinct resources enables efficient use of these resources with minimal idle time or switching overheads. Streaming trajectory states through a finite reservoir buffer $\sR$ (Fig.~\ref{fig:al-workflow}) bounds memory consumption regardless of the total dataset volume, and avoids I/O and disk storage entirely.

\subsection{Conditional Diffusion Generator}
\label{app:diffusion}

\paragraph{Choice of generative model.}
We parametrize the conditional generator $p_\phi(\vla\mid\teps)$ with a denoising diffusion probabilistic model (DDPM) \cite{ho_denoising_2020} equipped with classifier-free guidance (CFG) \cite{ho_classifier-free_2022}. In preliminary experiments, simpler CVAE variants were less stable and suffered from posterior collapse, whereas DDPMs gave a more reliable fit to the continuous conditional distributions encountered here and naturally supported guided sampling toward high-signal regions and scaled well with IC dimensionality. For discrete or mixed parameter spaces, the same framework could be paired with a mixed-type generative model instead of a Gaussian DDPM. The training and sampling pipelines follow the standard DDPM recipe; here we only specify the components that differ from a vanilla setup, namely (a) density-ratio reweighting and (b) the concrete denoiser/discriminator architectures and hyperparameters used in our asynchronous update loop.

\paragraph{Weighted diffusion objective.}
Let $q(x_k\mid \vla)$ denote the forward noising process at diffusion step $k\in\{1,\dots,K_{\mathrm{diff}}\}$ (as in \cite{ho_denoising_2020}), producing the noisy input $x_k$. The denoiser $\varepsilon_\phi(x_k,k,\teps)$ predicts the additive Gaussian noise $\varepsilon$ used to construct $x_k$. Under the density-ratio debiasing described in Sec.~\ref{app:theory_classifier}, we associate each training pair $(\vla,\teps)$ with an importance weight $w_g(\vla)$ (Eq.~\eqref{eq:app_ratio_classifier_weight}). The resulting weighted noise-prediction objective is
\begin{equation}
\mathcal{L}^{\mathrm{w}}_{\mathrm{diff}}(\phi)
=
\mathbb{E}_{(\vla,\teps)\sim \sB}
\mathbb{E}_{k\sim \mathrm{Unif}(\{1,\dots,K_{\mathrm{diff}}\}),\,\varepsilon\sim\mathcal{N}(0,I)}\Big[
w_g(\vla)\,\|\varepsilon-\varepsilon_\phi(x_k,k,\teps)\|_2^2
\Big].
\end{equation}
In all experiments, we clamp $w_g(\vla)$ to a bounded interval (Table~\ref{tab:diffusion-hparams}) to prevent unstable gradients.

\paragraph{Classifier-free guidance.}
We use standard classifier-free guidance~\cite{ho_classifier-free_2022}. During training, the conditioning signal $\teps$
is replaced by a learned null token with probability $p_{\text{drop}}=0.1$; during sampling, we use guidance scale $w=5$.
The standard conditional/unconditional interpolation is otherwise unchanged; the complete values are listed in
Table~\ref{tab:diffusion-hparams}.

\paragraph{Denoiser architecture.}
Our denoiser is an MLP-based network chosen for fast (asynchronous) updates. Let $d$ be the dimensionality of $\vla$ and let $m=1024$ be the hidden width. The model maps $(x_k,k,\teps)$ to a noise prediction in $\mathbb{R}^d$ using: (i) an input projection to width $m$, (ii) a sinusoidal time embedding, (iii) a learned condition embedding for $\teps$, and (iv) two residual MLP blocks modulated by the fused embeddings. Concretely:

\begin{enumerate}
  \item \textbf{Input projection.} Project $x_k\in\mathbb{R}^d$ to $h_0\in\mathbb{R}^m$ with a linear layer.
  \item \textbf{Embeddings.} Compute a sinusoidal embedding of the diffusion step $k$, map it to $\mathbb{R}^m$, then project to $e=4m=4096$. Independently embed the condition $\teps$ to $\mathbb{R}^{4096}$ (and embed the null token $\teps_\varnothing$ similarly). Sum the time and condition embeddings to obtain a fused embedding in $\mathbb{R}^{4096}$.
  \item \textbf{Residual denoising trunk.} Apply two identical residual MLP blocks, each of the form
  \[
  h \leftarrow h + \mathrm{Linear}\big(\mathrm{SiLU}(\mathrm{LayerNorm}(h))\big),
  \]
  where the fused embedding modulates each block via affine modulation (FiLM-style): per block we produce $(\gamma,\beta)\in\mathbb{R}^m\times\mathbb{R}^m$ from the fused embedding and apply
  \[
  \mathrm{LayerNorm}(h)\mapsto \gamma\odot \mathrm{LayerNorm}(h) + \beta.
  \]
  \item \textbf{Output head.} A final LayerNorm--Linear head maps the resulting hidden state back to $\mathbb{R}^d$ to output $\varepsilon_\phi(x_k,k,\teps)$.
\end{enumerate}

\paragraph{Density-ratio classifier.}
To compute $w_g(\vla)$ (Eq.~\eqref{eq:app_ratio_classifier_weight}), we train a lightweight discriminator $D_\psi$ to distinguish uniform reference samples $\vla\sim\mathcal{U}_\Lambda$ from buffer samples (Appendix~\ref{app:theory_classifier}). It is a 2-layer MLP,
\[
\text{Linear}(d\!\to\!64)\ \to\ \text{ReLU}\ \to\ \text{Linear}(64\!\to\!1),
\]
trained with AdamW on the binary cross-entropy plus a quadratic penalty on its logits $s_\psi(\vla)$:
\[
\mathcal{L}(\psi) \;=\; \mathcal{L}_{\mathrm{BCE}}(\psi) \;+\; \beta\,\mathbb{E}_{\vla}\!\left[s_\psi(\vla)^2\right],
\]
where the expectation runs over both uniform and buffer samples. We found that in high-dimensional parameter spaces, generated samples quickly become almost perfectly separable from uniform ones so that an unregularized classifier then drives its logits to $\pm\infty$, the importance weights of bred samples collapse to zero and the generator is effectively trained on uniform samples only. The penalty then acts as a zero-mean Gaussian prior on the log-ratio: it shrinks the estimated log density ratio towards $0$ where the classifier is over-confident, which keeps the correction smooth and the effective sample size of the weighted buffer high, while preserving the ranking of samples by their relative density.

\paragraph{Configuration.}
Table~\ref{tab:diffusion-hparams} summarizes the full configuration used for the diffusion generator and the density-ratio classifier.

\begin{table}[h]
\centering
\caption{Diffusion generator and density-ratio classifier configuration.}
\label{tab:diffusion-hparams}
\begin{tabular}{ll}
\toprule
\textbf{Component} & \textbf{Setting} \\
\midrule
\multicolumn{2}{l}{\textit{Diffusion model (DDPM + CFG)}} \\
Hidden width ($m$) & 1024 \\
Residual blocks & 2 (LayerNorm/SiLU/Linear with residual connection) \\
Time embedding & sinusoidal $\to \mathbb{R}^{m}$, projected to $e=4096$ \\
Condition embedding & $\teps \mapsto \mathbb{R}^{4096}$;\\
Modulation & FiLM-style modulation of each block by the fused embedding \\
Optimizer / LR & Adam, $2\times10^{-4}$ \\
Batch size & 128 \\
Update budget & $\le 50$ DDPM gradient steps per new data arrival \\
Noise schedule & linear $\beta_k$ from $10^{-4}$ to $0.02$ over $K_{\mathrm{diff}}=100$ steps \\
CFG dropout prob. ($p_{\text{drop}}$) & 0.1  \\
Guidance scale ($w$) & 5.0 \\
\midrule
\multicolumn{2}{l}{\textit{Density-ratio classifier (for $w_g(\vla)$)}} \\
Architecture & Linear($d\to64$)$\to$ReLU$\to$Linear($64\to1$) \\
Optimizer / LR & AdamW, $10^{-4}$, weight decay $10^{-2}$ \\
Logit penalty ($\beta$) & 0.03 \\
Log-weight clipping & $\log w_g(\vla)\in[-5,\,5]$, then $w_g(\vla)\le 1/\pi_{\mathcal{U}}$ 
\\
Normalization & unit mean over the batch \\
\bottomrule
\end{tabular}
\end{table}

\subsection{Online Training Loop and Stability Heuristics}
\label{app:ogas_implementation}\label{app:impl_stability}
`
Our setting is fully online: solver trajectories, surrogate updates, and generator updates run concurrently and exchange data
through finite-capacity buffers. Without additional safeguards, this pipeline can become unstable due to (i) stale
conditioning signals, (ii) generator overfitting when simulations lag, and (iii) changes in the effective number of surrogate
updates per simulated trajectory. Below we summarize the practical heuristics used to ensure stable and reproducible runs.

\paragraph{Decoupled asynchronous pipeline.}
We keep the surrogate training loop on the critical path and run active sampling in the background:
(i) training-signal computation and transmission are asynchronous, and (ii) generator (DDPM) updates never block surrogate
update. This prevents variations in generator load from affecting surrogate throughput and maximize GPU-ressources use.

\paragraph{Signal normalization.}
The raw acquisition values vary over training as the surrogate improves. We therefore maintain running statistics for
standardization using an exponential moving average with decay $\gamma=0.99$ (roughly a ${\sim}100$-batch horizon). Each
incoming minibatch updates $(\mu,\sigma)$, and we store the normalized signal $\tilde\eps=(\eps-\mu)/\sigma$ in the
buffer $\sB$.

\paragraph{Generator buffer}
We train the generator on a dedicated buffer $\sB$ that stores only recent $(\lambda,\tilde\eps)$ pairs; the timestep is
marginalized when the signal is logged.
Because the surrogate evolves throughout training, older signals can become miscalibrated; restricting $\sB$ to
recent samples reduces staleness and improves the match between the generator's conditioning variable and the current loss
landscape.

\paragraph{Generator update schedule.}
The DDPM denoiser is made faster to train than the surrogate, enabling frequent refreshes of the conditional sampler
without becoming a bottleneck. Generator updates are driven by data availability rather than wall-time: when new
$(\lambda,\teps)$ pairs arrive, we run at most $K_{\max}$ DDPM gradient steps (i.e., per surrogate update) then stop until the next surrogate update.
This prevents the generator from overfitting a small, rapidly changing $\sB$, or a stale one when simulations are slow, and
makes generator compute scale with the rate at which fresh $(\lambda,\tilde\eps)$ pairs are produced. We use $K_{\max}=50$ in 2D.

\paragraph{Mixture warmup.}
We use a linear warmup for the generator fraction $\varepsilon=1-\alpha$, increasing it from $0.0$ to $0.7$ early in
training. This ensures the surrogate first observes broad coverage of $\Lambda$ under uniform sampling, building a stable global approximation before the generator concentrates  compute on high-signal regions.

\paragraph{Speed limiter.}
\label{app:speed_limiter}
In an asynchronous client--server setup, cluster variability can change the relative speeds of simulation and training.
This changes the \textit{effective} number of surrogate updates per simulated trajectory and can lead to irreproducible
results even when random seeds are fixed.

We monitor two rates: the \textit{input flux} (new samples inserted into $\sR$ per second) and the \textit{output flux}
(minibatches consumed for surrogate update per second). If input flux dominates, successive batches contain mostly unseen samples,
yielding high variance updates; if output flux dominates, the model repeatedly trains on the same buffer content,
increasing overfitting risk. The problematic case is drift over time (e.g., due to node preemption or stragglers), which
changes this ratio mid-run.

We control this with a \textit{view ratio}, defined as the average number of batches performed on the surrogate per simulated trajectory
(or equivalently, the fraction of $\sR$ samples that have been ``seen'' by training). The speed limiter enforces a
target view ratio by throttling either simulation intake or surrogate execution so that the data/compute ratio remains stable
throughout training. For our 2D experiments, we set the target to $1.8$ surrogate batches per simulation, matching the
available cluster throughput and yielding consistent training dynamics across runs. 

\section{Experimental Setup}
\label{app:impl}

This appendix specifies the 2D data generation (PDEs, solver configurations, initial conditions), the surrogate architectures, the training protocol, and the baselines.

\subsection{PDEs and Solver Configurations}
\label{app:pde_solver_setup}\label{app:pde_solver_config}

\begin{figure*}[h]
  \centering
  \includegraphics[width=0.8\textwidth]{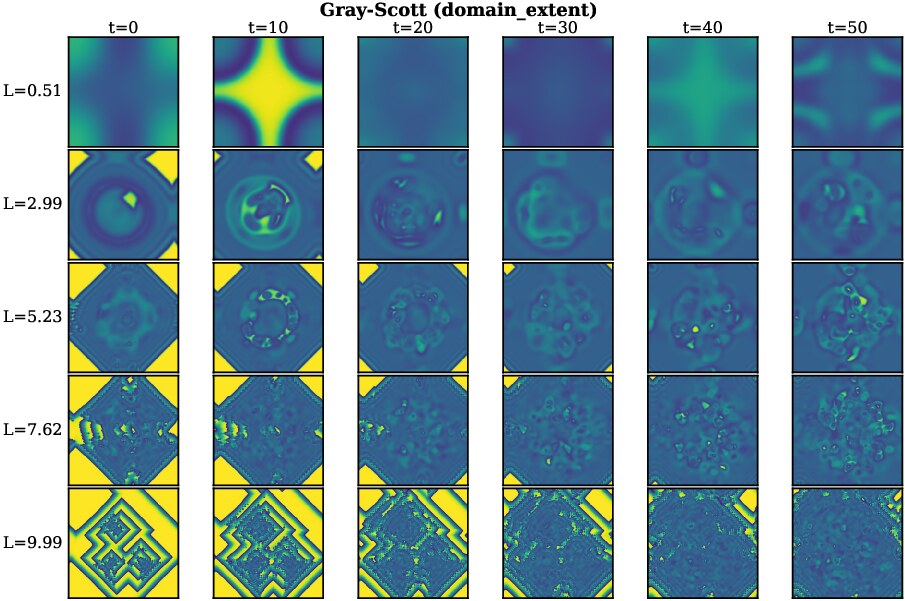}
  \Description{Grid of Gray--Scott concentration-field snapshots from several two-dimensional validation trajectories and time steps, illustrating reaction--diffusion pattern evolution.}
  \caption{\textit{Gray-Scott trajectory grid.} Visual snapshots from the 2D validation suite: Gray-Scott (reaction-diffusion patterns).}
  \label{fig:pde_gray_scott}
\end{figure*}

\begin{figure*}[h]
  \centering
  \includegraphics[width=0.8\textwidth]{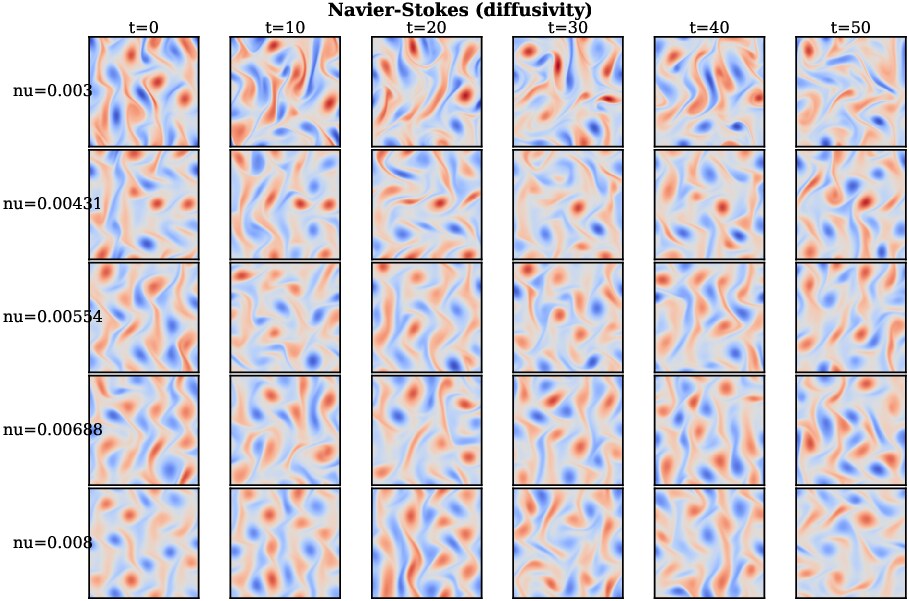}
  \Description{Grid of Navier--Stokes Kolmogorov-flow vorticity snapshots from several validation trajectories and time steps, illustrating fluid-dynamics evolution.}
  \caption{\textit{Navier-Stokes trajectory grid.} Visual snapshots from the 2D validation suite: Navier-Stokes Kolmogorov Flow (fluid dynamics).}
  \label{fig:pde_navier_stokes}
\end{figure*}

\begin{figure*}[h]
  \centering
  \includegraphics[width=0.8\textwidth]{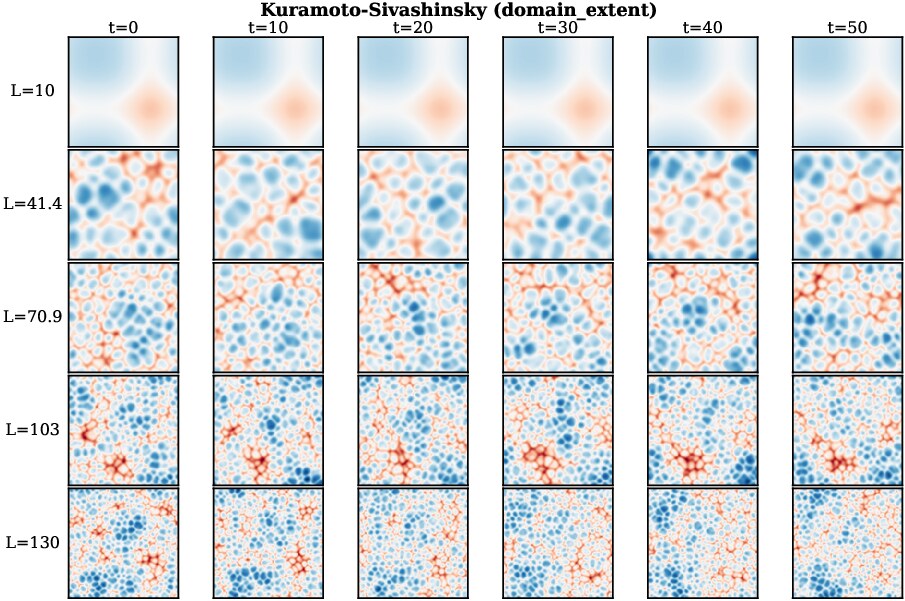}
  \Description{Grid of Kuramoto--Sivashinsky scalar-field snapshots from several two-dimensional validation trajectories and time steps, illustrating chaotic spatiotemporal evolution.}
  \caption{\textit{Kuramoto-Sivashinsky trajectory grid.} Visual snapshots from the 2D validation suite: Kuramoto-Sivashinsky (chaotic). }
  \label{fig:pde_ks}
\end{figure*}

\paragraph{Choice of PDEs and solver.}
PDEs used for the 2D evaluations were chosen to ensure a wide range of different PDEs, at high spatial resolutions, and with
varying physical quantities across simulations. To simulate them and produce the synthetic data, we relied on the Exponax Solver used in APEBench which implements a range of Exponential Time Differencing Runge-Kutta
methods operate
in Fourier space and as such do not allow for non-periodic domains or complex boundaries. To follow the design of the dataset of PDE transformer \cite{koupai-EfficientGenerativeTransformer-2025}, we use Exponax (as in APEBench) with our own data-generation interface/config.  We always use the physical solver interface across datasets to provide a simpler, unified way of handling physical quantities. Representative trajectories for these three equations are shown in Figure~\ref{fig:pde_gray_scott}, Figure~\ref{fig:pde_navier_stokes}, and Figure~\ref{fig:pde_ks}.
We provide the stepper configuration when using directly the Exponax solvers.
\paragraph{Common simulation setup (all 2D cases).}
All datasets are generated with Exponax pseudo-spectral steppers in Fourier space on a 2D periodic domain $\Omega=[0,L]^2$. Unless stated otherwise, we use a spectral grid of $N=256$ points per spatial dimension and compute spatial operators spectrally. Each recorded trajectory consists of a fixed number of \textit{stored} time steps: $t_s=32$ for training (training on two steps rollout following AL4PDE implementation) so as to yield 30 transitions per trajectory. The validation trajectories are longer and use $t_s^{\mathrm{val}}=51$ as validation is performed in one-step fashion so yielding here $50$ transitions. The validation set comprises 750 trajectories sampled via Halton sequences (total 37,500 samples), consistent with the main text. We distinguish three time-integration notions: (i) the \textit{stored} step size $\Delta t$ between consecutive saved states, (ii) the number of \textit{substeps} $n_{\mathrm{sub}}$ taken internally per stored step, and (iii) the corresponding internal integrator step $\delta t=\Delta t/n_{\mathrm{sub}}$. To reduce dependence on transients, we run an initial \textit{warmup} phase of a prescribed number of internal steps that are discarded before any states are recorded.

\paragraph{Kuramoto--Sivashinsky (2D).}
We simulate a single scalar field (density) with the Exponax Kuramoto--Sivashinsky stepper and periodic boundaries on $\Omega=[0,L]^2$. The domain extent is sampled as $L\sim\mathcal{U}[10,130]$. We discard a warmup of 200 internal steps prior to recording. The time delta used is $\Delta t=0.5$ and $n_{\mathrm{sub}}=5$ internal substeps, yielding a solver $\delta t=0.1$.

\paragraph{Gray--Scott reaction--diffusion (2D).}
We use the Exponax Gray--Scott stepper for two coupled concentration fields with periodic boundaries on $\Omega=[0,L]^2$. 2 Diffusivity and reaction parameters are fixed to
$D_a=2\times 10^{-5}$, $D_b=10^{-5}$, $F=0.02$, and $K=0.046$ to follow the $\beta$ variant of this equation used in PDETransformer \cite{koupai-EfficientGenerativeTransformer-2025}, while the domain extent is in our case sampled within the range $L\sim\mathcal{U}[0.5,10]$. We use a step size of $\Delta t=30$ with $n_{\mathrm{sub}}=30$ yielding $\delta t=1$ for the solver.

\paragraph{Navier--Stokes (2D Kolmogorov flow, vorticity form).}
We simulate a single vorticity field using the Exponax vorticity-formulation Navier--Stokes stepper with built-in Kolmogorov forcing (single-mode sinusoidal forcing in the vorticity equation) and periodic boundaries. The domain extent is fixed to $L=2\pi$, while viscosity is sampled as $\nu\sim\mathcal{U}[0.003,0.008]$. We discard a warmup of 50 internal steps prior to recording. We use a step size of $\Delta t=0.2$ and $n_{\mathrm{sub}}=100$, giving $\delta t=0.002$ for the solver.


\subsection{Initial Condition Families}
\label{app:ic_families} 

We use two families of random initial conditions, each designed to produce diverse spatial structures while remaining smooth enough to avoid numerical instabilities.

\begin{figure*}[h]
  \centering
  \includegraphics[width=0.8\textwidth]{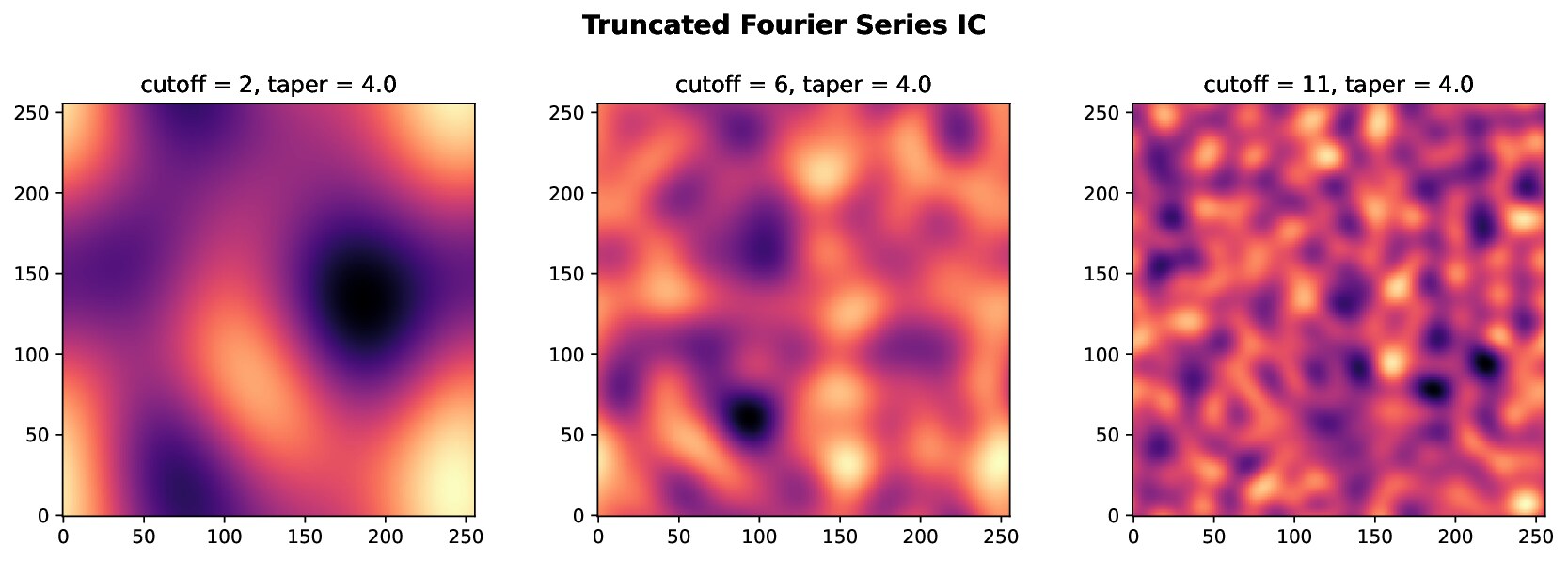}
  \Description{Examples of truncated Fourier-series initial conditions with different spectral cutoffs and taper powers.}
  \caption{Example of truncated fourier IC sampling with various cutoff and taper power}
  \label{fig:ic_fourier}
\end{figure*}

\paragraph{Truncated Fourier series.}
For the Kuramoto--Sivashinsky and Navier--Stokes equations, we sample initial conditions from a truncated Fourier series. On a square periodic domain $\Omega=[0,L]^2$ discretized on an $N\times N$ grid, we first draw a spectral cutoff $K\sim\mathcal{U}[2,8]$ and a taper exponent $p\sim\mathcal{U}[1,10]$. For each scalar field channel $c$ and each retained wavevector $\vk=(k_x,k_y)$ with $k_x\in\{0,\ldots,K\}$ and $k_y\in\{-K,\ldots,K\}$, we sample an amplitude $a_{c,\vk}\sim\mathcal{U}[-1,1]$ and phase $\phi_{c,\vk}\sim\mathcal{U}[0,2\pi]$. The initial field is then
\[
 u_c(\vx)\;=\;\sum_{\vk} a_{c,\vk}\,\exp\!\left(-\left(\frac{\|\vk\|}{K}\right)^{p}\right)\cos\!\left(\tfrac{2\pi}{L}\,\vk\cdot \vx+\phi_{c,\vk}\right).
\]
The exponential taper suppresses high-frequency modes smoothly, with stronger suppression for larger $p$. After sampling, each channel is normalized to have maximum absolute value 1. This parameterization covers a wide range of spatial scales---from nearly uniform fields (small $K$) to highly oscillatory patterns (large $K$)---while the taper power controls the smoothness of the transition. Examples are shown in Figure \ref{fig:ic_fourier}.

\paragraph{Gaussian blob superposition.}
For the Gray--Scott reaction--diffusion system, we construct initial conditions by superimposing $B=8$ anisotropic Gaussian blobs for the first channel. For each blob $i$ on channel $c$, we sample: an amplitude $A_{c,i}\sim\mathcal{U}[0,1]$; a center $(\mu_{ix},\mu_{iy})$ with each coordinate drawn from $\mathcal{U}[0.2L,0.8L]$; spread parameters $\sigma_{ix},\sigma_{iy}\sim\mathcal{U}[0.07,0.1]$ (relative to domain size); and a correlation coefficient $\rho_{c,i}\sim\mathcal{U}[0,0.1]$. The covariance matrix for blob $i$ is
\[
\Sigma_{c,i}=\begin{bmatrix}(L\sigma_{ix})^2 & \rho_{c,i}(L\sigma_{ix})(L\sigma_{iy})\\[2pt]
\rho_{c,i}(L\sigma_{ix})(L\sigma_{iy}) & (L\sigma_{iy})^2\end{bmatrix}.
\]
The initial field combines all blobs via amplitude-weighted averaging:
\[
 u_c(\vx)=\frac{1}{B}\sum_{i=1}^{B} A_{c,i}\exp\!\left(-\tfrac{1}{2}(\vx-\vmu_{c,i})^\top\Sigma_{c,i}^{-1}(\vx-\vmu_{c,i})\right).
\]
For Gray--Scott, which models two interacting chemical species, we set the second channel as the complement of the first: $u_1 \gets 1-u_0$. This results in 48 sampled parameters for the initial condition. This produces localized concentration patterns that evolve into the characteristic spots and stripes of the Gray--Scott system.

\begin{figure*}[h]
  \centering
  \includegraphics[width=0.8\textwidth]{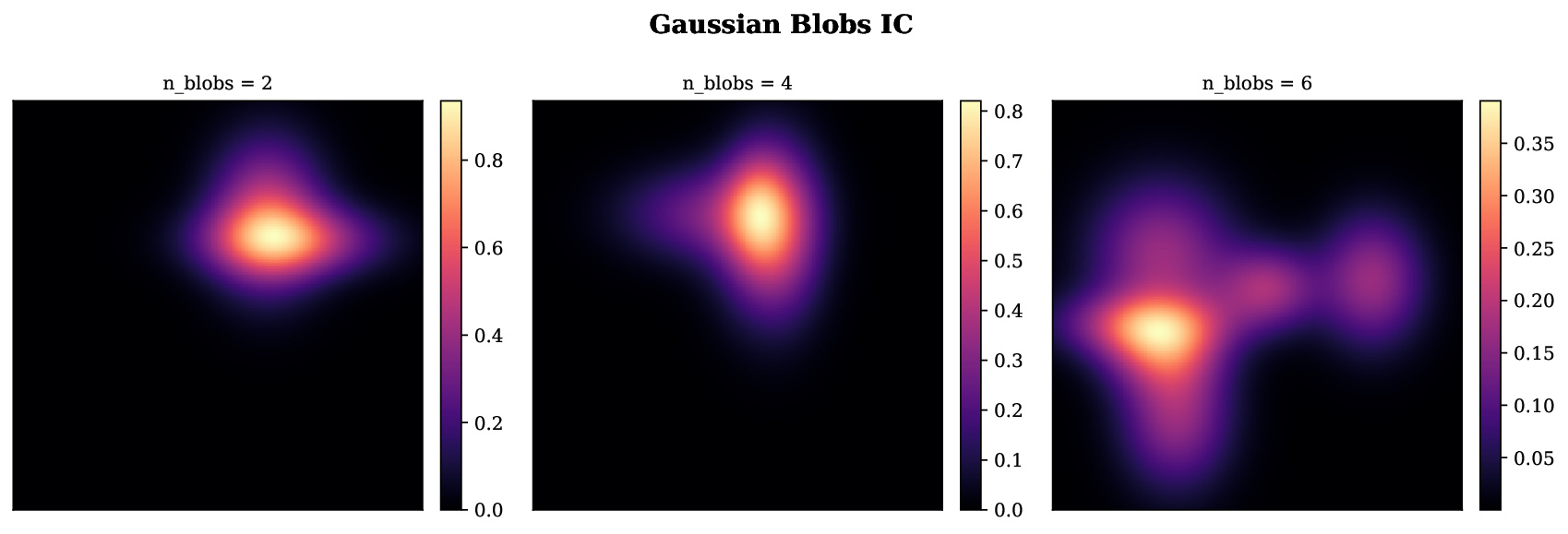}
  \Description{Examples of Gray--Scott initial conditions formed by Gaussian blobs, with different numbers and spatial arrangements of blobs.}
  \caption{ Example of gaussian blobs IC sampling with varying number of blobs}
  \label{fig:ic_blobs}
\end{figure*}

Examples are shown in Figure~\ref{fig:ic_blobs}.
\subsection{Surrogate Architectures}
\label{app:model_architecture}

All surrogate models operate in an autoregressive mode: given the current PDE state $u_t$ and physics parameters $\vla$, the model predicts the next state $u_{t+1}$. Thus when using 2-steps prediction in our setup we simply calculate two transitions applying the model two times and storing intermediaries gradients as it is usually done for a recurrent model. We evaluate three neural operator architectures spanning spectral, convolutional, and attention-based paradigms.

\paragraph{U-Net.}
We use a modern conditional U-Net adapted from AL4PDE~\citep{musekamp-ActiveLearningNeural-2025}. The encoder--decoder structure has 4 resolution levels with channel multipliers $[1, 2, 2, 4]$ relative to a base width of 8 channels, and 2 residual blocks per level. Physics conditioning is injected into each residual block via a learned linear projection (additive conditioning). For periodic domains, we use circular padding. The model predicts a scaled residual: $\hat{u}_{t+1} = u_t + 0.3\,\Delta u$, where the scaling factor improves training stability for stiff dynamics. GELU activations and group normalization are used throughout;

\paragraph{Fourier Neural Operator (FNO).}
The FNO~\citep{li2020fourier} performs learning directly in Fourier space. Our architecture uses 4 spectral convolution layers, each retaining the lowest 60 Fourier modes per spatial dimension. The hidden channel width is 32 throughout, with GELU activations and a final linear projection head (hidden dimension 128). Physics conditioning is achieved by broadcasting each physical coefficient as a new channel and concatenating with the input channels. The model predicts a residual $\Delta u$ that is added to the input state. Spatial coordinates are appended as additional input channels to encode positional information.

\paragraph{Scalable Operator Transformer (ScOT).}
ScOT~\citep{herde-PoseidonEfficientFoundation-2024} is a Swin-style vision transformer adapted for operator learning. While the initial model is provided with several configurations, they were mainly designed for 128x128 grid. In our setup we then chose to adapt the Tiny version of the model, raising the patch size from 4 to 8 and tripling then embedding size. This yield the $T256$ variant that we use in our setup. With the 8-size patch size, it yields a $32\times32$ token grid for $256\times256$ inputs. The architecture has 4 stages with depths $[4, 4, 4, 4]$ and head counts $[3, 6, 12, 24]$, using local attention with window size 8. The embedding dimension is 144 with MLP ratio 4. We also adapt physics conditioning to replace the existing mechanism of timestep conditioning to our autoregressive task. This mechanism is provided via a two-layer MLP that projects the parameter vector to the embedding dimension, which is then added as a learnable token (analogous to a time embedding in diffusion models). Skip connections link encoder and decoder stages, and a ConvNeXt-style residual block refines the output. The model predicts a residual that is added to the input state.

\subsection{Training Protocol}
\label{app:model_training_details}

\paragraph{Shared protocol.}
All models are trained online in the streaming setting. Each training batch contains consecutive state pairs $(u_t, u_{t+1}, u_{t+2})$; we train with a two-step loss as described in Sec.~\ref{sec:preliminaries},
where $\hat{u}_{t+2}$ is obtained by autoregressively applying the model to its own prediction $\hat{u}_{t+1}$. 

We use a batch size of 64 and normalize physics parameters to $[0,1]$ via Min-Max scaling using the bounds defined in the experiment config before being passed to the model. Training uses FP32 precision. Additionally, we apply adaptive gradient clipping following the setup of \citet{musekamp-ActiveLearningNeural-2025}, where the clipping threshold is an exponential moving average of current gradient norms (multiplier 5.0, EMA factor 0.05).

\paragraph{Per-model hyperparameters.}
Table~\ref{tab:training_hp} summarizes the optimizer and scheduler settings for each architecture.

\begin{table}[h]
\centering
\caption{Training hyperparameters by model architecture.}
\label{tab:training_hp}
\small
\begin{tabular}{lccc}
\toprule
& \textit{FNO} & \textit{U-Net} & \textit{ScOT} \\
\midrule
Optimizer & Adam & Adam & AdamW \\
Learning rate & $5\times10^{-4}$ & $3\times10^{-4}$ & $6\times10^{-4}$ \\
Weight decay & 0 & 0 & $10^{-2}$ \\
$(\beta_1, \beta_2)$ & $(0.9, 0.999)$ & $(0.9, 0.999)$ & $(0.9, 0.95)$ \\
Warmup steps & 1000 & 1500 & 500 \\
Scheduler & \multicolumn{3}{c}{Cosine decay to 0 (FNO, U-Net) or $10^{-5}$ (ScOT)} \\
Residual scale & 1.0 & 0.3 & 1.0 \\
\bottomrule
\end{tabular}
\end{table}

\subsection{Baselines}
\label{app:baselines}\label{app:sbal}

We detail how the active learning baselines of Sec.~\ref{sec:exp_baselines} are adapted to our streaming setting.

\paragraph{SBAL in AL4PDE.}
We follow the pool-based, batch active-learning loop: in each round, a batch of candidate inputs is selected from a pool, simulated with the numerical solver, and added to the training set. Uncertainty is measured via query-by-committee (QbC), i.e., ensemble disagreement, which serves as a sample informativeness score. Stochastic Batch Active Learning (SBAL) draws a batch \textit{without replacement} from the pool using a softened sampling distribution over uncertainties; a sharpness parameter $\beta$ interpolates between random sampling ($\beta=0$) and Top-$K$ selection ($\beta \rightarrow \infty$), thereby encouraging diversity beyond the highest-uncertainty mode:
\[
q_i = \frac{u_i^\beta}{\sum_j u_j^\beta}, \quad \beta \ge 0,
\]
where $u_i$ is the uncertainty score for candidate $i$ \citep{musekamp-ActiveLearningNeural-2025,kirschStochasticBatchAcquisition2023}.

\paragraph{Online adaptation.}
The original SBAL selects from a fixed pool and permanently removes simulated candidates. In our streaming setting, we instead draw a fresh uniform candidate pool at each selection (Sec.~\ref{sec:exp_baselines}). Since the reservoir $\sR$ discards older data, regions that remain uncertain may be selected again, while the soft, uncertainty-weighted batch selection of SBAL is preserved. Selection is triggered on the training rank about ten times per run, after a warm-up that lets the ensemble stabilize. At each trigger, the uncertainty of every candidate is estimated from ensemble rollouts and a batch is sampled with power-law weights (Algorithm~\ref{alg:online_sbal}); the cost of this step is analysed in Appendix~\ref{app:pool_based}.

\begin{algorithm}[h]
  \caption{Online SBAL for Streaming Surrogate Training}
  \label{alg:online_sbal}
  \begin{algorithmic}
  \Require Pool size $P$, batch size $K$, rollout fraction $\rho \in (0,1]$, sharpness $\beta \geq 0$, ensemble $\{f_{\theta}^{(m)}\}_{m=1}^{M}$
  \State \textbf{Construct candidate pool:} Sample $\{\lambda_i\}_{i=1}^{P} \sim \mathcal{U}_\Lambda$ uniformly
  \For{$i = 1, \ldots, P$}
    \State Sample initial condition $u_0 \sim p(\cdot \mid \lambda_i)$
    \State Set rollout horizon $T' \gets \max(1, \lfloor \rho T \rfloor)$
    \For{$t = 0, \ldots, T'-1$}
      \State Compute ensemble predictions $\{\hat{u}_{t+1}^{(m)}\}_{m=1}^{M}$ from $u_t$
      \State Record disagreement $d_t \gets \mathrm{Var}_m\bigl(\hat{u}_{t+1}^{(m)}\bigr)$ or other metric
      \State Update state $u_{t+1} \gets \bar{u}_{t+1}$ \Comment{ensemble mean}
    \EndFor
    \State Aggregate uncertainty: $u_i \gets \dfrac{1}{T'}\sum_{t=0}^{T'-1} d_t$
  \EndFor
  \State \textbf{Compute selection probabilities:} $q_i \gets \dfrac{u_i^\beta}{\sum_{j=1}^{P} u_j^\beta}$ for $i = 1, \ldots, P$
  \State \textbf{Sample batch:} Draw $K$ candidates $\{\lambda_{s_1}, \ldots, \lambda_{s_K}\}$ from $\{q_i\}$ without replacement
  \State \textbf{Return} selected parameters $\{\lambda_{s_k}\}_{k=1}^{K}$ to simulate
  \end{algorithmic}
\end{algorithm}

\paragraph{SBAL hyperparameters.}
We use pool size $P = 5000$, rollout fraction $\rho = 0.5$ (half the full trajectory length), and sharpness $\beta = 2$. This configuration balances computational cost against uncertainty estimation quality.

\paragraph{Breed.}
\label{app:breed_impl}
For the Breed baseline, we use a Gaussian mutation width of $\sigma=0.1$ relative to the parameter range, consistent with the original implementation but adapted for the streaming setting.

\subsection{Existing Software Assets and Licenses}
\label{app:existing_assets}

We use or adapt existing software assets only as credited components. APEBench and Exponax are MIT-licensed packages, and AL4PDE is released under the MIT license.

\section{Additional Analyses}
\label{app:extended_results}

\subsection{Per PDE and per Architecture Trends}
\label{app:rel_improvments_pde_and_arch}

Complete per-PDE tables are reported in Appendix~\ref{app:2d_results} (Tables~\ref{tab:navier-stokes-kolmogorov-flow-2d-low-res}--\ref{tab:gray-scott-beta-low-res}). We summarize their qualitative trends below; the mean--tail trade-off is analysed in Appendix~\ref{app:mean_tail_tradeoff}.

\paragraph{Navier--Stokes: OGAS improves robustness without sacrificing average accuracy.}
On 2D Navier--Stokes (Kolmogorov flow), OGAS consistently improves tail percentiles and worst-case errors across architectures while matching the mean error of the strongest baselines (Table~\ref{tab:navier-stokes-kolmogorov-flow-2d-low-res}). The uniform component of the mixture policy keeps enough coverage of the parameter space for the mean not to degrade.

\paragraph{Kuramoto--Sivashinsky and Gray--Scott: mild mean shifts for small models, large tail gains.}
On 2D KS and $\beta$-Gray--Scott (Tables~\ref{tab:kuramoto-sivashinsky-2d-low-res} and~\ref{tab:gray-scott-beta-low-res}), lower-capacity surrogates (U-Net and, to a lesser extent, FNO) show a trade-off: their mean error can be slightly worse than with uniform or space-filling baselines, while tail and worst-case errors improve substantially. The higher-capacity scOT shows little to no mean degradation and keeps the tail gains, which points to model capacity rather than to the sampling strategy (Appendix~\ref{app:mean_tail_tradeoff}).

\paragraph{Improvement ratios by architecture.}
Table~\ref{tab:rel_imp_archs} summarizes these trends as improvement ratios over Uniform, averaged across PDEs and seeds for each architecture.

\begin{table}[h]
\centering
\footnotesize
\caption{Improvement ratio vs Uniform across Architectures. Values are Mean $\pm$ Std. 1 is no improvement, higher than 1 is improvement, lower than 1 is degradation}
\label{tab:rel_imp_archs}
\begin{subtable}{\linewidth}
\centering
\caption{UNet}
\label{tab:rel_imp_unet_cond}
\begin{tabular}{l c c c c}
\toprule
\textbf{Strategy} & \textbf{mean} & \textbf{max} & \textbf{p99} & \textbf{std} \\
\midrule
OGAS-L & $0.94 \pm 0.10$ & $1.80 \pm 0.53$ & $1.45 \pm 0.21$ & $1.72 \pm 0.27$ \\
OGAS-U & $0.96 \pm 0.11$ & $1.64 \pm 0.16$ & $1.49 \pm 0.39$ & $1.72 \pm 0.43$ \\
Breed & $1.02 \pm 0.13$ & $1.58 \pm 0.56$ & $1.62 \pm 0.46$ & $1.89 \pm 0.44$ \\
SBAL & $0.97 \pm 0.08$ & $0.98 \pm 0.08$ & $0.93 \pm 0.21$ & $0.92 \pm 0.18$ \\
Top-K & $1.01 \pm 0.07$ & $0.98 \pm 0.08$ & $1.01 \pm 0.30$ & $0.97 \pm 0.24$ \\
Sobol & $0.95 \pm 0.05$ & $0.93 \pm 0.12$ & $0.91 \pm 0.10$ & $0.89 \pm 0.11$ \\
\bottomrule
\end{tabular}
\end{subtable}
\vspace{0.5em}
\begin{subtable}{\linewidth}
\centering
\caption{FNO}
\label{tab:rel_imp_fno_cond}
\begin{tabular}{l c c c c}
\toprule
\textbf{Strategy} & \textbf{mean} & \textbf{max} & \textbf{p99} & \textbf{std} \\
\midrule
OGAS-L & $0.93 \pm 0.06$ & $1.54 \pm 0.23$ & $1.50 \pm 0.33$ & $1.80 \pm 0.40$ \\
OGAS-U & $0.96 \pm 0.05$ & $1.30 \pm 0.11$ & $1.27 \pm 0.14$ & $1.47 \pm 0.24$ \\
Breed & $0.82 \pm 0.07$ & $1.12 \pm 0.10$ & $1.13 \pm 0.15$ & $1.36 \pm 0.24$ \\
SBAL & $0.97 \pm 0.06$ & $1.29 \pm 0.38$ & $1.20 \pm 0.28$ & $1.10 \pm 0.15$ \\
Top-K & $0.96 \pm 0.10$ & $1.30 \pm 0.32$ & $1.26 \pm 0.29$ & $1.27 \pm 0.27$ \\
Sobol & $0.98 \pm 0.05$ & $1.02 \pm 0.10$ & $0.99 \pm 0.03$ & $0.99 \pm 0.03$ \\
\bottomrule
\end{tabular}
\end{subtable}
\vspace{0.5em}
\begin{subtable}{\linewidth}
\centering
\caption{scOT}
\label{tab:rel_imp_scot_t256}
\begin{tabular}{l c c c c}
\toprule
\textbf{Strategy} & \textbf{mean} & \textbf{max} & \textbf{p99} & \textbf{std} \\
\midrule
OGAS-L & $1.04 \pm 0.11$ & $1.59 \pm 0.25$ & $1.32 \pm 0.09$ & $1.47 \pm 0.12$ \\
OGAS-U & $0.97 \pm 0.08$ & $1.47 \pm 0.15$ & $1.20 \pm 0.13$ & $1.30 \pm 0.15$ \\
Breed & $0.92 \pm 0.13$ & $0.88 \pm 0.34$ & $0.93 \pm 0.34$ & $1.01 \pm 0.52$ \\
SBAL & $1.01 \pm 0.09$ & $0.99 \pm 0.18$ & $0.97 \pm 0.14$ & $0.98 \pm 0.19$ \\
Top-K & $1.03 \pm 0.11$ & $0.92 \pm 0.25$ & $0.95 \pm 0.22$ & $0.90 \pm 0.26$ \\
Sobol & $1.05 \pm 0.08$ & $1.02 \pm 0.12$ & $1.01 \pm 0.10$ & $0.99 \pm 0.11$ \\
\bottomrule
\end{tabular}
\end{subtable}
\vspace{0.5em}
\end{table}

\subsection{Interpreting the Mean--Tail Tradeoff}
\label{app:mean_tail_tradeoff}

\paragraph{Capacity-limited learning under heterogeneous dynamics.}
KS and Gray--Scott in our 2D setup include large variations in the effective domain size parameter $L$, which changes the characteristic spatial scales of the dynamics at fixed grid resolution. This induces a particularly heterogeneous training distribution: different parameter regimes correspond to dynamics with different dominant wavelengths and temporal behaviors. For limited-capacity surrogates, fully fitting all regimes is difficult under a fixed simulation budget, and directing sampling toward high-signal regimes can emphasize harder cases that dominate tail metrics. In this regime, improved robustness can come before (or even temporarily trade against) improvements in average error. The fact that scOT largely avoids the mean degradation supports the interpretation that OGAS is primarily exposing the training process to genuinely difficult regimes, and that sufficient model capacity is required to translate this harder curriculum into improved mean performance within the same budget.

\paragraph{Training horizon relative to 1D.}
The 2D experiments are conducted under a comparatively tight training horizon: we run a fixed budget of 2D solver trajectories and, relative to the 1D suite, perform substantially fewer surrogate update steps (10k vs 30k simulations, 17k batches vs 60k batches). 
In the 1D long-horizon studies, improvements in tail metrics typically emerge early, whereas improvements in mean error continue to accrue later with additional updates once the model has absorbed the diversity of regimes emphasized by the sampling policy. This provides controlled evidence that the modest mean differences observed for U-Net/FNO on KS and Gray--Scott in 2D may partially reflect an early-to-mid training regime under limited budget, rather than a fundamental limitation of signal-guided sampling. We do not extrapolate beyond the reported 2D budgets; instead, we use 1D as a controlled setting to illustrate how the mean metric can continue improving with longer training under the same policy.

\subsection{Mean-error Gains Under Narrower Parameter Ranges}
\label{app:narrower_range}

The 2D headline experiments use intentionally broad parameter ranges to span diverse physical regimes ($L\in[10,130]$ for KS; $L\in[0.5,10]$ for GS), which produces highly heterogeneous dynamics relative to the surrogate's capacity. To test whether the mild mean-error degradation observed for low-capacity surrogates (U-Net) on KS and GS is driven by this heterogeneity rather than by \textit{OGAS} itself, we re-run UNet/KS and UNet/GS with narrower parameter ranges that make the task easier:
KS from $L\in[10,130]$ to $L\in[10,25]$; GS from $L\in[0.5,10]$ to $L\in[0.5,2.5]$.
All other settings are identical to the main 2D protocol (same architecture, optimizer, simulation budget $N\eq10^4$, seeds, validation set construction).

Table~\ref{tab:narrower_range} reports the resulting comparison against uniform sampling. Under these narrower ranges, \textit{OGAS-L} improves \emph{both} the mean RMSE and the tail/dispersion metrics on both PDEs, e.g., +4.4\% mean and +14.1\% \textit{p99} on GS; +4.3\% mean and +7.9\% \textit{p99} on KS. Combined with the per-architecture trends in Appendix~\ref{app:rel_improvments_pde_and_arch}, this is consistent with the interpretation that mean-error degradations on the broad-range 2D headline experiments reflect a capacity-limited curriculum effect rather than a fundamental robustness--accuracy trade-off intrinsic to OGAS. We treat this as a sensitivity check, not a headline result: the broad-range setting in the main paper is the harder, more realistic benchmark.

\begin{table}[h]
\centering
\footnotesize
\caption{\textbf{Narrower-range sensitivity check (UNet, $N\eq10^4$, 2D, 3 seeds).} \textit{OGAS-L} vs.\ Uniform on KS and GS with reduced domain-size ranges. Lower values are better; ``Rel.'' is the relative reduction of \textit{OGAS-L} over Uniform.}
\label{tab:narrower_range}
\setlength{\tabcolsep}{4pt}
\renewcommand{\arraystretch}{1.1}
\begin{tabular*}{\linewidth}{@{\extracolsep{\fill}}llcccc@{}}
\toprule
PDE / Model & Sampling & RMSE-mean & RMSE-std & RMSE-p95 & RMSE-p99 \\
\midrule
GS / UNet & Uniform & $2.80\mathrm{e}{-3}\!\pm\!9.0\mathrm{e}{-5}$ & $1.19\mathrm{e}{-3}\!\pm\!2.8\mathrm{e}{-5}$ & $4.48\mathrm{e}{-3}\!\pm\!1.8\mathrm{e}{-4}$ & $6.33\mathrm{e}{-3}\!\pm\!2.2\mathrm{e}{-4}$ \\
          & OGAS-L  & $2.68\mathrm{e}{-3}\!\pm\!6.3\mathrm{e}{-5}$ & $9.88\mathrm{e}{-4}\!\pm\!4.7\mathrm{e}{-5}$ & $4.20\mathrm{e}{-3}\!\pm\!6.2\mathrm{e}{-5}$ & $5.44\mathrm{e}{-3}\!\pm\!3.2\mathrm{e}{-5}$ \\
          & \textit{Rel.} & $+4.38\%$ & $+17.15\%$ & $+6.31\%$ & $+14.09\%$ \\
\midrule
KS / UNet & Uniform & $6.57\mathrm{e}{-4}\!\pm\!2.4\mathrm{e}{-5}$ & $1.57\mathrm{e}{-4}\!\pm\!1.3\mathrm{e}{-5}$ & $9.04\mathrm{e}{-4}\!\pm\!3.8\mathrm{e}{-5}$ & $1.17\mathrm{e}{-3}\!\pm\!5.2\mathrm{e}{-5}$ \\
          & OGAS-L  & $6.29\mathrm{e}{-4}\!\pm\!4.1\mathrm{e}{-5}$ & $1.32\mathrm{e}{-4}\!\pm\!2.4\mathrm{e}{-5}$ & $8.39\mathrm{e}{-4}\!\pm\!6.9\mathrm{e}{-5}$ & $1.08\mathrm{e}{-3}\!\pm\!1.3\mathrm{e}{-4}$ \\
          & \textit{Rel.} & $+4.28\%$ & $+15.76\%$ & $+7.27\%$ & $+7.93\%$ \\
\bottomrule
\end{tabular*}
\renewcommand{\arraystretch}{1.0}
\end{table}

\subsection{Hard DDPM Dataset Evaluation}
\label{app:hard_ddpm_eval}
We also evaluate the U-Net surrogates on a validation set drawn from the \textit{OGAS} generator: 250 parameter configurations from the final DDPM of each of the three models produced by the different seeds of the \textit{OGAS-Loss} runs.

\begin{figure*}[h]
  \centering
  \includegraphics[width=0.7\linewidth]{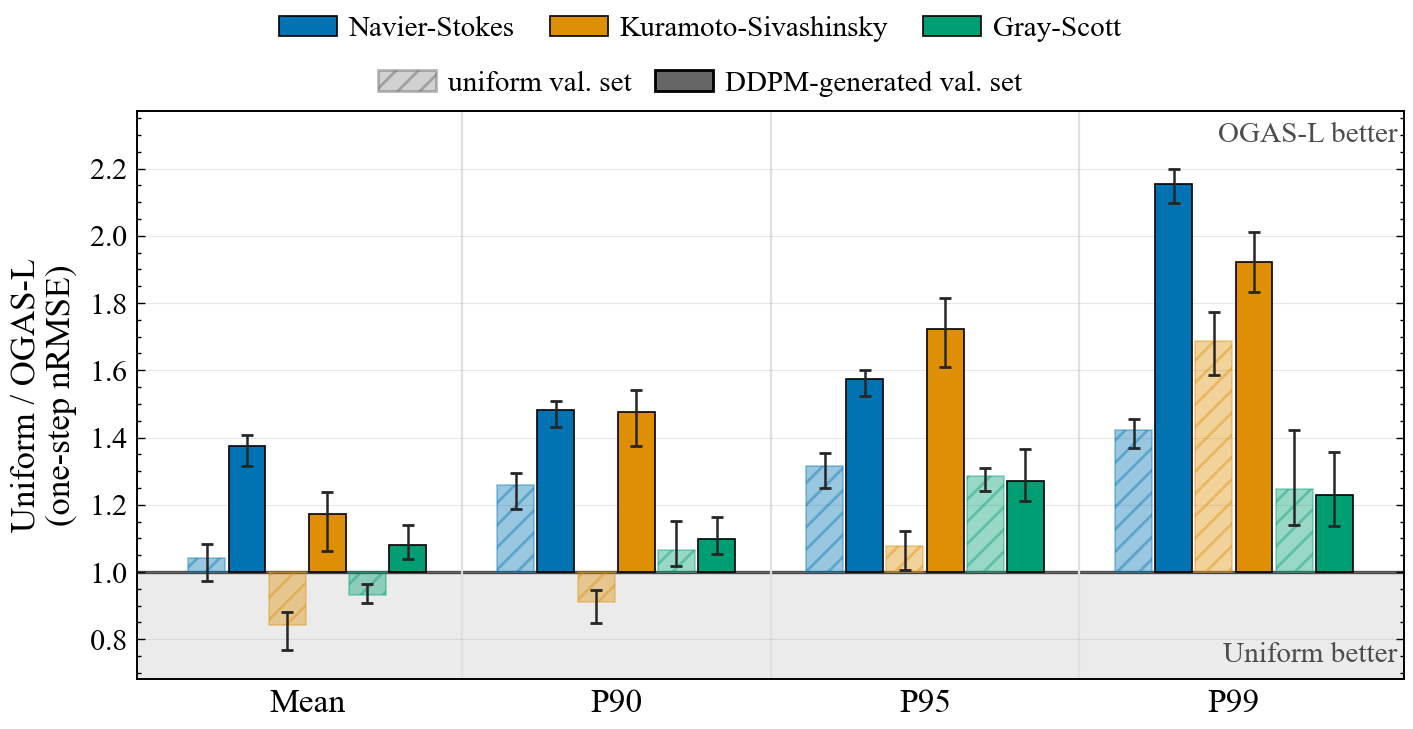}
  \caption{U-Net improvement ratios (Uniform / \textit{OGAS-Loss}) of the one-step normalized RMSE on the uniform validation set and on the DDPM validation set. Higher than 1 is an improvement of \textit{OGAS-Loss}.}
  \label{fig:hard_set_improvment}
\end{figure*}

Table~\ref{tab:hard_set} and Figure~\ref{fig:hard_set_improvment} compare \textit{OGAS-Loss} and \textit{Uniform} on both sets. On the uniform set, \textit{OGAS-Loss} improves the tail, with \textit{RMSE-p99} ratios between $1.24$ and $1.68$, while its mean error is similar or slightly higher. On the DDPM set, it improves every metric on the three PDEs, with \textit{RMSE-mean} ratios from $1.08$ to $1.38$ and \textit{RMSE-p99} ratios from $1.22$ to $2.16$. Both surrogates make larger errors on this set; the mean error of \textit{Uniform} on Navier Stokes, for instance, more than doubles. The configurations that \textit{OGAS} finds difficult are thus also difficult for a uniformly trained surrogate, suggesting that difficulty transfers between models.

\begin{table}[h]
\centering
\footnotesize
\caption{\textbf{DDPM validation set (U-Net, 2D).} One-step normalized RMSE ($\times 10^{-2}$) of \textit{OGAS-Loss} and Uniform on the uniform validation set of the main benchmark and on the validation set sampled from the final \textit{OGAS-Loss} DDPM. Values show mean $\pm$ std across seeds. Best per PDE and validation set in \textbf{bold}.}
\label{tab:hard_set}
\resizebox{\linewidth}{!}{%
\begin{tabular}{lllccccc}
\toprule
PDE & Val. set & Strategy & Mean & Std & P90 & P95 & P99 \\
\midrule
Navier--Stokes & Uniform & OGAS-Loss & $\mathbf{0.893 \pm 0.053}$ & $\mathbf{0.273 \pm 0.007}$ & $\mathbf{1.27 \pm 0.06}$ & $\mathbf{1.44 \pm 0.07}$ & $\mathbf{1.80 \pm 0.06}$ \\
 &  & Uniform & $0.927 \pm 0.020$ & $0.488 \pm 0.007$ & $1.59 \pm 0.02$ & $1.89 \pm 0.03$ & $2.56 \pm 0.03$ \\
\cmidrule(l){2-8}
 & DDPM & OGAS-Loss & $\mathbf{1.56 \pm 0.06}$ & $\mathbf{0.350 \pm 0.005}$ & $\mathbf{2.01 \pm 0.06}$ & $\mathbf{2.17 \pm 0.06}$ & $\mathbf{2.53 \pm 0.06}$ \\
 &  & Uniform & $2.15 \pm 0.03$ & $0.859 \pm 0.034$ & $2.97 \pm 0.03$ & $3.42 \pm 0.06$ & $5.46 \pm 0.33$ \\
\midrule
KS & Uniform & OGAS-Loss & $0.511 \pm 0.041$ & $\mathbf{0.132 \pm 0.005}$ & $0.640 \pm 0.040$ & $\mathbf{0.699 \pm 0.042}$ & $\mathbf{0.874 \pm 0.050}$ \\
 &  & Uniform & $\mathbf{0.428 \pm 0.006}$ & $0.261 \pm 0.014$ & $\mathbf{0.582 \pm 0.009}$ & $0.751 \pm 0.032$ & $1.47 \pm 0.06$ \\
\cmidrule(l){2-8}
 & DDPM & OGAS-Loss & $\mathbf{0.530 \pm 0.045}$ & $\mathbf{0.221 \pm 0.008}$ & $\mathbf{0.729 \pm 0.045}$ & $\mathbf{0.821 \pm 0.051}$ & $\mathbf{1.22 \pm 0.06}$ \\
 &  & Uniform & $0.619 \pm 0.034$ & $0.456 \pm 0.036$ & $1.07 \pm 0.08$ & $1.41 \pm 0.12$ & $2.34 \pm 0.21$ \\
\midrule
Gray--Scott & Uniform & OGAS-Loss & $1.37 \pm 0.04$ & $\mathbf{0.829 \pm 0.069}$ & $\mathbf{2.42 \pm 0.16}$ & $\mathbf{3.16 \pm 0.09}$ & $\mathbf{4.35 \pm 0.50}$ \\
 &  & Uniform & $\mathbf{1.27 \pm 0.23}$ & $1.14 \pm 0.41$ & $2.57 \pm 0.45$ & $4.05 \pm 1.49$ & $5.38 \pm 2.17$ \\
\cmidrule(l){2-8}
 & DDPM & OGAS-Loss & $\mathbf{1.98 \pm 0.09}$ & $\mathbf{0.915 \pm 0.065}$ & $\mathbf{2.93 \pm 0.15}$ & $\mathbf{3.71 \pm 0.24}$ & $\mathbf{4.74 \pm 0.42}$ \\
 &  & Uniform & $2.14 \pm 0.22$ & $1.23 \pm 0.32$ & $3.22 \pm 0.30$ & $4.70 \pm 1.51$ & $5.80 \pm 2.20$ \\
\bottomrule
\end{tabular}}
\end{table}

\begin{figure*}[h]
  \centering
  \includegraphics[width=\linewidth]{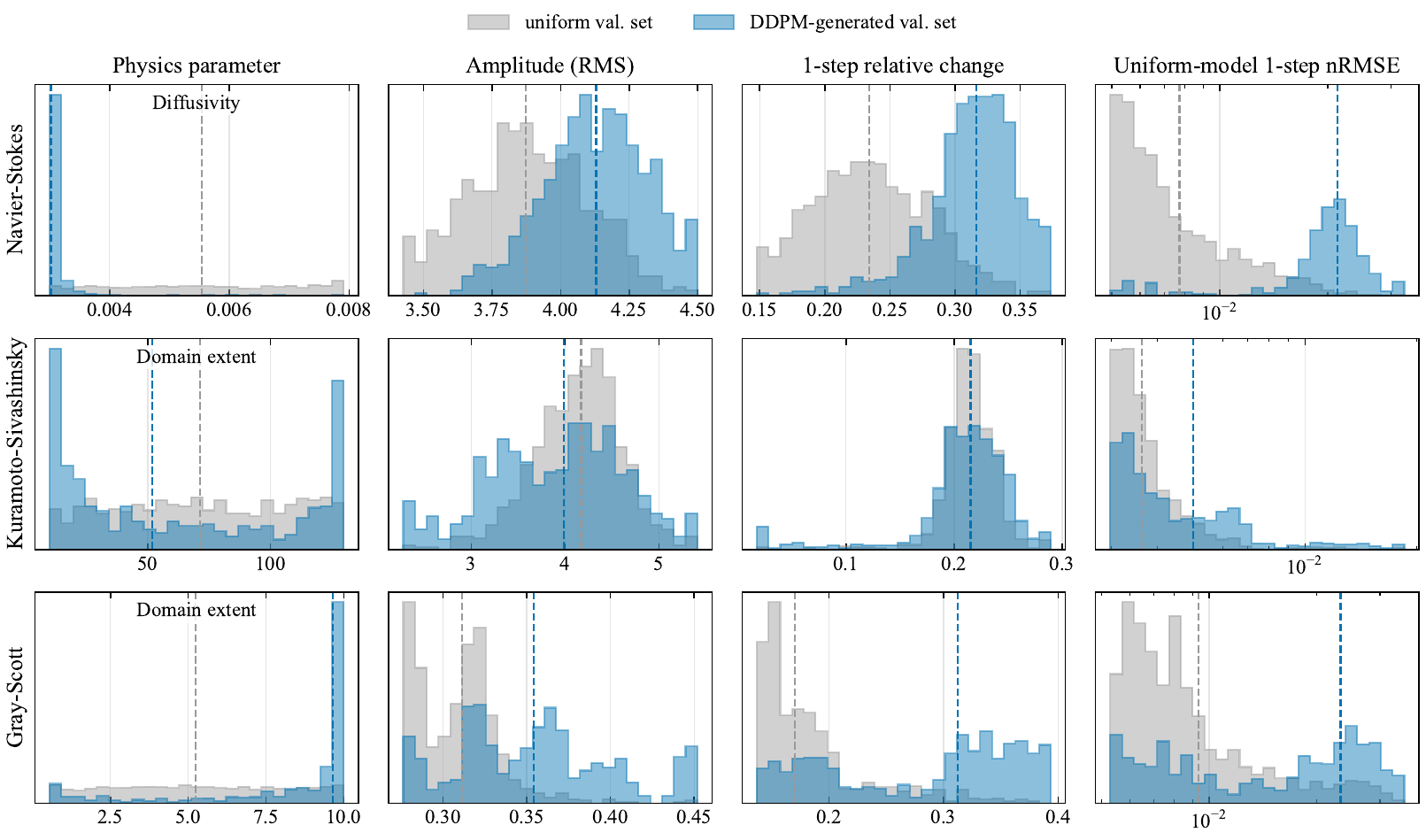}
  \caption{Distributions of trajectory descriptors on the uniform validation set (gray) and on the DDPM validation set (blue), for each 2D PDE: physics parameter, amplitude (RMS of the field), 1-step relative change $\|u_{t+1}-u_t\|/\|u_t\|$, and one-step normalized RMSE of the Uniform U-Net. Dashed lines are medians.}
  \label{fig:valset_descriptors}
\end{figure*}

Figure~\ref{fig:valset_descriptors} compares a number of  distributions between the two sets. In parameter space the DDPM set is concentrated: it targets low diffusivity on Navier Stokes, large domains on Gray Scott, and both ends of the domain range on KS. However, this does not make its trajectories less diverse as the mean pairwise distance between states is higher than in the uniform set on Gray Scott ($0.78$ vs.\ $0.62$ at the initial time) and Navier Stokes, and equal on KS. In terms of dynamics and difficulty, the DDPM set is even more evenly spread. On Gray Scott, the uniform set mostly contains slow trajectories that are easy to predict, whereas the DDPM set covers the whole range. On Navier Stokes, it shifts toward faster dynamics where the median error of Uniform is $2.8$ times larger.

Yet, neither validation set is neutral.  The uniform set is the common practice, but it follows the training distribution of \textit{Uniform}, which favors it on the mean error, and does not guarantee that difficult regimes are well covered. The DDPM set follows a distribution that has been learnt from the surrogate training which is adversarial: the generator targets the configurations where the \textit{OGAS} surrogate fails most, and this difficulty transfers to \textit{Uniform}, whose errors also increase on this set for a given type of surrogate. 
\subsection{Single-model Evaluation and Ensemble Size Ablation}
\label{app:single_model_eval}

In the base experiments, the loss signal of \textit{OGAS-L} is averaged over the two model ensemble that uncertainty based methods require (denoted \textit{OGAS-L2} here). To check that OGAS does not depend on this ensembling, we also train \textit{OGAS-L} with a single surrogate, and evaluate a single model for every method (the first ensemble member when an ensemble is trained). Tables~\ref{tab:navier-stokes-single}--\ref{tab:gs-single} (Appendix~\ref{app:single_model_tables}) report the results. Single-model \textit{OGAS-L} reduces \textit{RMSE-std} relative to Uniform in all nine PDE and architecture settings and \textit{RMSE-p99} in eight of them, but its mean error is higher than Uniform in all nine. It is also generally behind \textit{OGAS-L2}, unless for Gray-Scott with U-Net. This suggests that the ensemble-averaged loss is a less noisy training signal, while OGAS does not require it to improve tail metrics.

\label{app:ensemble_size}
Finally, we vary the size of the ensemble that provides the disagreement signal of \textit{OGAS-Unc}, with $M=2$, $3$ and $5$ members, on KS with U-Net. To compare the quality of the signal rather than the benefit of averaging predictions, every model is evaluated through its first member (Table~\ref{tab:abl-ensemble-size}). A larger ensemble barely changes the results: from two to five members, the mean and 99th-percentile errors decrease by less than one standard deviation, and all sizes remain behind \textit{OGAS-Loss} on the tail. This is on par with \cite{musekamp-ActiveLearningNeural-2025}, who report minor gains beyond two models, and supports the use of $M=2$ in the main experiments.

\begin{table}[h]
\centering
\footnotesize
\caption{\textbf{Ensemble size ablation (2D Kuramoto-Sivashinsky, UNet).} \textit{OGAS-Unc} with $M\in\{2,3,5\}$ ensemble members; every model is evaluated through its first member. Values show mean $\pm$ std across seeds. Best in \textbf{bold}.}
\label{tab:abl-ensemble-size}
\begin{tabular}{lccccc}
\toprule
Strategy & Mean ($\times 10^{-3}$) & Std ($\times 10^{-3}$) & Max ($\times 10^{-2}$) & P99 ($\times 10^{-2}$) & P95 ($\times 10^{-3}$) \\
\midrule
OGAS-Loss & $6.38 \pm 0.30$ & $\mathbf{1.54 \pm 0.16}$ & $\mathbf{5.43 \pm 0.18}$ & $\mathbf{1.05 \pm 0.05}$ & $8.46 \pm 0.27$\\
OGAS-Unc, $M{=}2$ & $6.29 \pm 0.09$ & $1.92 \pm 0.15$ & $6.91 \pm 0.67$ & $1.34 \pm 0.06$ & $8.77 \pm 0.27$ \\
OGAS-Unc, $M{=}3$ & $6.17 \pm 0.31$ & $1.78 \pm 0.26$ & $6.87 \pm 0.94$ & $1.31 \pm 0.12$ & $8.38 \pm 0.34$ \\
OGAS-Unc, $M{=}5$ & $6.14 \pm 0.14$ & $1.89 \pm 0.03$ & $7.34 \pm 0.75$ & $1.31 \pm 0.04$ & $\mathbf{8.25 \pm 0.22}$\\
Uniform & $\mathbf{5.28 \pm 0.41}$ & $2.95 \pm 0.55$ & $10.05 \pm 0.84$ & $1.70 \pm 0.31$ & $9.06 \pm 1.35$\\
\bottomrule
\end{tabular}
\end{table}

\subsection{Ablation on Using CRPS as Another Conditioning Metric}
\label{app:crps_ablation}

Ensemble disagreement measures how much the members differ, not whether they are accurate. As an alternative signal that accounts for both mixing loss and uncertainty, we condition \textit{OGAS-Unc} on the energy score of the ensemble prediction (eCRPS), computed on each training sample as $\frac{1}{M}\sum_{m}|\hat{u}_m-u|-\frac{1}{2M^2}\sum_{m,m'}|\hat{u}_m-\hat{u}_{m'}|$ averaged over the grid, with $\hat{u}_m$ the prediction of member $m$ and $u$ the solver target. Table~\ref{tab:abl-ecrps} compares it with \textit{OGAS-Unc} and \textit{OGAS-Loss} on KS for the three architectures (3 seeds, $M{=}2$, ensemble prediction).

Here the eCRPS signal gives the best or near-best tail metrics on all three architectures. Relative to the disagreement, it lowers \textit{RMSE-p99} from $0.94$ to $0.87$ (U-Net), $2.04$ to $1.43$ (FNO) and $1.82$ to $1.71\times10^{-2}$ (scOT), and it matches \textit{OGAS-Loss} on U-Net and FNO. On scOT, it also gives the lowest \textit{RMSE-mean} ($1.10$ vs.\ $1.16\times10^{-2}$ for Uniform).

\begin{table}[h]
\centering
\footnotesize
\caption{\textbf{Acquisition signal ablation (2D Kuramoto-Sivashinsky).} \textit{OGAS-Unc} conditioned on the ensemble disagreement (main experiments) or on the energy CRPS of the ensemble prediction; $M{=}2$, ensemble prediction. Values show mean $\pm$ std across seeds. Best per architecture in \textbf{bold}.}
\label{tab:abl-ecrps}
\begin{subtable}{\linewidth}
\centering
\caption{UNet}
\begin{tabular}{lccccc}
\toprule
Strategy & Mean ($\times 10^{-3}$) & Std ($\times 10^{-3}$) & Max ($\times 10^{-2}$) & P99 ($\times 10^{-2}$) & P95 ($\times 10^{-3}$) \\
\midrule
OGAS-Loss & $5.11 \pm 0.41$ & $1.32 \pm 0.05$ & $\mathbf{5.08 \pm 0.41}$ & $\mathbf{0.87 \pm 0.05}$ & $6.99 \pm 0.42$ \\
OGAS-Unc (disagreement) & $4.86 \pm 0.46$ & $1.37 \pm 0.13$ & $6.07 \pm 0.81$ & $0.94 \pm 0.12$ & $\mathbf{6.70 \pm 0.59}$ \\
OGAS-Unc (eCRPS) & $4.95 \pm 0.47$ & $\mathbf{1.28 \pm 0.11}$ & $5.48 \pm 0.38$ & $\mathbf{0.87 \pm 0.09}$ & $6.80 \pm 0.69$ \\
Uniform & $\mathbf{4.28 \pm 0.06}$ & $2.61 \pm 0.14$ & $9.53 \pm 0.07$ & $1.47 \pm 0.06$ & $7.51 \pm 0.32$ \\
\bottomrule
\end{tabular}
\end{subtable}
\vspace{0.5em}
\begin{subtable}{\linewidth}
\centering
\caption{FNO}
\begin{tabular}{lccccc}
\toprule
Strategy & Mean ($\times 10^{-3}$) & Std ($\times 10^{-3}$) & Max ($\times 10^{-2}$) & P99 ($\times 10^{-2}$) & P95 ($\times 10^{-2}$) \\
\midrule
OGAS-Loss & $9.49 \pm 0.28$ & $1.76 \pm 0.06$ & $3.26 \pm 0.18$ & $1.45 \pm 0.03$ & $1.22 \pm 0.04$ \\
OGAS-Unc (disagreement) & $8.63 \pm 0.02$ & $2.73 \pm 0.16$ & $4.88 \pm 0.28$ & $2.04 \pm 0.09$ & $1.29 \pm 0.04$ \\
OGAS-Unc (eCRPS) & $9.40 \pm 0.41$ & $\mathbf{1.70 \pm 0.11}$ & $\mathbf{3.23 \pm 0.12}$ & $\mathbf{1.43 \pm 0.07}$ & $\mathbf{1.19 \pm 0.06}$ \\
Uniform & $\mathbf{8.30 \pm 0.32}$ & $4.02 \pm 0.16$ & $5.95 \pm 0.12$ & $2.79 \pm 0.07$ & $1.53 \pm 0.06$ \\
\bottomrule
\end{tabular}
\end{subtable}
\vspace{0.5em}
\begin{subtable}{\linewidth}
\centering
\caption{scOT}
\begin{tabular}{lccccc}
\toprule
Strategy & Mean ($\times 10^{-2}$) & Std ($\times 10^{-3}$) & Max ($\times 10^{-2}$) & P99 ($\times 10^{-2}$) & P95 ($\times 10^{-2}$) \\
\midrule
OGAS-Loss & $1.22 \pm 0.12$ & $\mathbf{1.92 \pm 0.25}$ & $3.20 \pm 0.23$ & $\mathbf{1.71 \pm 0.15}$ & $1.51 \pm 0.16$ \\
OGAS-Unc (disagreement) & $1.30 \pm 0.15$ & $2.16 \pm 0.35$ & $\mathbf{3.19 \pm 0.28}$ & $1.82 \pm 0.23$ & $1.63 \pm 0.22$ \\
OGAS-Unc (eCRPS) & $\mathbf{1.10 \pm 0.10}$ & $1.97 \pm 0.28$ & $3.39 \pm 0.33$ & $\mathbf{1.71 \pm 0.22}$ & $\mathbf{1.40 \pm 0.14}$ \\
Uniform & $1.16 \pm 0.11$ & $2.82 \pm 0.95$ & $4.80 \pm 0.39$ & $2.32 \pm 0.58$ & $1.66 \pm 0.35$ \\
\bottomrule
\end{tabular}
\end{subtable}
\end{table}

\subsection{Evolution of the Sampling Distribution}
\label{app:mixture_sampling}

\begin{figure*}[h]
\centering
\begin{subfigure}{0.32\textwidth}
  \includegraphics[width=\linewidth]{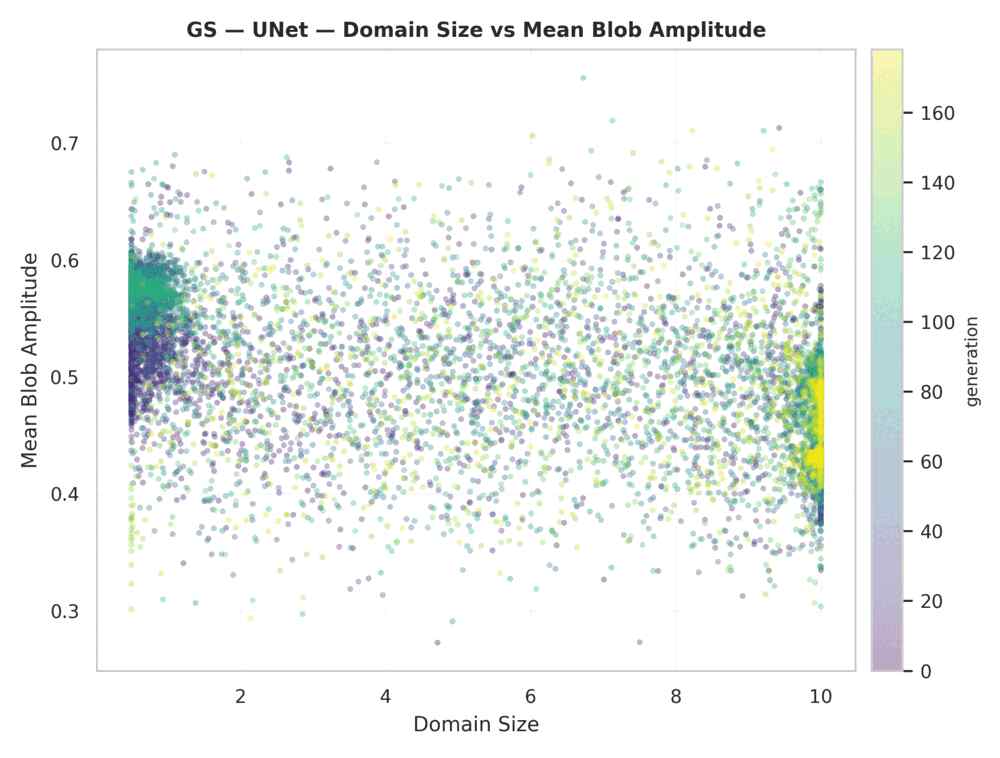}
  \caption{U-Net (Domain vs Amplitude)}
\end{subfigure}
\hfill
\begin{subfigure}{0.32\textwidth}
  \includegraphics[width=\linewidth]{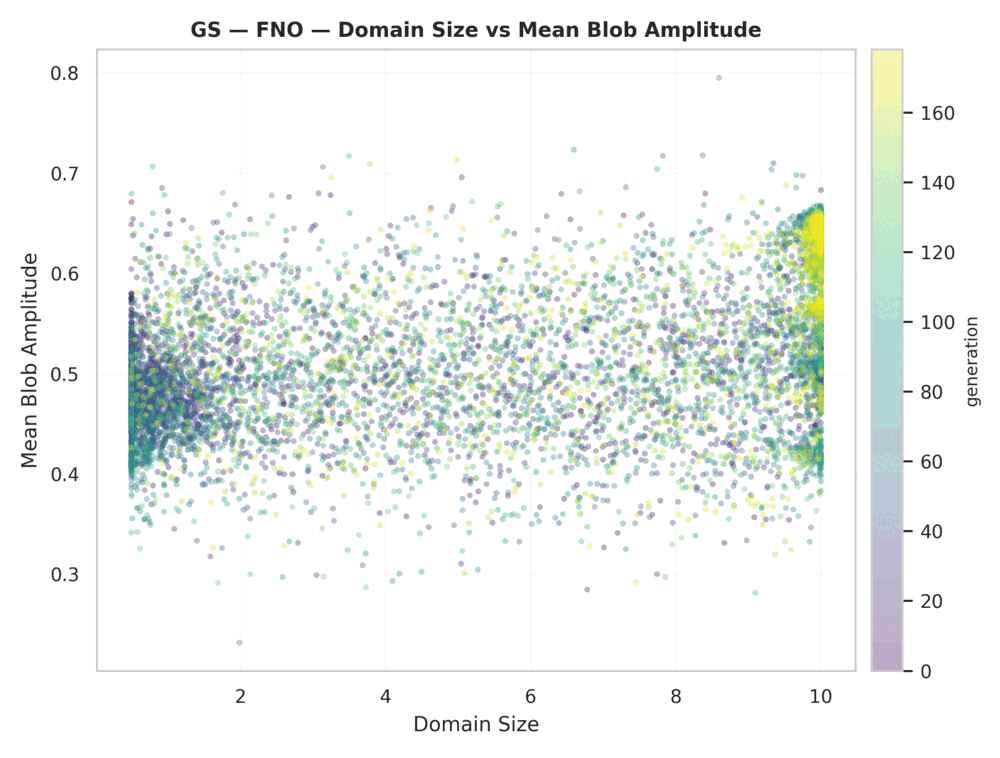}
  \caption{FNO (Domain vs Amplitude)}
\end{subfigure}
\hfill
\begin{subfigure}{0.32\textwidth}
  \includegraphics[width=\linewidth]{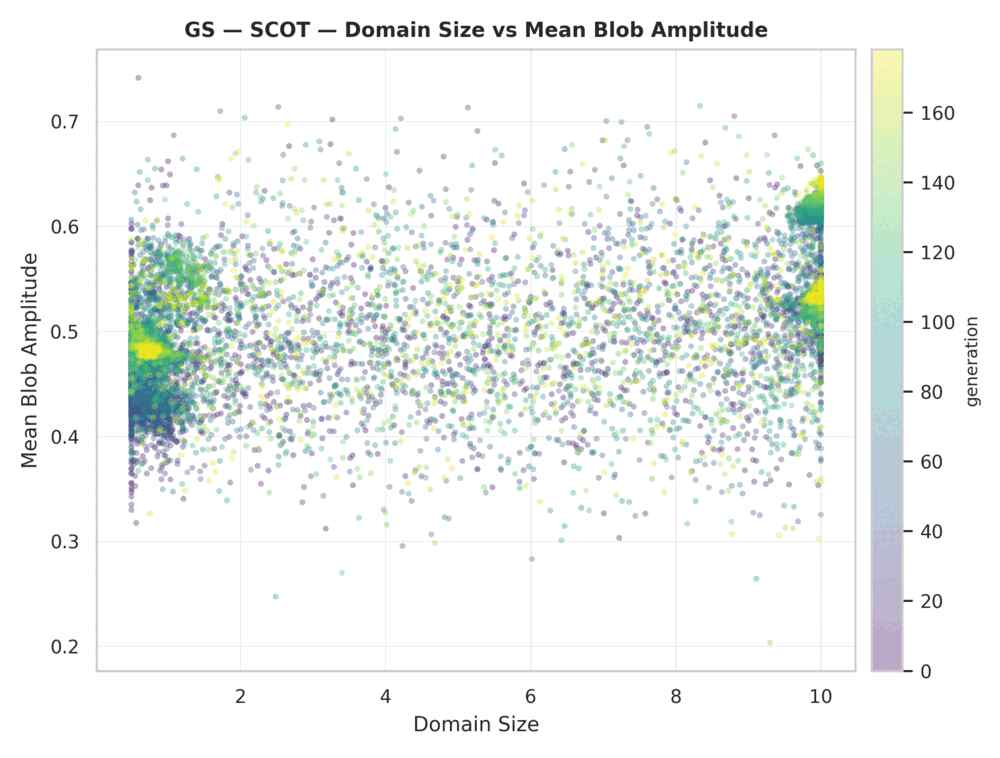}
  \caption{ScOT (Domain vs Amplitude)}
\end{subfigure}

\vspace{0.5em}

\begin{subfigure}{0.32\textwidth}
  \includegraphics[width=\linewidth]{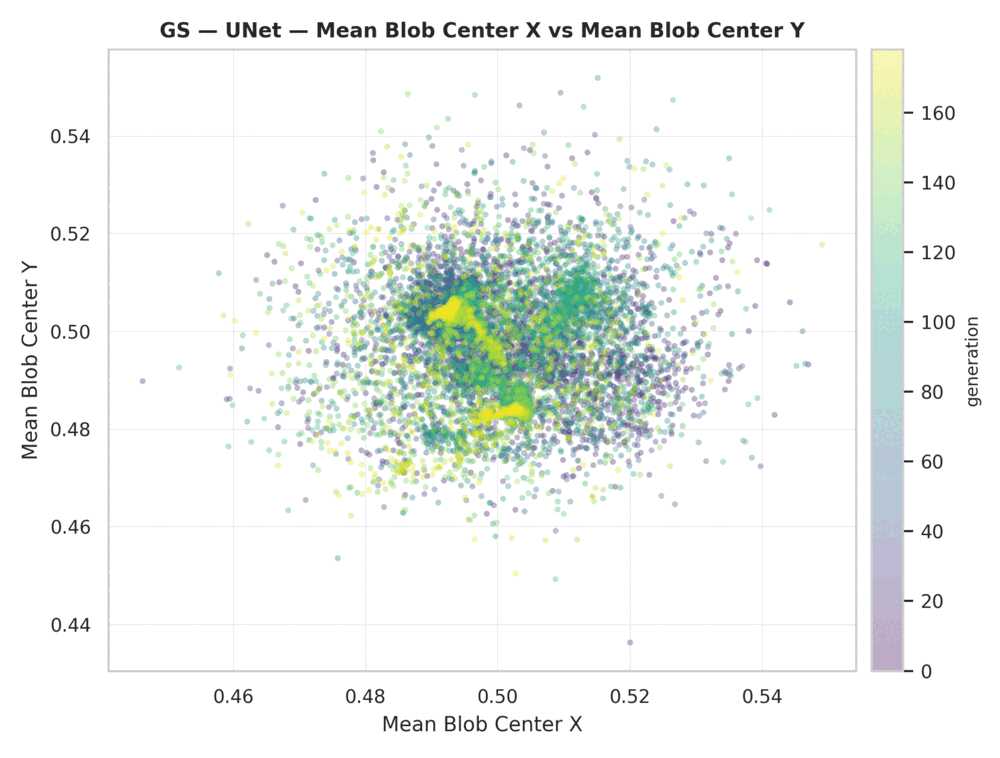}
  \caption{U-Net (Blob Centers)}
\end{subfigure}
\hfill
\begin{subfigure}{0.32\textwidth}
  \includegraphics[width=\linewidth]{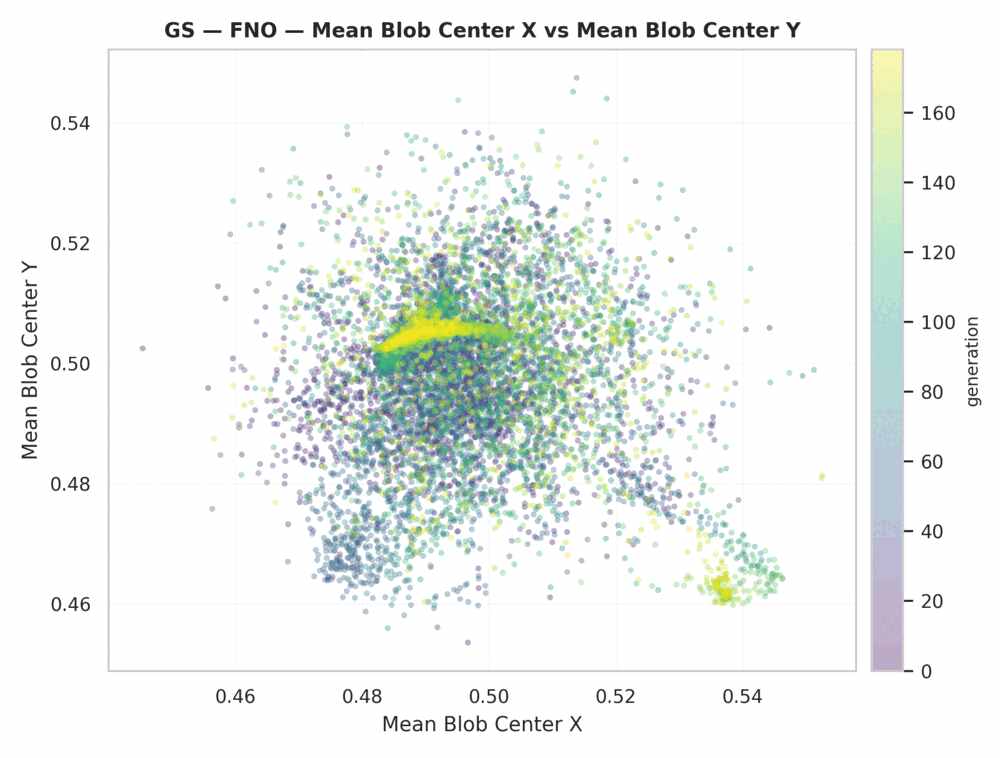}
  \caption{FNO (Blob Centers)}
\end{subfigure}
\hfill
\begin{subfigure}{0.32\textwidth}
  \includegraphics[width=\linewidth]{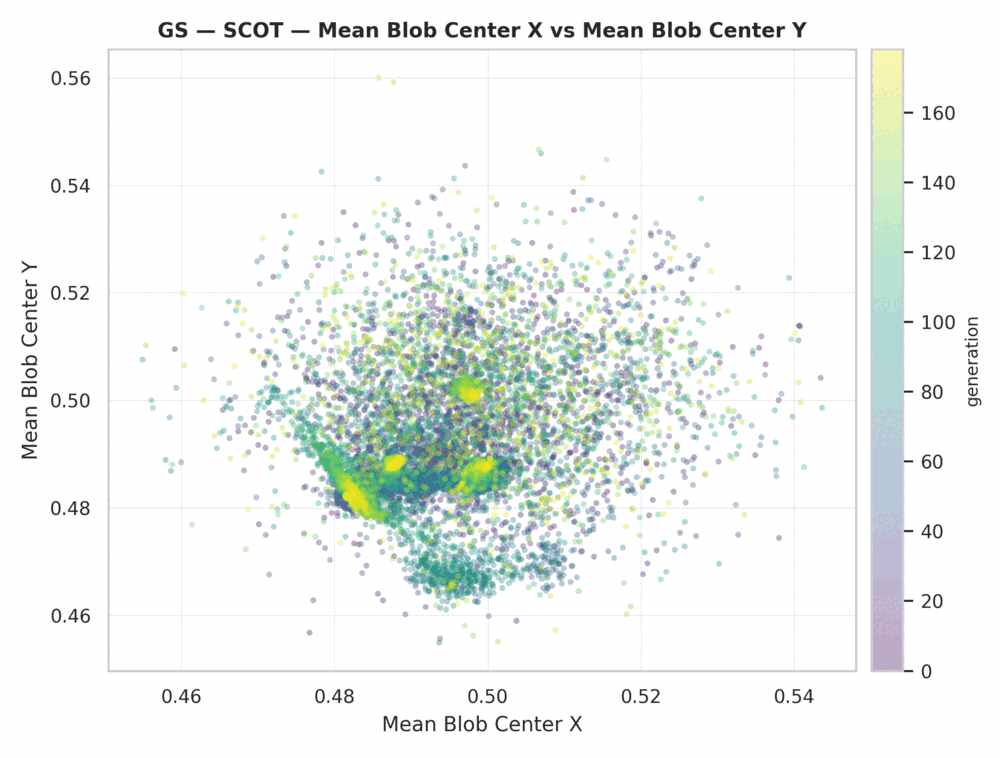}
  \caption{ScOT (Blob Centers)}
\end{subfigure}

\Description{Six-panel Gray--Scott plot showing how U-Net, FNO, and scOT sample distributions evolve for domain extent versus blob amplitude and for the two mean blob-center coordinates.}
\caption{Mixture sampling evolution for Gray-Scott.}
\label{fig:gs_mixture_evolution}
\end{figure*}

\begin{figure*}[h]
\centering
\begin{subfigure}{0.32\textwidth}
  \includegraphics[width=\linewidth]{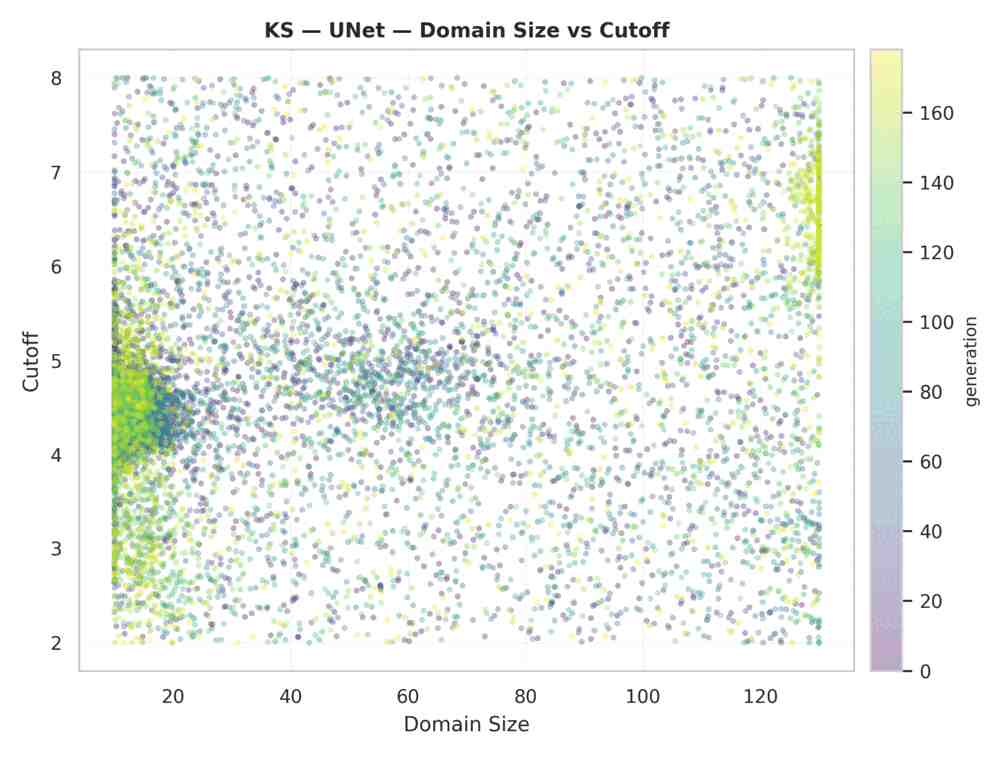}
  \caption{U-Net}
\end{subfigure}
\hfill
\begin{subfigure}{0.32\textwidth}
  \includegraphics[width=\linewidth]{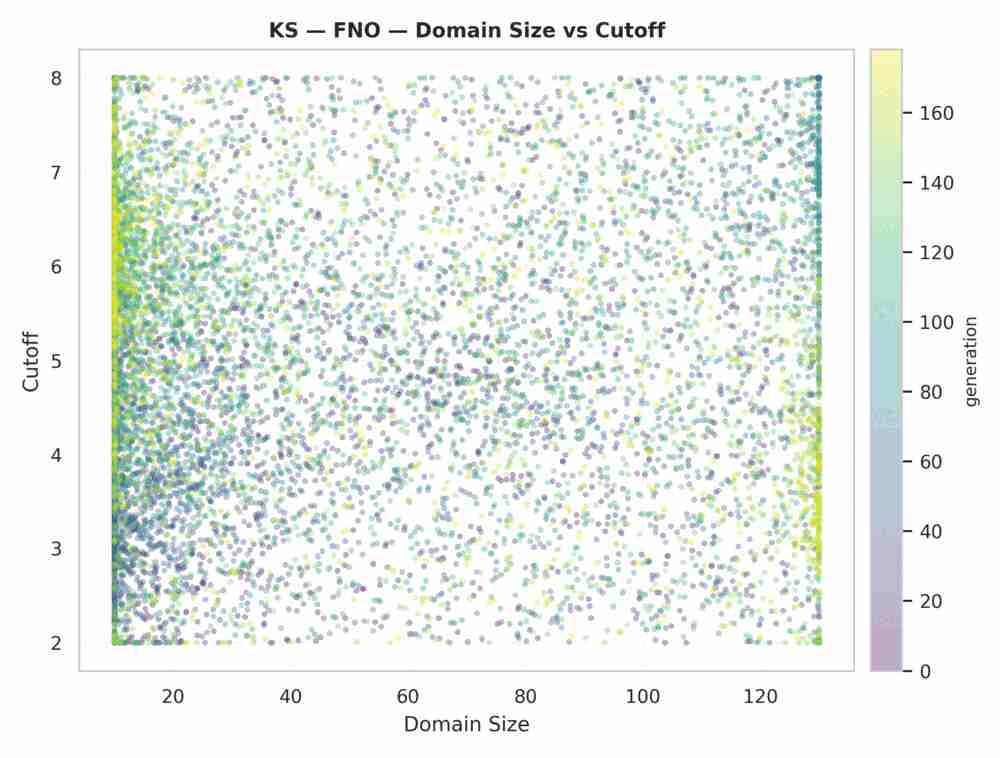}
  \caption{FNO}
\end{subfigure}
\hfill
\begin{subfigure}{0.32\textwidth}
  \includegraphics[width=\linewidth]{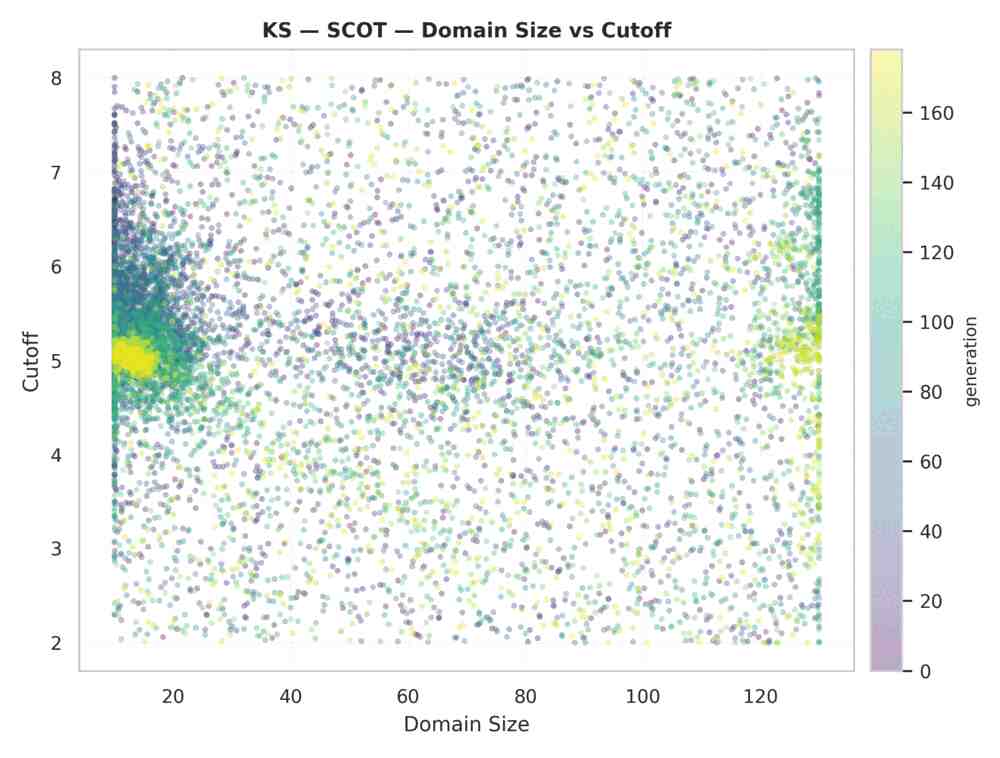}
  \caption{ScOT}
\end{subfigure}
\Description{Three-panel Kuramoto--Sivashinsky plot showing the evolution of sampled domain extent and spectral cutoff distributions for U-Net, FNO, and scOT.}
\caption{Mixture sampling evolution for Kuramoto-Sivashinsky.}
\label{fig:ks_mixture_evolution}
\end{figure*}

\begin{figure*}[h]
\centering
\begin{subfigure}{0.32\textwidth}
  \includegraphics[width=\linewidth]{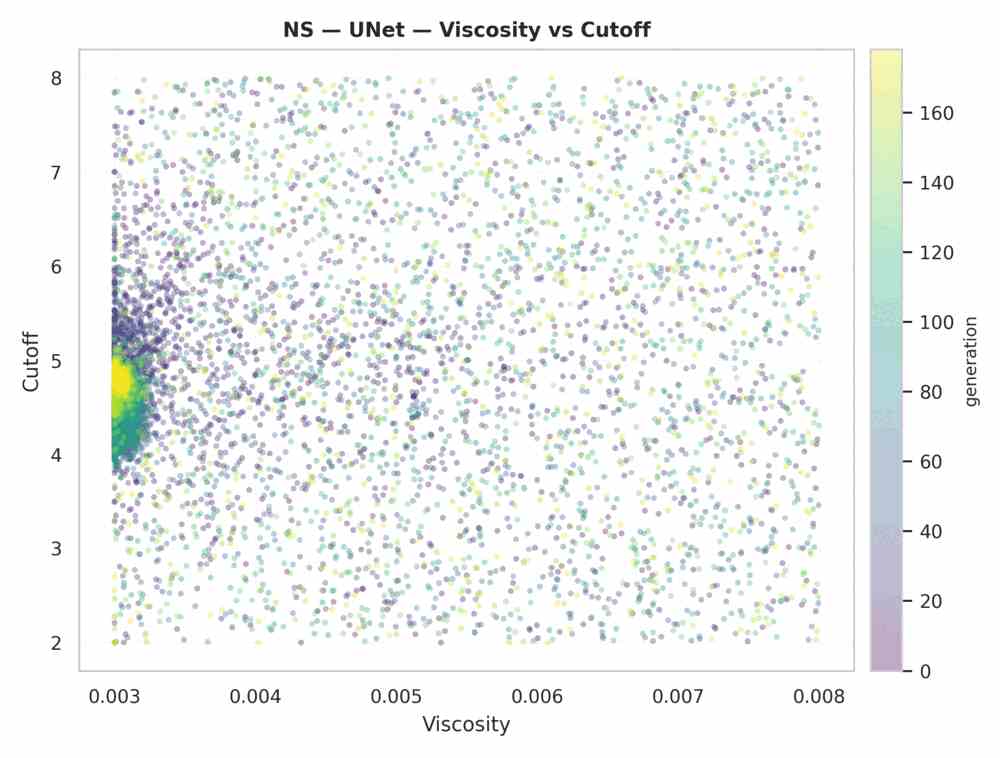}
  \caption{U-Net}
\end{subfigure}
\hfill
\begin{subfigure}{0.32\textwidth}
  \includegraphics[width=\linewidth]{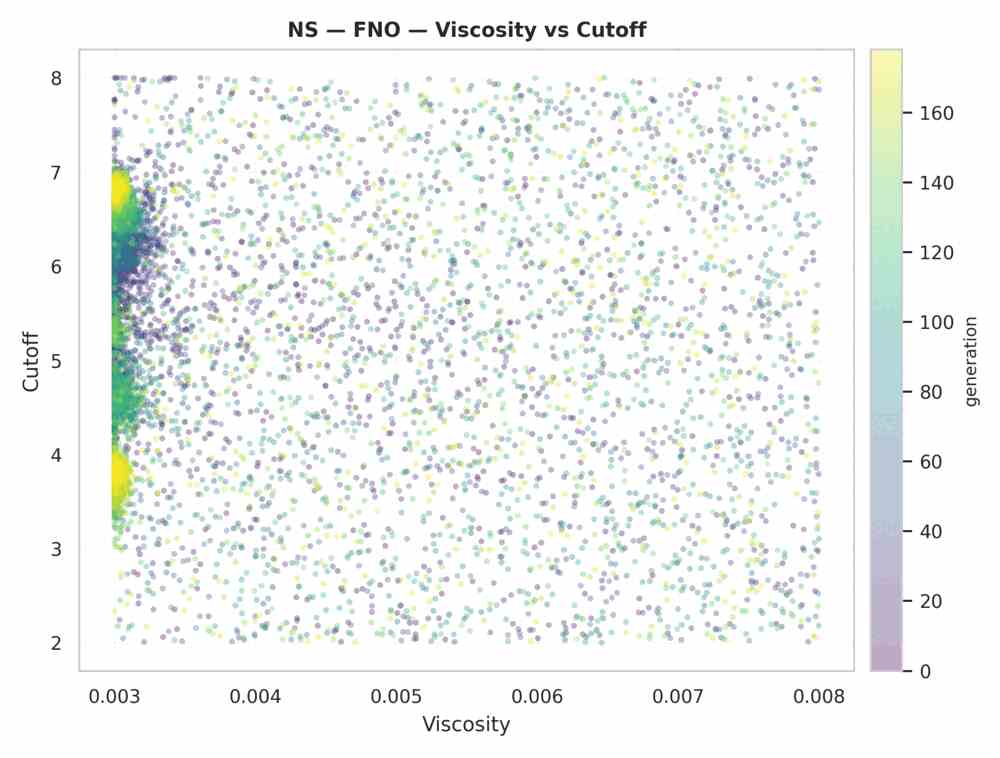}
  \caption{FNO}
\end{subfigure}
\hfill
\begin{subfigure}{0.32\textwidth}
  \includegraphics[width=\linewidth]{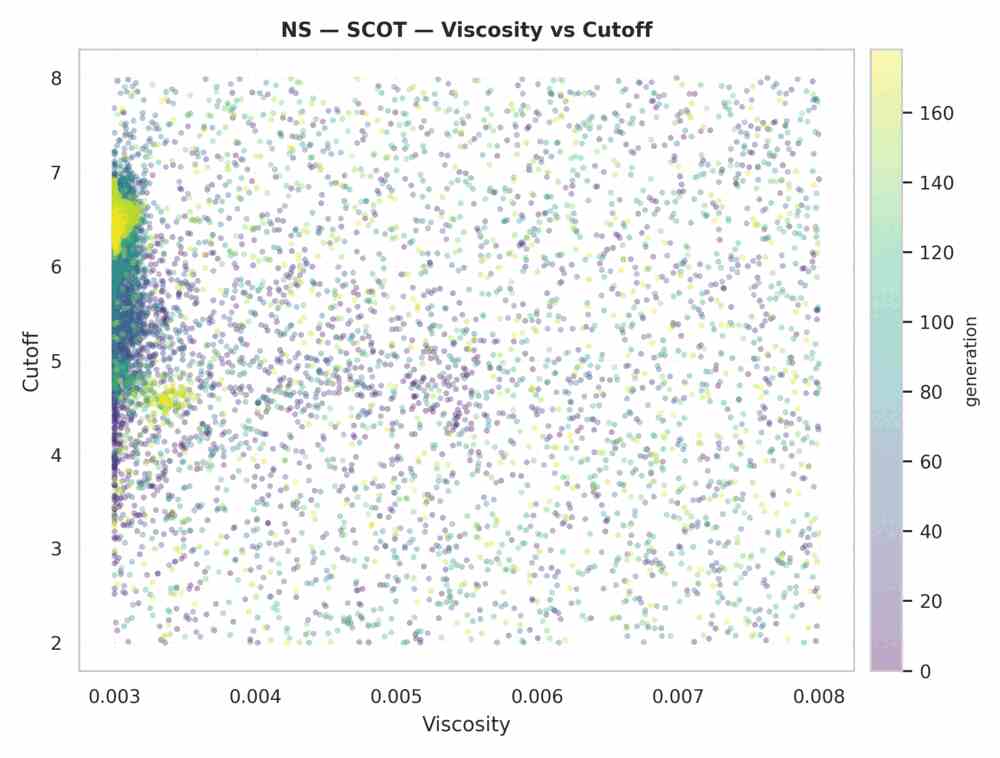}
  \caption{ScOT}
\end{subfigure}
\Description{Three-panel Navier--Stokes plot showing the evolution of sampled diffusivity and spectral cutoff distributions for U-Net, FNO, and scOT.}
\caption{Mixture sampling evolution for Navier-Stokes.}
\label{fig:ns_mixture_evolution}
\end{figure*}

Figures~\ref{fig:gs_mixture_evolution}, \ref{fig:ks_mixture_evolution}, and \ref{fig:ns_mixture_evolution} illustrate the evolution of the sampled parameter distributions across generations for the Gray-Scott, Kuramoto-Sivashinsky, and Navier-Stokes equations, respectively. We visualize the joint distributions of physical coefficients (e.g., viscosity for Navier-Stokes) and key initial condition (IC) parameters, such as the mean position of Gaussian blobs or the spectral cutoff frequency. These plots demonstrate the distribution shift over generations, highlighting the convergence towards specific points or subspaces within the parameter domain.

\subsection{Computational Overhead}
\label{app:gpu_overhead}\label{app:gpu_time_sharing}

\paragraph{OGAS overhead.}
OGAS relies a lightweight generator and discriminator trained asynchronously alongside the surrogate.
The generator is a fast MLP-based diffusion model.  In our experiments, OGAS runs on a standalone GPU, causing
minimal interferences with the main GPUs dedicated to the surrogate training. Measured  wall-time overheads are negligible ($<1\%$) compared to uniform sampling. These overheads are expected to grow moderately when scaling  as OGAS cost is not related to the size of the surrogate model or dimensionality of the training data, but rather on the dimension  of the inferred parameter space and surrogate batch size. 

\paragraph{GPU sharing.}
We  also ran tests where OGAS shares one GPU with the surrogate training,  using 2D PDEs and  the Gray-Scott PDE and scOT models. The experiment trains two ensembles while OGAS uses the last available GPU for DDPM training across two hardware setups on A100 GPUs:

\begin{itemize}
    \item \textbf{Separate GPU configuration:} Ensemble0 on GPU0, Ensemble1 on GPU1, DDPM on GPU2 (3 GPUs total).
    \item \textbf{Time-shared configuration:} Ensemble0 on GPU0, Ensemble1 + DDPM on GPU1 (2 GPUs total).
\end{itemize}

No custom implementation is required for GPU sharing beyond requesting fewer devices from the workload manager; CUDA transparently time-slices and schedules kernels from both processes on the shared device.

\vspace{0.5em}

\begin{center}
    \centering
    \captionof{table}{Performance comparison between dedicated (\textit{Separate}) and shared (\textit{Time-shared}) GPU configurations on A100 GPUs.}
    \label{tab:gpu_sharing_results}
    \vspace{2mm}
    \resizebox{\linewidth}{!}{%
    \begin{tabular}{lcccccc}
        \toprule
        Configuration & GPUs & Walltime (h:m) & Avg. GPU Utilization & Avg. GPU Memory & Total GPU-Hours \\
        \midrule
        Separate    & 3 & 4:40 & 90\%, 90\%, 22\% & 56\%, 56\%, 3\% & $\sim$14.0 \\
        Time-shared & 2 & 6:15 & 68\%, 94\%       & 56\%, 58\%      & $\sim$12.5 \\
        \bottomrule
    \end{tabular}}
    
    \vspace{1.5em} 

    \includegraphics[width=\textwidth]{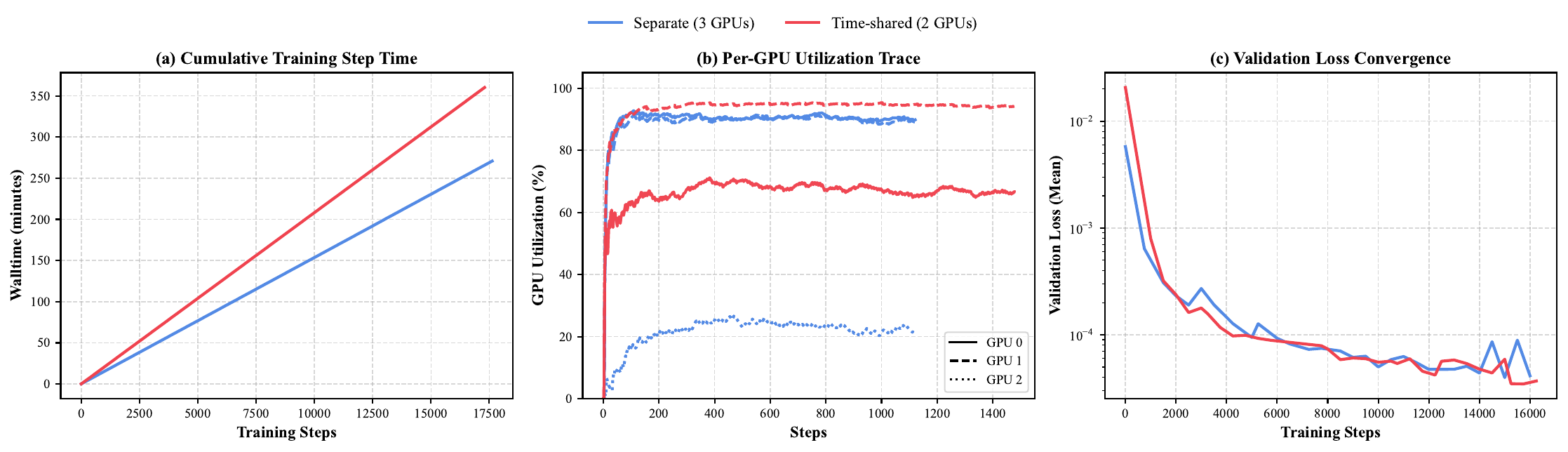}
    \captionof{figure}{GPU sharing performance analysis}
    \label{fig:gpu_sharing_analysis}
\end{center}

\vspace{0.5em}

\begin{itemize}
    \item \textbf{Wall-Clock vs. GPU-Hour Trade-Off:} Time-sharing introduces a $20\text{--}30\%$ wall-clock overhead due to synchronization delays during lockstep ensemble updates. However, it reduces the overall physical GPU requirement from three devices to two and lowers total GPU-hour consumption by approximately two hours (a $\sim10\%$ reduction in total compute cost).
    \item \textbf{Utilization Profile:} Sharing GPU1 increases its average utilization by $4\%$ as the DDPM workload consumes otherwise idle capacity. However, because ensemble updates are strictly synchronized, GPU0 incurs additional idle time while waiting for Ensemble1 to finish its step. This drops GPU0's utilization by $\sim 22\%$, which matches the stand-alone utilization of the DDPM on a dedicated GPU ($22\%$)
    \item \textbf{Model Quality:} Both configurations yielded comparable training convergence and validation metrics, confirming that time-sharing does not degrade output quality. 
\end{itemize}

\paragraph{Cost of pool-based baselines.}
\label{app:pool_based}
Pool-based methods (SBAL, Top-$K$) introduce a substantial inference bottleneck that scales linearly with the pool size $P$. In our setting with $P=5{,}000$, selecting a batch requires rolling out the ensemble on all $5{,}000$ candidates.
\begin{enumerate}[leftmargin=1.2em]
    \item \textit{Inference cost}: Each selection step requires $P \times M \times T_{\text{rollout}}$ forward passes. For $P=5{,}000$, $M=2$, and $T_{\text{rollout}}=15$, this totals $150{,}000$ forward passes per batch selection.
    \item \textit{Throughput impact}: In our online experiment, this scoring step must block the simulation or training process until the next batch is selected. Even with parallelization, the new generation takes at least 3 minutes for each model under our configuration. This explains why we limit the resampling period to 1000 simulations (10 resamplings per experiment) in order to add approx. 17.6 min of delay, which still significantly increases the time overhead compared to OGAS.
\end{enumerate}

\section{1D Benchmark and Ablations}
\label{app:1d_exp}

We conduct extended experiments on 1D PDEs to validate hyperparameter choices in a controlled, lower-dimensional setting before scaling to 2D.

\subsection{Setup}
\label{app:1d_setup}

\paragraph{PDEs and solver parameters.}
We evaluate on three 1D PDEs from APEBench~\cite{koehler-APEBenchBenchmarkAutoregressive-2024}:
\begin{itemize}[leftmargin=1.2em,itemsep=0.1em]
  \item \textit{Kuramoto--Sivashinsky (KS):} A fourth-order chaotic PDE exhibiting spatiotemporal chaos. We use diffusion coefficient $-82.5$, hyperdiffusion $-18{,}750$, and advection $-150$.
  \item \textit{Conservative KS:} A variant using $\partial_x(u^2)$ instead of $u\partial_x u$ for the nonlinear term. Coefficients: $-62.5$, $-12{,}500$, $-5$.
  \item \textit{Korteweg--de Vries (KdV):} A dispersive PDE with coefficients: dispersion $-2{,}500$, hyperdiffusion $-3{,}125$, convection $-10$.
\end{itemize}
Initial conditions are sinusoidal superpositions $u_0(x;\lambda) = \sum_{m=1}^M A_m\sin(mx + \phi_m)$ with $M\in\{2,3\}$ modes, yielding parameter spaces of dimension $\dim(\Lambda)\in\{4,6\}$.

\paragraph{Training and evaluation protocol.}
Table~\ref{tab:1d_setup} summarizes the experimental configuration. We train the U-Net surrogate provided in APEBench (6 hidden channels, 5 levels) using a simulation budget of $N=30{,}000$ trajectories and a batch size of $256$. To ensure reproducibility across runs with variable cluster loads, we employ a speed limiter that maintains a fixed ratio of 2.15 gradient updates per simulated trajectory. Validation is performed on 1{,}500 held-out trajectories.

\begin{table}[h]
\centering
\caption{1D experimental configuration.}
\label{tab:1d_setup}
\small
\begin{tabular}{ll}
\toprule
\textit{Parameter} & \textit{Value} \\
\midrule
\multicolumn{2}{l}{\textit{Solver \& data}} \\
\quad Spatial grid & $N_x = 800$ points \\
\quad Trajectory length & $T = 75$ timesteps \\
\quad Simulation budget & $N = 30{,}000$ trajectories \\
\quad Validation set & $1{,}500$ trajectories (LHS) \\
\midrule
\multicolumn{2}{l}{\textit{Surrogate training}} \\
\quad Architecture & U-Net (6 channels, 5 levels) \\
\quad Solver clients & 56 concurrent processes \\
\quad Speed limiter & 2.15 batches/trajectory \\
\quad Total updates & ${\approx}64{,}500$ \\
\midrule
\multicolumn{2}{l}{\textit{Generator (DDPM)}} \\
\quad Model dimension & $m = 128$ \\
\quad Embedding dimension & $e = 512$ \\
\quad Sampling steps & 50 \\
\quad CFG guidance scale & $w = 5.0$ \\
\quad Max updates per arrival & 10 \\
\quad AL buffer size & 5{,}000 pairs \\
\quad Signal normalization & EMA ($\gamma = 0.9$) \\
\bottomrule
\end{tabular}
\end{table}

\paragraph{Evaluation metrics.}
Following the main experiments (Sec.~\ref{sec:exp_baselines}), we report distributional statistics of the one-step normalized $L^2$ error over the validation set. Note that for these early experiments, we did not compute quantiles or the normalized loss. Thus, we report the mean, max, and std metrics using the standard $L^2$ loss over 5 random seeds. The first two ablations were performed on the KS equation only, and the last study was performed on all three equations (KS, conservative KS, and KdV).

\subsection{Ablation: Buffer Size and Resampling Cadence}
\label{app:1d_buffer}

We first investigate how the surrogate buffer size and resampling frequency affect learning dynamics. Our intuition was that a smaller surrogate buffer would allow for a more accurate distribution of parameters towards the latest resampling, enabling quicker adaptation to the surrogate's evolving difficulty using the most recent data. Larger buffers, however, could provide more stable gradient estimates at the cost of slower reaction to distribution shifts.
 
We sweep buffer sizes across three levels and define capacity in terms of the percentage of trajectories retained from the total budget. The three levels are: small (SB, 150 trajectories $\approx$ 0.5\% of total budget), medium (MB, 750 = 2.5\%), and large  (DB, 1{,}500 = 5\%). We also vary resampling cadences relative to buffer capacity: fast (FR, $0.3\times$ buffer size), normal (NR, $1\times$), and slow (SR, $3\times$). Table~\ref{tab:1d_buffer_ablation} reports results on 1D Kuramoto--Sivashinsky.

\begin{table}[h]
\centering
\caption{Buffer size and resampling cadence ablation on 1D KS. Smaller buffers with faster refresh yield better robustness (lower Max and Std). Best values in bold.}
\label{tab:1d_buffer_ablation}
\small
\begin{tabular}{lccc}
\toprule
\textit{Configuration} & \textit{Max} & \textit{Std} & \textit{Mean} \\
\midrule
\multicolumn{4}{l}{\textit{Large buffer (DB = 1{,}500 trajectories)}} \\
\quad + Slow refresh (SR) & $0.120 \pm 0.057$ & $9.57\times10^{-4}$ & $2.03\times10^{-5}$ \\
\quad + Normal refresh (NR) & $0.101 \pm 0.024$ & $7.48\times10^{-4}$ & $1.47\times10^{-5}$ \\
\quad + Fast refresh (FR) & $0.088 \pm 0.027$ & $6.17\times10^{-4}$ & $1.20\times10^{-5}$ \\
\midrule
\multicolumn{4}{l}{\textit{Medium buffer (MB = 750 trajectories)}} \\
\quad + Slow refresh (SR) & $0.116 \pm 0.049$ & $9.09\times10^{-4}$ & $1.67\times10^{-5}$ \\
\quad + Normal refresh (NR) & $0.089 \pm 0.029$ & $6.75\times10^{-4}$ & $1.21\times10^{-5}$ \\
\quad + Fast refresh (FR) & $0.072 \pm 0.023$ & $4.94\times10^{-4}$ & $9.78\times10^{-6}$ \\
\midrule
\multicolumn{4}{l}{\textit{Small buffer (SB = 150 trajectories)}} \\
\quad + Slow refresh (SR) & $0.089 \pm 0.029$ & $6.48\times10^{-4}$ & $1.45\times10^{-5}$ \\
\quad + Normal refresh (NR) & $0.079 \pm 0.028$ & $5.39\times10^{-4}$ & $1.20\times10^{-5}$ \\
\quad + Fast refresh (FR) & $\mathbf{0.066 \pm 0.015}$ & $\mathbf{3.98\times10^{-4}}$ & $\mathbf{8.03\times10^{-6}}$ \\
\bottomrule
\end{tabular}
\end{table}

These results indicate that the surrogate buffer size is the dominant factor: moving from large (DB) to small (SB) buffers reduces the worst-case error by 45\% and dispersion by 58\%. Within each buffer size, faster resampling consistently improves all metrics. The combination SB+FR achieves the best results across all three metrics with the lowest variance across seeds, confirming that rapid adaptation to the evolving difficulty surface outweighs potential instabilities from smaller sample sizes.

\subsection{Ablation: Tail Targeting and Mixture Rate}
\label{app:1d_tail}
Using the SB+FR configuration, we investigate how aggressively the generator should target high-loss regions. We sample $k(\tilde\eps)$ from the observed training-signal distribution by (i) drawing \textit{uniformly} within a tail slice $[\text{Quantile}_\rho, \max]$, with $\rho \in \{0.5, 0.7, 0.9\}$, and (ii) mixing this with uniform coverage using rate $\alpha \in \{0.15, 0.30, 0.50\}$.

After selecting $\alpha=0.3$ from this ablation, we later introduced \textit{proportional sampling} (initially in 2D experiments): instead of uniform sampling within a fixed tail (requiring $\rho$), we sample $k(\tilde\eps)$ with probability proportional to the observed signal magnitude (Sec.~\ref{sec:method}), while keeping the same mixture mechanism. We report the proportional result at $\alpha=0.3$ for comparison.

\begin{table}[h]
\centering
\caption{Tail quantile and mixture rate ablation on 1D KS}
\label{tab:1d_tail_ablation}
\small
\begin{tabular}{lccc}
\toprule
\textit{Configuration} ($\alpha$, $\rho$) & \textit{Max} & \textit{Std} & \textit{Mean} \\
\midrule
\multicolumn{4}{l}{\textit{High uniform mixture ($\alpha = 0.50$)}} \\
\quad $\rho = 0.7$ & $0.107 \pm 0.045$ & $8.30\times10^{-4}$ & $1.49\times10^{-5}$ \\
\quad $\rho = 0.9$ & $0.094 \pm 0.038$ & $6.92\times10^{-4}$ & $1.32\times10^{-5}$ \\
\midrule
\multicolumn{4}{l}{\textit{Moderate uniform ($\alpha = 0.30$)}} \\
\quad $\rho = 0.7$ & $0.092 \pm 0.025$ & $7.14\times10^{-4}$ & $1.33\times10^{-5}$ \\
\quad $\rho = 0.9$ & $0.082 \pm 0.032$ & $5.87\times10^{-4}$ & $1.13\times10^{-5}$ \\
\quad Proportional & $\mathbf{0.076 \pm 0.028}$ & $\mathbf{5.40\times10^{-4}}$ & $1.19\times10^{-5}$ \\
\midrule
\multicolumn{4}{l}{\textit{Low uniform mixture ($\alpha = 0.15$)}} \\
\quad $\rho = 0.5$ & $0.099 \pm 0.041$ & $7.63\times10^{-4}$ & $1.35\times10^{-5}$ \\
\quad $\rho = 0.7$ & $0.079 \pm 0.030$ & $5.90\times10^{-4}$ & $1.18\times10^{-5}$ \\
\quad $\rho = 0.9$ & $0.076 \pm 0.036$ & $5.64\times10^{-4}$ & $\mathbf{1.10\times10^{-5}}$ \\
\bottomrule
\end{tabular}
\end{table}

We found that targeting higher quantiles ($\rho = 0.9$) consistently improves all metrics across mixture rates, confirming that focusing on the hardest 10\% of observed signals is beneficial. The mixture rate presents a robustness--optimality trade-off: while $\alpha = 0.15$ attains the best tail-uniform \textit{Max} values, it exhibits higher seed-to-seed variability. The moderate setting $\alpha = 0.30$ with $\rho = 0.9$ offers near-optimal performance with better stability, motivating our choice of $\alpha=0.3$ for subsequent experiments. Proportional sampling at $\alpha=0.3$ gives similar or better results (lower \textit{Max} and \textit{Std} in Table~\ref{tab:1d_tail_ablation}) without tuning $\rho$, so we adopt it.

\textit{Adopted configuration:} $\alpha = 0.30$, with proportional sampling.

\subsection{Cross-PDE Comparison and Effect of Debiasing}
\label{app:1d_cross_pde}
 
Using the configurations selected above (SB+FR, $\alpha = 0.30$), we compare OGAS against baselines across all three 1D PDEs. We also evaluate the impact of density-ratio debiasing by comparing three OGAS variants: without debiasing, with random reference samples only, and with the full classifier-based ratio estimation.

\begin{table}[h]
\centering
\caption{Summary across PDEs and sampling strategies (reported as mean $\pm$ half-range).
 OGAS = main method with ratio debiasing, OGAS-NC = no ratio debiasing, OGAS-R = no ratio debiasing but trained only on uniformly sampled lambdas.
Best (lowest) per PDE and metric is in \textbf{bold}.}
\label{tab:all_pdes_one_table_boldbest}
\resizebox{\linewidth}{!}{%
\begin{tabular}{llccc}
\toprule
\textit{PDE} & \textit{Strategy} & \textit{Mean } & \textit{Max } & \textit{Std } \\
\midrule

KS & Uniform & 4.297e-05 $\pm$ 1.464e-05 & 1.784e-01 $\pm$ 5.716e-02 & 1.529e-03 $\pm$ 7.587e-04 \\
KS & Breed   & 2.365e-05 $\pm$ 1.004e-05 & 1.422e-01 $\pm$ 5.990e-02 & 1.129e-03 $\pm$ 6.433e-04 \\
KS & OGAS-NC & 1.408e-05 $\pm$ 5.786e-06 & 9.650e-02 $\pm$ 3.730e-02 & 7.437e-04 $\pm$ 3.567e-04 \\
KS & OGAS-R  & 2.642e-05 $\pm$ 3.971e-06 & 1.360e-01 $\pm$ 3.209e-02 & 1.019e-03 $\pm$ 4.161e-04 \\
KS & OGAS    & \textbf{1.154e-05 $\pm$ 3.750e-06} & \textbf{8.082e-02 $\pm$ 2.172e-02} & \textbf{5.879e-04 $\pm$ 2.426e-04} \\
\midrule

KdV & Uniform & 2.423e-07 $\pm$ 1.152e-07 & 9.703e-05 $\pm$ 2.170e-04 & 1.503e-06 $\pm$ 3.182e-06 \\
KdV & Breed   & \textbf{2.091e-07 $\pm$ 6.997e-08} & 1.673e-06 $\pm$ 7.005e-07 & 1.343e-07 $\pm$ 3.474e-08 \\
KdV & OGAS-NC & 2.419e-07 $\pm$ 1.627e-07 & \textbf{1.530e-06 $\pm$ 6.596e-07} & 1.379e-07 $\pm$ 5.827e-08 \\
KdV & OGAS-R  & 2.426e-07 $\pm$ 1.374e-07 & 1.622e-06 $\pm$ 7.050e-07 & \textbf{1.232e-07 $\pm$ 4.375e-08} \\
KdV & OGAS    & 2.367e-07 $\pm$ 1.155e-07 & 1.828e-06 $\pm$ 7.517e-07 & 1.447e-07 $\pm$ 4.021e-08 \\
\midrule

KS (cons.) & Uniform & 6.078e-06 $\pm$ 1.292e-06 & 5.020e-03 $\pm$ 1.045e-04 & 5.160e-05 $\pm$ 1.569e-06 \\
KS (cons.) & Breed   & 5.180e-06 $\pm$ 1.199e-06 & 4.186e-03 $\pm$ 1.325e-04 & 4.046e-05 $\pm$ 1.509e-06 \\
KS (cons.) & OGAS-NC & 4.882e-06 $\pm$ 1.151e-06 & 2.965e-03 $\pm$ 8.037e-04 & 2.244e-05 $\pm$ 7.773e-06 \\
KS (cons.) & OGAS-R  & 5.893e-06 $\pm$ 1.420e-06 & 4.911e-03 $\pm$ 3.141e-04 & 4.876e-05 $\pm$ 4.682e-06 \\
KS (cons.) & OGAS    & \textbf{4.866e-06 $\pm$ 8.904e-07} & \textbf{2.108e-03 $\pm$ 1.441e-03} & \textbf{1.712e-05 $\pm$ 1.072e-05} \\
\bottomrule
\end{tabular}}
\end{table}

 OGAS with classifier-based density-ratio debiasing achieves 2--3$\times$ lower dispersion than uniform sampling and outperforms Breed-style importance sampling by 40--50\% across all PDEs. The improvement is consistent across equation types with varying dynamical regimes (chaotic, conservative, dispersive) and removing debiasing increases dispersion by 20\% relative to the full method, confirming the importance of correcting for buffer-induced sampling bias.

\section{Complete Results}
\label{app:complete_results}
\subsection{2D Benchmark Results}
\label{app:2d_results}
We provide full tabular results for the 2D experiments across three PDEs and three architectures, complementing the summary plots in the main text. Metrics reported are Mean, Standard Deviation (Std), Maximum (Max), and tail percentiles (P99, P95, P75, P50) of the normalized one-step error on the validation set.

\begin{table}[h]
\centering
\footnotesize
\caption{\textbf{2D Navier-Stokes Results.} Active learning strategy comparison on Kolmogorov flow. Values show mean $\pm$ std across seeds. Best per architecture in \textbf{bold}.}
\label{tab:navier-stokes-kolmogorov-flow-2d-low-res}
\begin{subtable}{\linewidth}
\centering
\caption{UNet}
\resizebox{\linewidth}{!}{%
\begin{tabular}{lccccccc}
\toprule
Strategy & Mean ($\times 10^{-3}$) & Std ($\times 10^{-3}$) & Max ($\times 10^{-2}$) & P99 ($\times 10^{-2}$) & P95 ($\times 10^{-2}$) & P75 ($\times 10^{-2}$) & P50 ($\times 10^{-3}$) \\
\midrule
OGAS-L & $8.93 \pm 0.53$ & $2.73 \pm 0.07$ & $4.16 \pm 0.79$ & $1.80 \pm 0.06$ & $1.44 \pm 0.07$ & $\mathbf{1.02 \pm 0.06}$ & $8.22 \pm 0.54$ \\
OGAS-U & $8.91 \pm 0.16$ & $\mathbf{2.72 \pm 0.10}$ & $\mathbf{3.72 \pm 0.49}$ & $\mathbf{1.78 \pm 0.05}$ & $\mathbf{1.42 \pm 0.02}$ & $1.02 \pm 0.01$ & $8.26 \pm 0.16$ \\
Breed & $\mathbf{8.88 \pm 0.36}$ & $3.24 \pm 0.18$ & $4.82 \pm 0.10$ & $1.99 \pm 0.09$ & $1.52 \pm 0.08$ & $1.03 \pm 0.05$ & $7.95 \pm 0.28$ \\
Uniform & $9.27 \pm 0.20$ & $4.88 \pm 0.07$ & $6.16 \pm 0.10$ & $2.56 \pm 0.03$ & $1.89 \pm 0.03$ & $1.14 \pm 0.02$ & $7.67 \pm 0.22$ \\
SBAL & $9.38 \pm 0.39$ & $5.01 \pm 0.22$ & $6.12 \pm 0.37$ & $2.60 \pm 0.11$ & $1.93 \pm 0.08$ & $1.17 \pm 0.05$ & $7.74 \pm 0.34$ \\
Top-K & $9.15 \pm 0.02$ & $4.85 \pm 0.06$ & $6.59 \pm 0.07$ & $2.52 \pm 0.02$ & $1.87 \pm 0.02$ & $1.13 \pm 0.00$ & $\mathbf{7.56 \pm 0.04}$ \\
Sobol & $9.30 \pm 0.06$ & $4.90 \pm 0.04$ & $6.09 \pm 0.24$ & $2.56 \pm 0.01$ & $1.90 \pm 0.00$ & $1.15 \pm 0.01$ & $7.70 \pm 0.06$ \\
\bottomrule
\end{tabular}}
\end{subtable}
\vspace{0.5em}
\begin{subtable}{\linewidth}
\centering
\caption{FNO}
\resizebox{\linewidth}{!}{%
\begin{tabular}{lccccccc}
\toprule
Strategy & Mean ($\times 10^{-2}$) & Std ($\times 10^{-3}$) & Max ($\times 10^{-2}$) & P99 ($\times 10^{-2}$) & P95 ($\times 10^{-2}$) & P75 ($\times 10^{-2}$) & P50 ($\times 10^{-3}$) \\
\midrule
OGAS-L & $1.04 \pm 0.01$ & $\mathbf{3.42 \pm 0.12}$ & $4.19 \pm 0.14$ & $2.22 \pm 0.04$ & $1.71 \pm 0.03$ & $1.19 \pm 0.01$ & $9.59 \pm 0.17$ \\
OGAS-U & $1.03 \pm 0.02$ & $3.44 \pm 0.10$ & $\mathbf{4.09 \pm 0.18}$ & $\mathbf{2.22 \pm 0.05}$ & $\mathbf{1.70 \pm 0.04}$ & $1.17 \pm 0.03$ & $9.45 \pm 0.22$ \\
Breed & $1.23 \pm 0.03$ & $4.78 \pm 0.24$ & $5.60 \pm 0.30$ & $2.86 \pm 0.11$ & $2.17 \pm 0.08$ & $1.44 \pm 0.05$ & $11.03 \pm 0.28$ \\
Uniform & $1.04 \pm 0.02$ & $5.99 \pm 0.12$ & $5.84 \pm 0.11$ & $3.02 \pm 0.05$ & $2.24 \pm 0.05$ & $1.32 \pm 0.03$ & $8.52 \pm 0.20$ \\
SBAL & $1.02 \pm 0.01$ & $5.74 \pm 0.18$ & $5.70 \pm 0.13$ & $2.93 \pm 0.06$ & $2.17 \pm 0.05$ & $1.28 \pm 0.03$ & $8.40 \pm 0.10$ \\
Top-K & $\mathbf{0.98 \pm 0.02}$ & $4.58 \pm 0.24$ & $5.03 \pm 0.12$ & $2.55 \pm 0.08$ & $1.89 \pm 0.06$ & $\mathbf{1.17 \pm 0.03}$ & $8.46 \pm 0.19$ \\
Sobol & $1.02 \pm 0.02$ & $5.88 \pm 0.10$ & $5.66 \pm 0.10$ & $2.98 \pm 0.05$ & $2.20 \pm 0.03$ & $1.29 \pm 0.02$ & $\mathbf{8.38 \pm 0.15}$ \\
\bottomrule
\end{tabular}}
\end{subtable}
\vspace{0.5em}
\begin{subtable}{\linewidth}
\centering
\caption{scOT}
\resizebox{\linewidth}{!}{%
\begin{tabular}{lccccccc}
\toprule
Strategy & Mean ($\times 10^{-2}$) & Std ($\times 10^{-3}$) & Max ($\times 10^{-2}$) & P99 ($\times 10^{-2}$) & P95 ($\times 10^{-2}$) & P75 ($\times 10^{-2}$) & P50 ($\times 10^{-2}$) \\
\midrule
OGAS-L & $\mathbf{1.72 \pm 0.19}$ & $\mathbf{4.42 \pm 0.23}$ & $\mathbf{5.45 \pm 0.34}$ & $\mathbf{3.08 \pm 0.25}$ & $\mathbf{2.55 \pm 0.24}$ & $\mathbf{1.96 \pm 0.21}$ & $1.63 \pm 0.19$ \\
OGAS-U & $1.86 \pm 0.04$ & $4.75 \pm 0.13$ & $5.75 \pm 0.14$ & $3.31 \pm 0.08$ & $2.76 \pm 0.06$ & $2.11 \pm 0.05$ & $1.76 \pm 0.04$ \\
Breed & $2.15 \pm 0.21$ & $7.96 \pm 1.07$ & $10.10 \pm 1.44$ & $4.87 \pm 0.58$ & $3.64 \pm 0.40$ & $2.49 \pm 0.25$ & $1.95 \pm 0.19$ \\
Uniform & $1.91 \pm 0.22$ & $6.54 \pm 0.47$ & $8.83 \pm 0.99$ & $4.03 \pm 0.33$ & $3.14 \pm 0.29$ & $2.23 \pm 0.24$ & $1.76 \pm 0.21$ \\
SBAL & $1.82 \pm 0.18$ & $6.30 \pm 0.29$ & $8.32 \pm 0.34$ & $3.86 \pm 0.23$ & $3.01 \pm 0.22$ & $2.12 \pm 0.19$ & $1.66 \pm 0.17$ \\
Top-K & $1.75 \pm 0.16$ & $6.19 \pm 0.61$ & $7.97 \pm 0.76$ & $3.75 \pm 0.34$ & $2.92 \pm 0.27$ & $2.06 \pm 0.18$ & $\mathbf{1.61 \pm 0.14}$ \\
Sobol & $1.77 \pm 0.18$ & $6.35 \pm 0.16$ & $8.27 \pm 0.26$ & $3.85 \pm 0.21$ & $2.96 \pm 0.21$ & $2.07 \pm 0.20$ & $1.61 \pm 0.19$ \\
\bottomrule
\end{tabular}}
\end{subtable}
\end{table}

\begin{table}[h]
\centering
\footnotesize
\caption{\textbf{2D Kuramoto-Sivashinsky Results.} Active learning strategy comparison. Values show mean $\pm$ std across seeds. Best per architecture in \textbf{bold}.}
\label{tab:kuramoto-sivashinsky-2d-low-res}
\begin{subtable}{\linewidth}
\centering
\caption{UNet}
\resizebox{\linewidth}{!}{%
\begin{tabular}{lccccccc}
\toprule
Strategy & Mean ($\times 10^{-3}$) & Std ($\times 10^{-3}$) & Max ($\times 10^{-2}$) & P99 ($\times 10^{-2}$) & P95 ($\times 10^{-3}$) & P75 ($\times 10^{-3}$) & P50 ($\times 10^{-3}$) \\
\midrule
OGAS-L & $5.11 \pm 0.41$ & $\mathbf{1.32 \pm 0.05}$ & $5.08 \pm 0.41$ & $\mathbf{0.87 \pm 0.05}$ & $6.99 \pm 0.42$ & $5.64 \pm 0.39$ & $4.96 \pm 0.40$ \\
OGAS-U & $4.86 \pm 0.46$ & $1.37 \pm 0.13$ & $6.07 \pm 0.81$ & $0.94 \pm 0.12$ & $\mathbf{6.70 \pm 0.59}$ & $5.21 \pm 0.42$ & $4.67 \pm 0.39$ \\
Breed & $4.91 \pm 0.46$ & $1.54 \pm 0.21$ & $\mathbf{4.50 \pm 0.95}$ & $1.10 \pm 0.16$ & $7.07 \pm 1.00$ & $5.16 \pm 0.50$ & $4.59 \pm 0.40$ \\
Uniform & $\mathbf{4.28 \pm 0.06}$ & $2.61 \pm 0.14$ & $9.53 \pm 0.07$ & $1.47 \pm 0.06$ & $7.51 \pm 0.32$ & $\mathbf{4.44 \pm 0.07}$ & $3.66 \pm 0.09$ \\
SBAL & $4.44 \pm 0.17$ & $3.24 \pm 0.34$ & $10.56 \pm 0.15$ & $1.83 \pm 0.10$ & $8.57 \pm 0.60$ & $4.51 \pm 0.15$ & $3.62 \pm 0.13$ \\
Top-K & $4.41 \pm 0.06$ & $3.37 \pm 0.16$ & $10.15 \pm 0.66$ & $1.86 \pm 0.12$ & $8.87 \pm 0.03$ & $4.57 \pm 0.15$ & $\mathbf{3.51 \pm 0.07}$ \\
Sobol & $4.43 \pm 0.02$ & $3.01 \pm 0.18$ & $11.14 \pm 1.68$ & $1.59 \pm 0.03$ & $7.79 \pm 0.24$ & $4.55 \pm 0.08$ & $3.73 \pm 0.03$ \\
\bottomrule
\end{tabular}}
\end{subtable}
\vspace{0.5em}
\begin{subtable}{\linewidth}
\centering
\caption{FNO}
\resizebox{\linewidth}{!}{%
\begin{tabular}{lccccccc}
\toprule
Strategy & Mean ($\times 10^{-3}$) & Std ($\times 10^{-3}$) & Max ($\times 10^{-2}$) & P99 ($\times 10^{-2}$) & P95 ($\times 10^{-2}$) & P75 ($\times 10^{-2}$) & P50 ($\times 10^{-3}$) \\
\midrule
OGAS-L & $9.49 \pm 0.28$ & $\mathbf{1.76 \pm 0.06}$ & $\mathbf{3.26 \pm 0.18}$ & $\mathbf{1.45 \pm 0.03}$ & $\mathbf{1.22 \pm 0.04}$ & $1.02 \pm 0.03$ & $9.28 \pm 0.29$ \\
OGAS-U & $8.63 \pm 0.02$ & $2.73 \pm 0.16$ & $4.88 \pm 0.28$ & $2.04 \pm 0.09$ & $1.29 \pm 0.04$ & $0.93 \pm 0.01$ & $7.97 \pm 0.10$ \\
Breed & $11.32 \pm 0.35$ & $2.53 \pm 0.41$ & $5.02 \pm 0.58$ & $2.18 \pm 0.26$ & $1.61 \pm 0.07$ & $1.17 \pm 0.04$ & $10.65 \pm 0.42$ \\
Uniform & $\mathbf{8.30 \pm 0.32}$ & $4.02 \pm 0.16$ & $5.95 \pm 0.12$ & $2.79 \pm 0.07$ & $1.53 \pm 0.06$ & $\mathbf{0.87 \pm 0.04}$ & $\mathbf{6.93 \pm 0.25}$ \\
SBAL & $9.26 \pm 0.36$ & $3.23 \pm 0.42$ & $3.34 \pm 0.11$ & $1.80 \pm 0.12$ & $1.57 \pm 0.12$ & $1.11 \pm 0.08$ & $8.18 \pm 0.27$ \\
Top-K & $9.84 \pm 0.16$ & $2.66 \pm 0.36$ & $3.49 \pm 0.23$ & $1.72 \pm 0.05$ & $1.50 \pm 0.04$ & $1.10 \pm 0.08$ & $9.41 \pm 0.26$ \\
Sobol & $8.56 \pm 0.47$ & $4.11 \pm 0.12$ & $6.46 \pm 0.16$ & $2.86 \pm 0.12$ & $1.56 \pm 0.05$ & $0.89 \pm 0.05$ & $7.20 \pm 0.46$ \\
\bottomrule
\end{tabular}}
\end{subtable}
\vspace{0.5em}
\begin{subtable}{\linewidth}
\centering
\caption{scOT}
\resizebox{\linewidth}{!}{%
\begin{tabular}{lccccccc}
\toprule
Strategy & Mean ($\times 10^{-2}$) & Std ($\times 10^{-3}$) & Max ($\times 10^{-2}$) & P99 ($\times 10^{-2}$) & P95 ($\times 10^{-2}$) & P75 ($\times 10^{-2}$) & P50 ($\times 10^{-2}$) \\
\midrule
OGAS-L & $1.22 \pm 0.12$ & $\mathbf{1.92 \pm 0.25}$ & $3.20 \pm 0.23$ & $\mathbf{1.71 \pm 0.15}$ & $\mathbf{1.51 \pm 0.16}$ & $1.31 \pm 0.14$ & $1.21 \pm 0.12$ \\
OGAS-U & $1.30 \pm 0.15$ & $2.16 \pm 0.35$ & $\mathbf{3.19 \pm 0.28}$ & $1.82 \pm 0.23$ & $1.63 \pm 0.22$ & $1.40 \pm 0.17$ & $1.28 \pm 0.15$ \\
Breed & $1.43 \pm 0.15$ & $5.09 \pm 1.31$ & $8.86 \pm 1.20$ & $3.55 \pm 0.67$ & $2.37 \pm 0.53$ & $1.48 \pm 0.13$ & $1.27 \pm 0.09$ \\
Uniform & $1.16 \pm 0.11$ & $2.82 \pm 0.95$ & $4.80 \pm 0.39$ & $2.32 \pm 0.58$ & $1.66 \pm 0.35$ & $1.23 \pm 0.10$ & $1.09 \pm 0.07$ \\
SBAL & $1.23 \pm 0.11$ & $3.46 \pm 1.70$ & $5.78 \pm 1.98$ & $2.72 \pm 0.86$ & $1.85 \pm 0.46$ & $1.30 \pm 0.10$ & $1.14 \pm 0.04$ \\
Top-K & $1.20 \pm 0.17$ & $5.02 \pm 2.12$ & $7.47 \pm 2.19$ & $3.30 \pm 1.12$ & $2.09 \pm 0.61$ & $1.33 \pm 0.20$ & $1.06 \pm 0.07$ \\
Sobol & $\mathbf{1.11 \pm 0.11}$ & $3.05 \pm 0.52$ & $5.17 \pm 0.82$ & $2.42 \pm 0.34$ & $1.64 \pm 0.24$ & $\mathbf{1.18 \pm 0.10}$ & $\mathbf{1.03 \pm 0.09}$ \\
\bottomrule
\end{tabular}}
\end{subtable}
\end{table}

\begin{table}[h]
\centering
\footnotesize
\caption{\textbf{2D Gray-Scott Results.} Active learning strategy comparison. Values show mean $\pm$ std across seeds. Best per architecture in \textbf{bold}.}
\label{tab:gray-scott-beta-low-res}
\begin{subtable}{\linewidth}
\centering
\caption{UNet}
\resizebox{\linewidth}{!}{%
\begin{tabular}{lccccccc}
\toprule
Strategy & Mean ($\times 10^{-2}$) & Std ($\times 10^{-2}$) & Max ($\times 10^{-1}$) & P99 ($\times 10^{-2}$) & P95 ($\times 10^{-2}$) & P75 ($\times 10^{-2}$) & P50 ($\times 10^{-3}$) \\
\midrule
OGAS-L & $1.37 \pm 0.04$ & $0.83 \pm 0.07$ & $\mathbf{1.00 \pm 0.39}$ & $4.35 \pm 0.50$ & $3.16 \pm 0.09$ & $1.65 \pm 0.10$ & $10.91 \pm 0.55$ \\
OGAS-U & $1.35 \pm 0.23$ & $1.02 \pm 0.71$ & $1.08 \pm 0.06$ & $4.89 \pm 3.55$ & $4.18 \pm 2.98$ & $1.48 \pm 0.03$ & $10.16 \pm 0.18$ \\
Breed & $\mathbf{1.13 \pm 0.08}$ & $\mathbf{0.47 \pm 0.02}$ & $1.52 \pm 0.51$ & $\mathbf{2.42 \pm 0.00}$ & $\mathbf{1.89 \pm 0.12}$ & $\mathbf{1.34 \pm 0.11}$ & $10.05 \pm 0.80$ \\
Uniform & $1.27 \pm 0.23$ & $1.14 \pm 0.41$ & $1.78 \pm 0.26$ & $5.38 \pm 2.17$ & $4.05 \pm 1.49$ & $1.55 \pm 0.29$ & $8.41 \pm 0.67$ \\
SBAL & $1.36 \pm 0.25$ & $1.29 \pm 0.49$ & $1.77 \pm 0.19$ & $6.12 \pm 2.81$ & $4.50 \pm 1.86$ & $1.65 \pm 0.13$ & $8.61 \pm 0.85$ \\
Top-K & $1.22 \pm 0.13$ & $1.06 \pm 0.28$ & $1.67 \pm 0.07$ & $4.81 \pm 1.56$ & $3.81 \pm 1.08$ & $1.50 \pm 0.19$ & $\mathbf{7.94 \pm 0.11}$ \\
Sobol & $1.42 \pm 0.03$ & $1.43 \pm 0.22$ & $2.02 \pm 0.39$ & $6.69 \pm 1.15$ & $5.21 \pm 0.91$ & $1.71 \pm 0.14$ & $8.49 \pm 1.13$ \\
\bottomrule
\end{tabular}}
\end{subtable}
\vspace{0.5em}
\begin{subtable}{\linewidth}
\centering
\caption{FNO}
\resizebox{\linewidth}{!}{%
\begin{tabular}{lccccccc}
\toprule
Strategy & Mean ($\times 10^{-2}$) & Std ($\times 10^{-3}$) & Max ($\times 10^{-2}$) & P99 ($\times 10^{-2}$) & P95 ($\times 10^{-2}$) & P75 ($\times 10^{-2}$) & P50 ($\times 10^{-2}$) \\
\midrule
OGAS-L & $1.41 \pm 0.01$ & $\mathbf{6.21 \pm 0.17}$ & $\mathbf{7.36 \pm 0.43}$ & $\mathbf{3.21 \pm 0.03}$ & $\mathbf{2.71 \pm 0.03}$ & $1.69 \pm 0.01$ & $1.25 \pm 0.02$ \\
OGAS-U & $1.42 \pm 0.02$ & $7.06 \pm 0.35$ & $8.11 \pm 0.26$ & $3.58 \pm 0.14$ & $2.99 \pm 0.12$ & $1.67 \pm 0.02$ & $1.23 \pm 0.01$ \\
Breed & $1.47 \pm 0.06$ & $7.01 \pm 0.03$ & $9.07 \pm 0.24$ & $3.71 \pm 0.17$ & $2.81 \pm 0.05$ & $1.82 \pm 0.06$ & $1.24 \pm 0.05$ \\
Uniform & $\mathbf{1.29 \pm 0.03}$ & $8.49 \pm 0.33$ & $10.22 \pm 0.70$ & $3.90 \pm 0.13$ & $3.23 \pm 0.11$ & $\mathbf{1.55 \pm 0.03}$ & $\mathbf{1.00 \pm 0.01}$ \\
SBAL & $1.31 \pm 0.01$ & $8.40 \pm 0.02$ & $9.74 \pm 1.76$ & $3.89 \pm 0.01$ & $3.21 \pm 0.01$ & $1.56 \pm 0.02$ & $1.01 \pm 0.01$ \\
Top-K & $1.35 \pm 0.03$ & $8.70 \pm 0.15$ & $10.08 \pm 0.75$ & $4.01 \pm 0.06$ & $3.32 \pm 0.05$ & $1.61 \pm 0.04$ & $1.04 \pm 0.05$ \\
Sobol & $1.35 \pm 0.08$ & $8.63 \pm 0.30$ & $9.24 \pm 0.82$ & $3.99 \pm 0.15$ & $3.29 \pm 0.13$ & $1.63 \pm 0.11$ & $1.06 \pm 0.08$ \\
\bottomrule
\end{tabular}}
\end{subtable}
\vspace{0.5em}
\begin{subtable}{\linewidth}
\centering
\caption{scOT}
\resizebox{\linewidth}{!}{%
\begin{tabular}{lccccccc}
\toprule
Strategy & Mean ($\times 10^{-2}$) & Std ($\times 10^{-3}$) & Max ($\times 10^{-2}$) & P99 ($\times 10^{-2}$) & P95 ($\times 10^{-2}$) & P75 ($\times 10^{-2}$) & P50 ($\times 10^{-2}$) \\
\midrule
OGAS-L & $\mathbf{1.44 \pm 0.07}$ & $6.13 \pm 0.40$ & $\mathbf{6.41 \pm 1.57}$ & $\mathbf{3.25 \pm 0.17}$ & $2.77 \pm 0.13$ & $\mathbf{1.66 \pm 0.10}$ & $1.26 \pm 0.06$ \\
OGAS-U & $1.53 \pm 0.05$ & $7.32 \pm 0.47$ & $7.47 \pm 1.04$ & $3.80 \pm 0.31$ & $3.10 \pm 0.10$ & $1.80 \pm 0.08$ & $1.30 \pm 0.03$ \\
Breed & $1.44 \pm 0.16$ & $\mathbf{5.73 \pm 1.71}$ & $8.84 \pm 2.41$ & $3.47 \pm 1.13$ & $\mathbf{2.54 \pm 0.53}$ & $1.66 \pm 0.16$ & $1.26 \pm 0.09$ \\
Uniform & $1.48 \pm 0.14$ & $8.80 \pm 0.66$ & $9.98 \pm 2.53$ & $4.18 \pm 0.54$ & $3.28 \pm 0.18$ & $1.84 \pm 0.14$ & $1.16 \pm 0.12$ \\
SBAL & $1.46 \pm 0.03$ & $9.07 \pm 0.05$ & $10.13 \pm 1.59$ & $4.29 \pm 0.11$ & $3.31 \pm 0.03$ & $1.82 \pm 0.06$ & $\mathbf{1.13 \pm 0.03}$ \\
Top-K & $1.47 \pm 0.09$ & $9.02 \pm 1.09$ & $10.71 \pm 2.03$ & $4.28 \pm 0.69$ & $3.30 \pm 0.18$ & $1.83 \pm 0.12$ & $1.15 \pm 0.06$ \\
Sobol & $1.46 \pm 0.04$ & $8.89 \pm 0.78$ & $9.63 \pm 1.44$ & $4.12 \pm 0.41$ & $3.29 \pm 0.12$ & $1.83 \pm 0.08$ & $1.14 \pm 0.02$ \\
\bottomrule
\end{tabular}}
\end{subtable}
\end{table}

\subsection{Single Model Results}
\label{app:single_model_tables}
Single-model counterparts of the tables above, discussed in Appendix~\ref{app:single_model_eval}.

\begin{table}[h]
\centering
\footnotesize
\caption{\textbf{Single-Model Navier-Stokes Results.} Evaluation is performed one single model trained or on the first model of the ensemble when ensemble is required for uncertainty computation. Values show mean $\pm$ std across seeds. Best per architecture in \textbf{bold}.}
\label{tab:navier-stokes-single}
\begin{subtable}{\linewidth}
\centering
\caption{UNet}
\resizebox{\linewidth}{!}{%
\begin{tabular}{lccccccc}
\toprule
Strategy & Mean ($\times 10^{-3}$) & Std ($\times 10^{-3}$) & Max ($\times 10^{-2}$) & P99 ($\times 10^{-2}$) & P95 ($\times 10^{-2}$) & P75 ($\times 10^{-2}$) & P50 ($\times 10^{-3}$) \\
\midrule
OGAS-L & $11.22 \pm 0.64$ & $3.09 \pm 0.07$ & $4.71 \pm 0.36$ & $2.13 \pm 0.06$ & $1.72 \pm 0.06$ & $1.27 \pm 0.05$ & $10.51 \pm 0.64$ \\
OGAS-L2 & $8.93 \pm 0.53$ & $2.73 \pm 0.07$ & $4.16 \pm 0.79$ & $1.80 \pm 0.06$ & $1.44 \pm 0.07$ & $\mathbf{1.02 \pm 0.06}$ & $8.22 \pm 0.54$ \\
OGAS-U & $\mathbf{8.91 \pm 0.16}$ & $\mathbf{2.72 \pm 0.10}$ & $\mathbf{3.72 \pm 0.49}$ & $\mathbf{1.78 \pm 0.05}$ & $\mathbf{1.42 \pm 0.02}$ & $1.02 \pm 0.01$ & $8.26 \pm 0.16$ \\
SBAL & $9.38 \pm 0.39$ & $5.01 \pm 0.22$ & $6.12 \pm 0.37$ & $2.60 \pm 0.11$ & $1.93 \pm 0.08$ & $1.17 \pm 0.05$ & $7.74 \pm 0.34$ \\
Top-K & $9.15 \pm 0.02$ & $4.85 \pm 0.06$ & $6.59 \pm 0.07$ & $2.52 \pm 0.02$ & $1.87 \pm 0.02$ & $1.13 \pm 0.00$ & $\mathbf{7.56 \pm 0.04}$ \\
Sobol & $9.30 \pm 0.06$ & $4.90 \pm 0.04$ & $6.09 \pm 0.24$ & $2.56 \pm 0.01$ & $1.90 \pm 0.00$ & $1.15 \pm 0.01$ & $7.70 \pm 0.06$ \\
Uniform & $9.27 \pm 0.20$ & $4.88 \pm 0.07$ & $6.16 \pm 0.10$ & $2.56 \pm 0.03$ & $1.89 \pm 0.03$ & $1.14 \pm 0.02$ & $7.67 \pm 0.22$ \\
\bottomrule
\end{tabular}}
\end{subtable}
\vspace{0.5em}
\begin{subtable}{\linewidth}
\centering
\caption{FNO}
\resizebox{\linewidth}{!}{%
\begin{tabular}{lccccccc}
\toprule
Strategy & Mean ($\times 10^{-2}$) & Std ($\times 10^{-3}$) & Max ($\times 10^{-2}$) & P99 ($\times 10^{-2}$) & P95 ($\times 10^{-2}$) & P75 ($\times 10^{-2}$) & P50 ($\times 10^{-3}$) \\
\midrule
OGAS-L & $1.32 \pm 0.04$ & $3.81 \pm 0.18$ & $4.46 \pm 0.11$ & $2.60 \pm 0.09$ & $2.06 \pm 0.08$ & $1.50 \pm 0.05$ & $12.39 \pm 0.31$ \\
OGAS-L2 & $1.04 \pm 0.01$ & $\mathbf{3.42 \pm 0.12}$ & $4.19 \pm 0.14$ & $2.22 \pm 0.04$ & $1.71 \pm 0.03$ & $1.19 \pm 0.01$ & $9.59 \pm 0.17$ \\
OGAS-U & $1.03 \pm 0.02$ & $3.44 \pm 0.10$ & $\mathbf{4.09 \pm 0.18}$ & $\mathbf{2.22 \pm 0.05}$ & $\mathbf{1.70 \pm 0.04}$ & $1.17 \pm 0.03$ & $9.45 \pm 0.22$ \\
SBAL & $1.02 \pm 0.01$ & $5.74 \pm 0.18$ & $5.70 \pm 0.13$ & $2.93 \pm 0.06$ & $2.17 \pm 0.05$ & $1.28 \pm 0.03$ & $8.40 \pm 0.10$ \\
Top-K & $\mathbf{0.98 \pm 0.02}$ & $4.58 \pm 0.24$ & $5.03 \pm 0.12$ & $2.55 \pm 0.08$ & $1.89 \pm 0.06$ & $\mathbf{1.17 \pm 0.03}$ & $8.46 \pm 0.19$ \\
Sobol & $1.02 \pm 0.02$ & $5.88 \pm 0.10$ & $5.66 \pm 0.10$ & $2.98 \pm 0.05$ & $2.20 \pm 0.03$ & $1.29 \pm 0.02$ & $\mathbf{8.38 \pm 0.15}$ \\
Uniform & $1.04 \pm 0.02$ & $5.99 \pm 0.12$ & $5.84 \pm 0.11$ & $3.02 \pm 0.05$ & $2.24 \pm 0.05$ & $1.32 \pm 0.03$ & $8.52 \pm 0.20$ \\
\bottomrule
\end{tabular}}
\end{subtable}
\vspace{0.5em}
\begin{subtable}{\linewidth}
\centering
\caption{scOT}
\resizebox{\linewidth}{!}{%
\begin{tabular}{lccccccc}
\toprule
Strategy & Mean ($\times 10^{-2}$) & Std ($\times 10^{-3}$) & Max ($\times 10^{-2}$) & P99 ($\times 10^{-2}$) & P95 ($\times 10^{-2}$) & P75 ($\times 10^{-2}$) & P50 ($\times 10^{-2}$) \\
\midrule
OGAS-L & $2.34 \pm 0.22$ & $5.37 \pm 0.19$ & $6.33 \pm 0.18$ & $3.96 \pm 0.23$ & $3.35 \pm 0.25$ & $2.63 \pm 0.23$ & $2.23 \pm 0.22$ \\
OGAS-L2 & $\mathbf{1.72 \pm 0.19}$ & $\mathbf{4.42 \pm 0.23}$ & $\mathbf{5.45 \pm 0.34}$ & $\mathbf{3.08 \pm 0.25}$ & $\mathbf{2.55 \pm 0.24}$ & $\mathbf{1.96 \pm 0.21}$ & $1.63 \pm 0.19$ \\
OGAS-U & $1.86 \pm 0.04$ & $4.75 \pm 0.13$ & $5.75 \pm 0.14$ & $3.31 \pm 0.08$ & $2.76 \pm 0.06$ & $2.11 \pm 0.05$ & $1.76 \pm 0.04$ \\
SBAL & $1.82 \pm 0.18$ & $6.30 \pm 0.29$ & $8.32 \pm 0.34$ & $3.86 \pm 0.23$ & $3.01 \pm 0.22$ & $2.12 \pm 0.19$ & $1.66 \pm 0.17$ \\
Top-K & $1.75 \pm 0.16$ & $6.19 \pm 0.61$ & $7.97 \pm 0.76$ & $3.75 \pm 0.34$ & $2.92 \pm 0.27$ & $2.06 \pm 0.18$ & $\mathbf{1.61 \pm 0.14}$ \\
Sobol & $1.77 \pm 0.18$ & $6.35 \pm 0.16$ & $8.27 \pm 0.26$ & $3.85 \pm 0.21$ & $2.96 \pm 0.21$ & $2.07 \pm 0.20$ & $1.61 \pm 0.19$ \\
Uniform & $1.91 \pm 0.22$ & $6.54 \pm 0.47$ & $8.83 \pm 0.99$ & $4.03 \pm 0.33$ & $3.14 \pm 0.29$ & $2.23 \pm 0.24$ & $1.76 \pm 0.21$ \\
\bottomrule
\end{tabular}}
\end{subtable}
\end{table}

\begin{table}[h]
\centering
\footnotesize
\caption{\textbf{Single-Model Kuramoto-Sivashinsky Results.} Evaluation is performed one single model trained or on the first model of the ensemble when ensemble is required for uncertainty computation. Values show mean $\pm$ std across seeds. Best per architecture in \textbf{bold}.}
\label{tab:ks-single}
\begin{subtable}{\linewidth}
\centering
\caption{UNet}
\resizebox{\linewidth}{!}{%
\begin{tabular}{lccccccc}
\toprule
Strategy & Mean ($\times 10^{-3}$) & Std ($\times 10^{-3}$) & Max ($\times 10^{-2}$) & P99 ($\times 10^{-2}$) & P95 ($\times 10^{-3}$) & P75 ($\times 10^{-3}$) & P50 ($\times 10^{-3}$) \\
\midrule
OGAS-L & $6.98 \pm 0.63$ & $1.79 \pm 0.20$ & $6.43 \pm 1.06$ & $1.20 \pm 0.08$ & $9.42 \pm 0.94$ & $7.63 \pm 0.62$ & $6.77 \pm 0.53$ \\
OGAS-L2 & $5.11 \pm 0.41$ & $\mathbf{1.32 \pm 0.05}$ & $\mathbf{5.08 \pm 0.41}$ & $\mathbf{0.87 \pm 0.05}$ & $6.99 \pm 0.42$ & $5.64 \pm 0.39$ & $4.96 \pm 0.40$ \\
OGAS-U & $4.86 \pm 0.46$ & $1.37 \pm 0.13$ & $6.07 \pm 0.81$ & $0.94 \pm 0.12$ & $\mathbf{6.70 \pm 0.59}$ & $5.21 \pm 0.42$ & $4.67 \pm 0.39$ \\
SBAL & $4.44 \pm 0.17$ & $3.24 \pm 0.34$ & $10.56 \pm 0.15$ & $1.83 \pm 0.10$ & $8.57 \pm 0.60$ & $4.51 \pm 0.15$ & $3.62 \pm 0.13$ \\
Top-K & $4.41 \pm 0.06$ & $3.37 \pm 0.16$ & $10.15 \pm 0.66$ & $1.86 \pm 0.12$ & $8.87 \pm 0.03$ & $4.57 \pm 0.15$ & $\mathbf{3.51 \pm 0.07}$ \\
Sobol & $4.43 \pm 0.02$ & $3.01 \pm 0.18$ & $11.14 \pm 1.68$ & $1.59 \pm 0.03$ & $7.79 \pm 0.24$ & $4.55 \pm 0.08$ & $3.73 \pm 0.03$ \\
Uniform & $\mathbf{4.28 \pm 0.06}$ & $2.61 \pm 0.14$ & $9.53 \pm 0.07$ & $1.47 \pm 0.06$ & $7.51 \pm 0.32$ & $\mathbf{4.44 \pm 0.07}$ & $3.66 \pm 0.09$ \\
\bottomrule
\end{tabular}}
\end{subtable}
\vspace{0.5em}
\begin{subtable}{\linewidth}
\centering
\caption{FNO}
\resizebox{\linewidth}{!}{%
\begin{tabular}{lccccccc}
\toprule
Strategy & Mean ($\times 10^{-3}$) & Std ($\times 10^{-3}$) & Max ($\times 10^{-2}$) & P99 ($\times 10^{-2}$) & P95 ($\times 10^{-2}$) & P75 ($\times 10^{-2}$) & P50 ($\times 10^{-3}$) \\
\midrule
OGAS-L & $11.00 \pm 0.58$ & $\mathbf{1.75 \pm 0.11}$ & $3.42 \pm 0.19$ & $1.57 \pm 0.11$ & $1.33 \pm 0.06$ & $1.17 \pm 0.06$ & $10.92 \pm 0.59$ \\
OGAS-L2 & $9.49 \pm 0.28$ & $1.76 \pm 0.06$ & $\mathbf{3.26 \pm 0.18}$ & $\mathbf{1.45 \pm 0.03}$ & $\mathbf{1.22 \pm 0.04}$ & $1.02 \pm 0.03$ & $9.28 \pm 0.29$ \\
OGAS-U & $8.63 \pm 0.02$ & $2.73 \pm 0.16$ & $4.88 \pm 0.28$ & $2.04 \pm 0.09$ & $1.29 \pm 0.04$ & $0.93 \pm 0.01$ & $7.97 \pm 0.10$ \\
SBAL & $9.26 \pm 0.36$ & $3.23 \pm 0.42$ & $3.34 \pm 0.11$ & $1.80 \pm 0.12$ & $1.57 \pm 0.12$ & $1.11 \pm 0.08$ & $8.18 \pm 0.27$ \\
Top-K & $9.84 \pm 0.16$ & $2.66 \pm 0.36$ & $3.49 \pm 0.23$ & $1.72 \pm 0.05$ & $1.50 \pm 0.04$ & $1.10 \pm 0.08$ & $9.41 \pm 0.26$ \\
Sobol & $8.56 \pm 0.47$ & $4.11 \pm 0.12$ & $6.46 \pm 0.16$ & $2.86 \pm 0.12$ & $1.56 \pm 0.05$ & $0.89 \pm 0.05$ & $7.20 \pm 0.46$ \\
Uniform & $\mathbf{8.30 \pm 0.32}$ & $4.02 \pm 0.16$ & $5.95 \pm 0.12$ & $2.79 \pm 0.07$ & $1.53 \pm 0.06$ & $\mathbf{0.87 \pm 0.04}$ & $\mathbf{6.93 \pm 0.25}$ \\
\bottomrule
\end{tabular}}
\end{subtable}
\vspace{0.5em}
\begin{subtable}{\linewidth}
\centering
\caption{scOT}
\resizebox{\linewidth}{!}{%
\begin{tabular}{lccccccc}
\toprule
Strategy & Mean ($\times 10^{-2}$) & Std ($\times 10^{-3}$) & Max ($\times 10^{-2}$) & P99 ($\times 10^{-2}$) & P95 ($\times 10^{-2}$) & P75 ($\times 10^{-2}$) & P50 ($\times 10^{-2}$) \\
\midrule
OGAS-L & $1.55 \pm 0.38$ & $2.45 \pm 0.71$ & $4.26 \pm 1.85$ & $2.15 \pm 0.56$ & $1.93 \pm 0.51$ & $1.67 \pm 0.42$ & $1.54 \pm 0.37$ \\
OGAS-L2 & $1.22 \pm 0.12$ & $\mathbf{1.92 \pm 0.25}$ & $3.20 \pm 0.23$ & $\mathbf{1.71 \pm 0.15}$ & $\mathbf{1.51 \pm 0.16}$ & $1.31 \pm 0.14$ & $1.21 \pm 0.12$ \\
OGAS-U & $1.30 \pm 0.15$ & $2.16 \pm 0.35$ & $\mathbf{3.19 \pm 0.28}$ & $1.82 \pm 0.23$ & $1.63 \pm 0.22$ & $1.40 \pm 0.17$ & $1.28 \pm 0.15$ \\
SBAL & $1.23 \pm 0.11$ & $3.46 \pm 1.70$ & $5.78 \pm 1.98$ & $2.72 \pm 0.86$ & $1.85 \pm 0.46$ & $1.30 \pm 0.10$ & $1.14 \pm 0.04$ \\
Top-K & $1.20 \pm 0.17$ & $5.02 \pm 2.12$ & $7.47 \pm 2.19$ & $3.30 \pm 1.12$ & $2.09 \pm 0.61$ & $1.33 \pm 0.20$ & $1.06 \pm 0.07$ \\
Sobol & $\mathbf{1.11 \pm 0.11}$ & $3.05 \pm 0.52$ & $5.17 \pm 0.82$ & $2.42 \pm 0.34$ & $1.64 \pm 0.24$ & $\mathbf{1.18 \pm 0.10}$ & $\mathbf{1.03 \pm 0.09}$ \\
Uniform & $1.16 \pm 0.11$ & $2.82 \pm 0.95$ & $4.80 \pm 0.39$ & $2.32 \pm 0.58$ & $1.66 \pm 0.35$ & $1.23 \pm 0.10$ & $1.09 \pm 0.07$ \\
\bottomrule
\end{tabular}}
\end{subtable}
\end{table}

\begin{table}[h]
\centering
\footnotesize
\caption{\textbf{Single-Model Gray-Scott Results.} Evaluation is performed one single model trained or on the first model of the ensemble when ensemble is required for uncertainty computation. Values show mean $\pm$ std across seeds. Best per architecture in \textbf{bold}.}
\label{tab:gs-single}
\begin{subtable}{\linewidth}
\centering
\caption{UNet}
\resizebox{\linewidth}{!}{%
\begin{tabular}{lccccccc}
\toprule
Strategy & Mean ($\times 10^{-2}$) & Std ($\times 10^{-2}$) & Max ($\times 10^{-1}$) & P99 ($\times 10^{-2}$) & P95 ($\times 10^{-2}$) & P75 ($\times 10^{-2}$) & P50 ($\times 10^{-3}$) \\
\midrule
OGAS-L & $1.41 \pm 0.11$ & $\mathbf{0.50 \pm 0.04}$ & $\mathbf{0.73 \pm 0.16}$ & $\mathbf{2.78 \pm 0.23}$ & $\mathbf{2.38 \pm 0.20}$ & $1.64 \pm 0.13$ & $12.60 \pm 0.92$ \\
OGAS-L2 & $1.37 \pm 0.04$ & $0.83 \pm 0.07$ & $1.00 \pm 0.39$ & $4.35 \pm 0.50$ & $3.16 \pm 0.09$ & $1.65 \pm 0.10$ & $10.91 \pm 0.55$ \\
OGAS-U & $1.35 \pm 0.23$ & $1.02 \pm 0.71$ & $1.08 \pm 0.06$ & $4.89 \pm 3.55$ & $4.18 \pm 2.98$ & $\mathbf{1.48 \pm 0.03}$ & $10.16 \pm 0.18$ \\
SBAL & $1.36 \pm 0.25$ & $1.29 \pm 0.49$ & $1.77 \pm 0.19$ & $6.12 \pm 2.81$ & $4.50 \pm 1.86$ & $1.65 \pm 0.13$ & $8.61 \pm 0.85$ \\
Top-K & $\mathbf{1.22 \pm 0.13}$ & $1.06 \pm 0.28$ & $1.67 \pm 0.07$ & $4.81 \pm 1.56$ & $3.81 \pm 1.08$ & $1.50 \pm 0.19$ & $\mathbf{7.94 \pm 0.11}$ \\
Sobol & $1.42 \pm 0.03$ & $1.43 \pm 0.22$ & $2.02 \pm 0.39$ & $6.69 \pm 1.15$ & $5.21 \pm 0.91$ & $1.71 \pm 0.14$ & $8.49 \pm 1.13$ \\
Uniform & $1.27 \pm 0.23$ & $1.14 \pm 0.41$ & $1.78 \pm 0.26$ & $5.38 \pm 2.17$ & $4.05 \pm 1.49$ & $1.55 \pm 0.29$ & $8.41 \pm 0.67$ \\
\bottomrule
\end{tabular}}
\end{subtable}
\vspace{0.5em}
\begin{subtable}{\linewidth}
\centering
\caption{FNO}
\resizebox{\linewidth}{!}{%
\begin{tabular}{lccccccc}
\toprule
Strategy & Mean ($\times 10^{-2}$) & Std ($\times 10^{-3}$) & Max ($\times 10^{-2}$) & P99 ($\times 10^{-2}$) & P95 ($\times 10^{-2}$) & P75 ($\times 10^{-2}$) & P50 ($\times 10^{-2}$) \\
\midrule
OGAS-L & $1.62 \pm 0.22$ & $6.94 \pm 0.84$ & $8.83 \pm 0.56$ & $3.61 \pm 0.51$ & $3.03 \pm 0.35$ & $1.98 \pm 0.27$ & $1.42 \pm 0.18$ \\
OGAS-L2 & $1.41 \pm 0.01$ & $\mathbf{6.21 \pm 0.17}$ & $\mathbf{7.36 \pm 0.43}$ & $\mathbf{3.21 \pm 0.03}$ & $\mathbf{2.71 \pm 0.03}$ & $1.69 \pm 0.01$ & $1.25 \pm 0.02$ \\
OGAS-U & $1.42 \pm 0.02$ & $7.06 \pm 0.35$ & $8.11 \pm 0.26$ & $3.58 \pm 0.14$ & $2.99 \pm 0.12$ & $1.67 \pm 0.02$ & $1.23 \pm 0.01$ \\
SBAL & $1.31 \pm 0.01$ & $8.40 \pm 0.02$ & $9.74 \pm 1.76$ & $3.89 \pm 0.01$ & $3.21 \pm 0.01$ & $1.56 \pm 0.02$ & $1.01 \pm 0.01$ \\
Top-K & $1.35 \pm 0.03$ & $8.70 \pm 0.15$ & $10.08 \pm 0.75$ & $4.01 \pm 0.06$ & $3.32 \pm 0.05$ & $1.61 \pm 0.04$ & $1.04 \pm 0.05$ \\
Sobol & $1.35 \pm 0.08$ & $8.63 \pm 0.30$ & $9.24 \pm 0.82$ & $3.99 \pm 0.15$ & $3.29 \pm 0.13$ & $1.63 \pm 0.11$ & $1.06 \pm 0.08$ \\
Uniform & $\mathbf{1.29 \pm 0.03}$ & $8.49 \pm 0.33$ & $10.22 \pm 0.70$ & $3.90 \pm 0.13$ & $3.23 \pm 0.11$ & $\mathbf{1.55 \pm 0.03}$ & $\mathbf{1.00 \pm 0.01}$ \\
\bottomrule
\end{tabular}}
\end{subtable}
\vspace{0.5em}
\begin{subtable}{\linewidth}
\centering
\caption{scOT}
\resizebox{\linewidth}{!}{%
\begin{tabular}{lccccccc}
\toprule
Strategy & Mean ($\times 10^{-2}$) & Std ($\times 10^{-3}$) & Max ($\times 10^{-2}$) & P99 ($\times 10^{-2}$) & P95 ($\times 10^{-2}$) & P75 ($\times 10^{-2}$) & P50 ($\times 10^{-2}$) \\
\midrule
OGAS-L & $1.95 \pm 0.14$ & $7.80 \pm 0.10$ & $9.30 \pm 1.18$ & $4.40 \pm 0.08$ & $3.47 \pm 0.14$ & $2.34 \pm 0.14$ & $1.69 \pm 0.16$ \\
OGAS-L2 & $\mathbf{1.44 \pm 0.07}$ & $\mathbf{6.13 \pm 0.40}$ & $\mathbf{6.41 \pm 1.57}$ & $\mathbf{3.25 \pm 0.17}$ & $\mathbf{2.77 \pm 0.13}$ & $\mathbf{1.66 \pm 0.10}$ & $1.26 \pm 0.06$ \\
OGAS-U & $1.53 \pm 0.05$ & $7.32 \pm 0.47$ & $7.47 \pm 1.04$ & $3.80 \pm 0.31$ & $3.10 \pm 0.10$ & $1.80 \pm 0.08$ & $1.30 \pm 0.03$ \\
SBAL & $1.46 \pm 0.03$ & $9.07 \pm 0.05$ & $10.13 \pm 1.59$ & $4.29 \pm 0.11$ & $3.31 \pm 0.03$ & $1.82 \pm 0.06$ & $\mathbf{1.13 \pm 0.03}$ \\
Top-K & $1.47 \pm 0.09$ & $9.02 \pm 1.09$ & $10.71 \pm 2.03$ & $4.28 \pm 0.69$ & $3.30 \pm 0.18$ & $1.83 \pm 0.12$ & $1.15 \pm 0.06$ \\
Sobol & $1.46 \pm 0.04$ & $8.89 \pm 0.78$ & $9.63 \pm 1.44$ & $4.12 \pm 0.41$ & $3.29 \pm 0.12$ & $1.83 \pm 0.08$ & $1.14 \pm 0.02$ \\
Uniform & $1.48 \pm 0.14$ & $8.80 \pm 0.66$ & $9.98 \pm 2.53$ & $4.18 \pm 0.54$ & $3.28 \pm 0.18$ & $1.84 \pm 0.14$ & $1.16 \pm 0.12$ \\
\bottomrule
\end{tabular}}
\end{subtable}
\end{table}

\subsection{Trajectory Samples}
\label{app:trajectory_samples}

\begin{figure}[h]
    \centering
    \begin{subfigure}[b]{1.0\linewidth}
        \includegraphics[width=\linewidth]{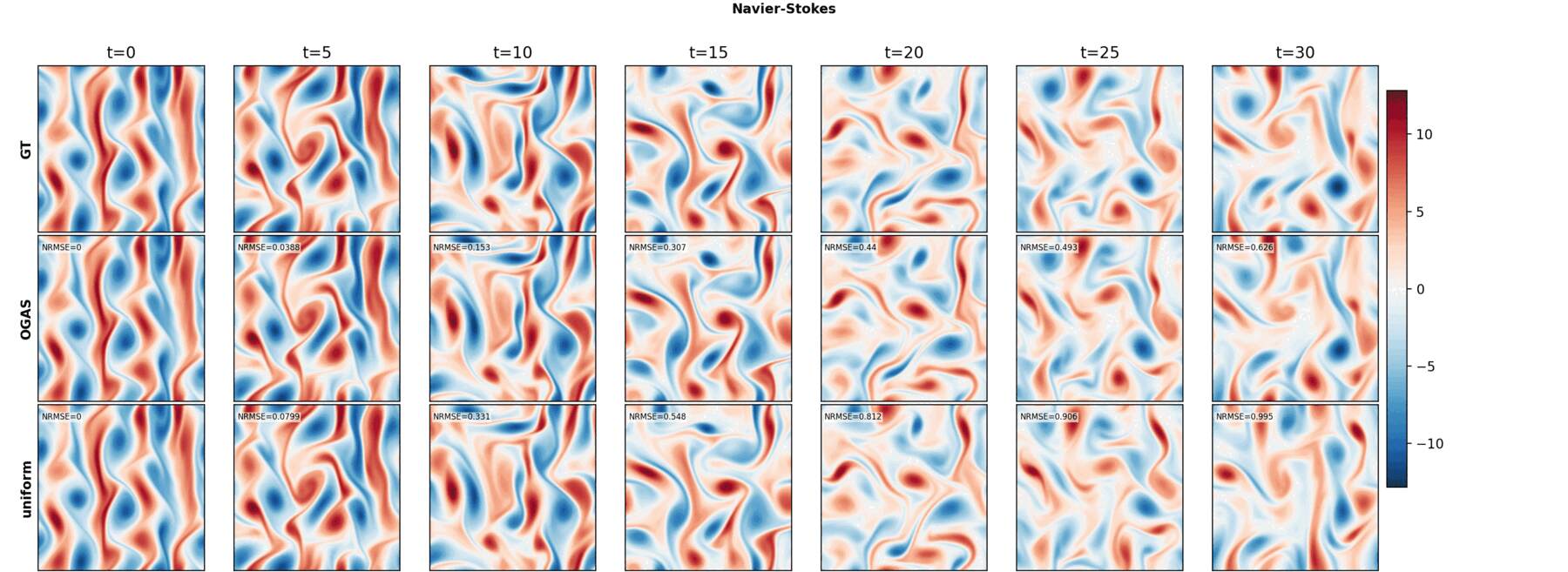}
        \caption{U-Net}
    \end{subfigure}
    \vspace{0.5em}
    \begin{subfigure}[b]{1.0\linewidth}
        \includegraphics[width=\linewidth]{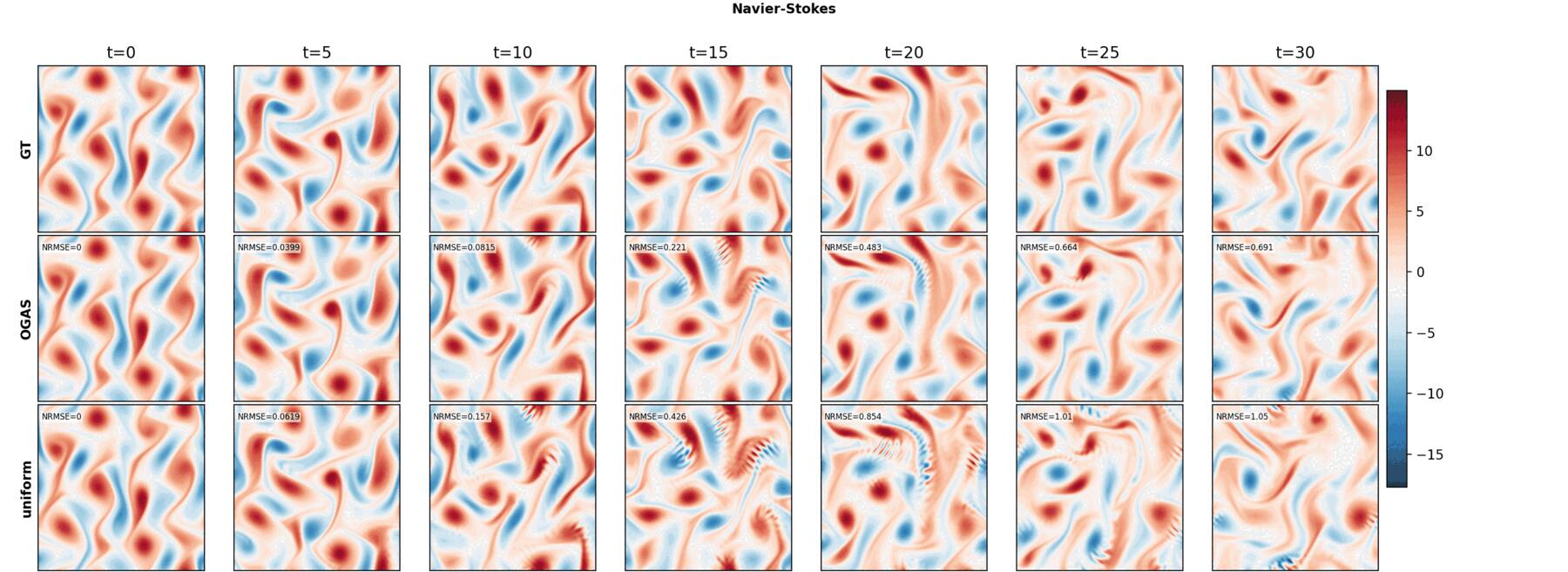}
        \caption{FNO}
    \end{subfigure}
    \vspace{0.5em}
    \begin{subfigure}[b]{1.0\linewidth}
        \includegraphics[width=\linewidth]{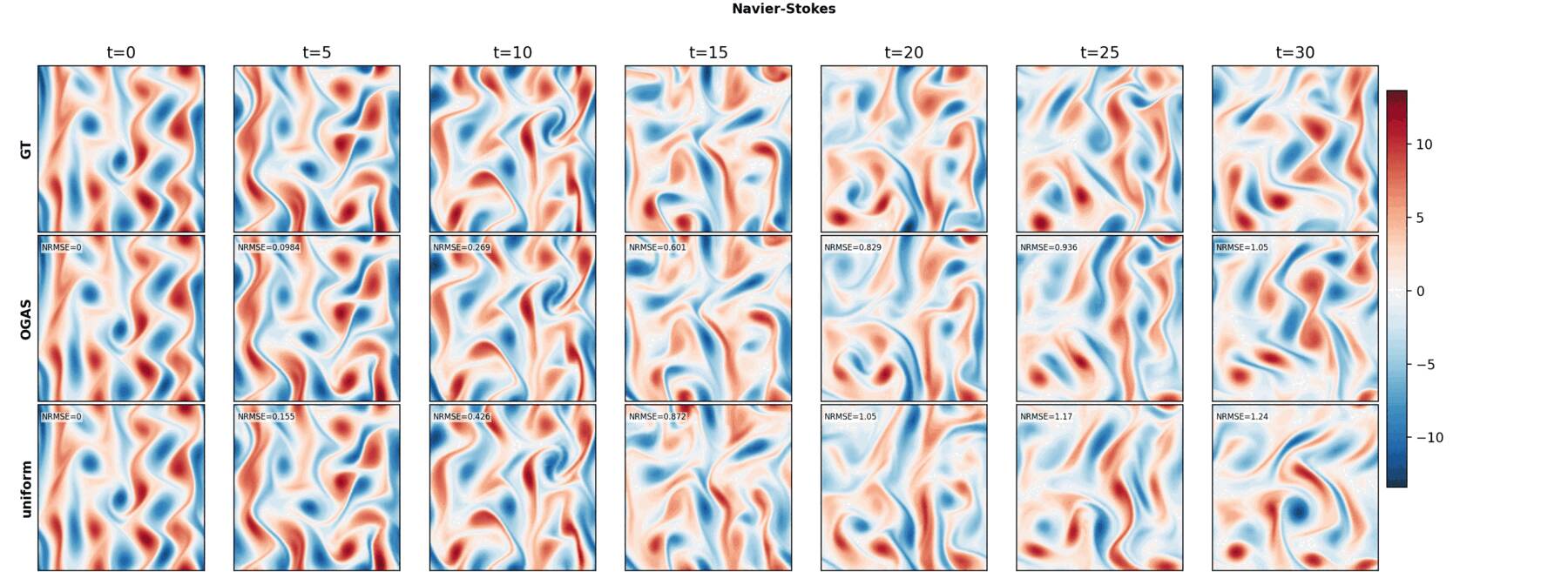}
        \caption{scOT}
    \end{subfigure}
    \Description{Three panels compare Navier--Stokes rollout predictions from U-Net, FNO, and scOT against ground truth and uniform- versus OGAS-trained surrogates on difficult trajectories.}
    \caption{\textbf{Navier-Stokes Trajectory Samples.} Comparison of rollout predictions on difficult trajectories. We compare the ground truth (GT) against surrogates trained with Uniform sampling and OGAS}
    \label{fig:ns_samples}
\end{figure}

\begin{figure}[h]
    \centering
    \begin{subfigure}[b]{1.0\linewidth}
        \includegraphics[width=\linewidth]{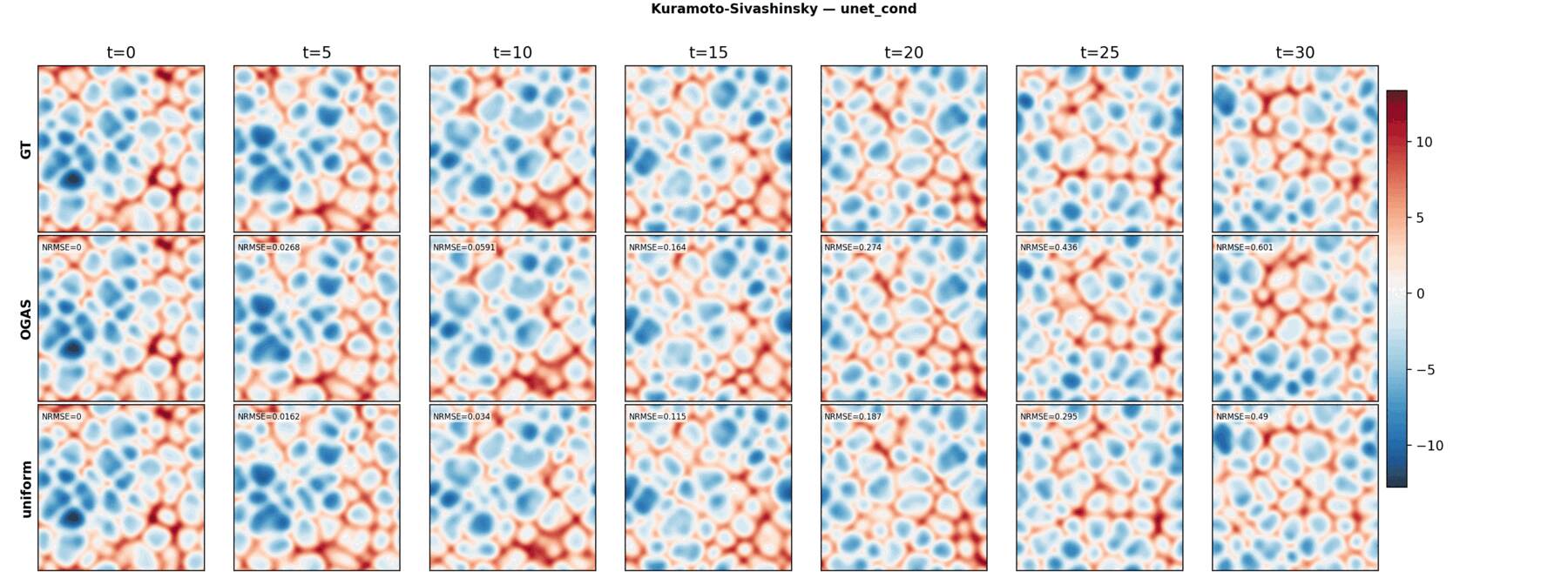}
        \caption{U-Net}
    \end{subfigure}
    \vspace{0.5em}
    \begin{subfigure}[b]{1.0\linewidth}
        \includegraphics[width=\linewidth]{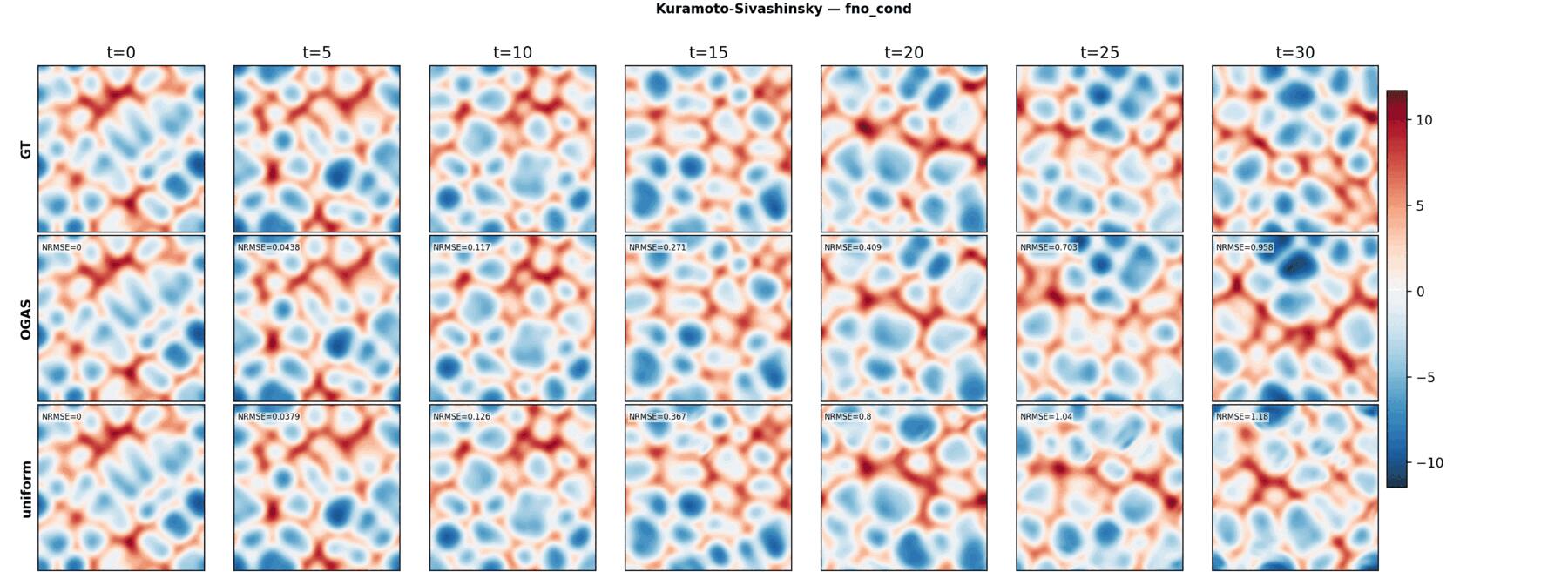}
        \caption{FNO}
    \end{subfigure}
    \vspace{0.5em}
    \begin{subfigure}[b]{1.0\linewidth}
        \includegraphics[width=\linewidth]{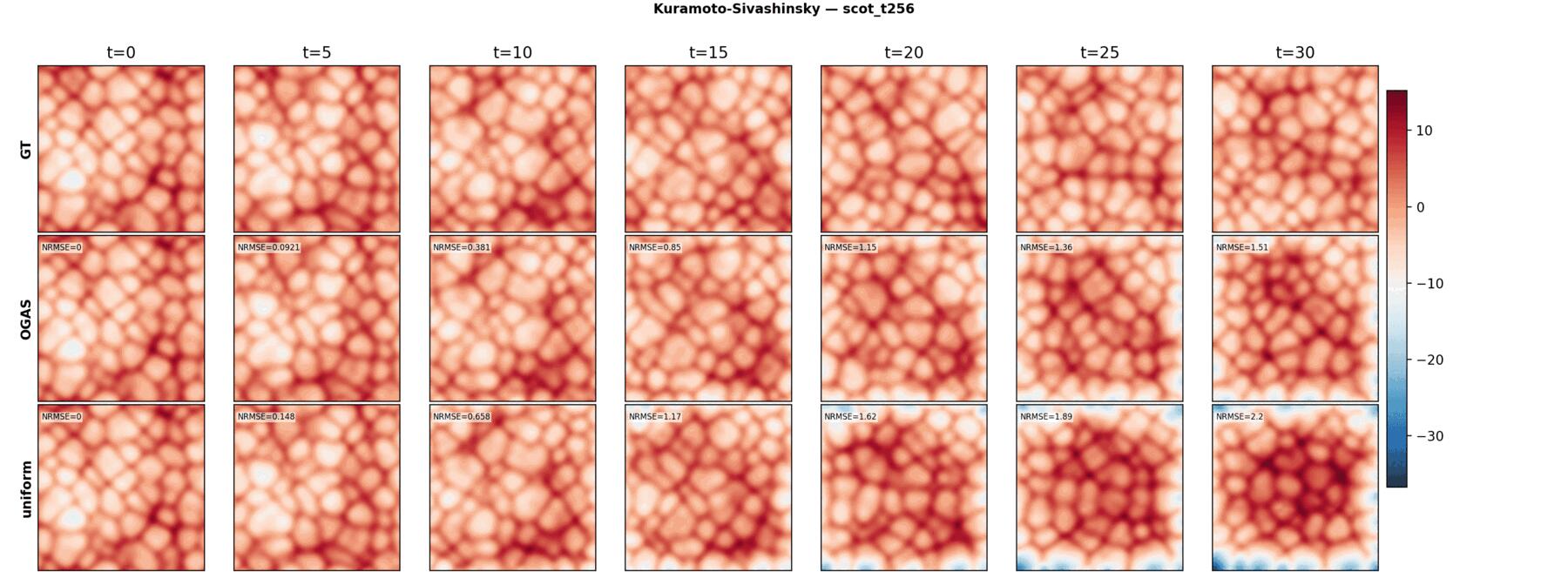}
        \caption{scOT}
    \end{subfigure}
    \Description{Three panels compare Kuramoto--Sivashinsky rollout predictions from U-Net, FNO, and scOT against ground truth and uniform- versus OGAS-trained surrogates on difficult trajectories.}
    \caption{\textbf{Kuramoto-Sivashinsky Trajectory Samples.} Comparison of rollout predictions on difficult trajectories. We compare the ground truth (GT) against surrogates trained with Uniform sampling and OGAS.}
    \label{fig:ks_samples}
\end{figure}

\begin{figure}[h]
    \centering
    \begin{subfigure}[b]{1.0\linewidth}
        \includegraphics[width=\linewidth]{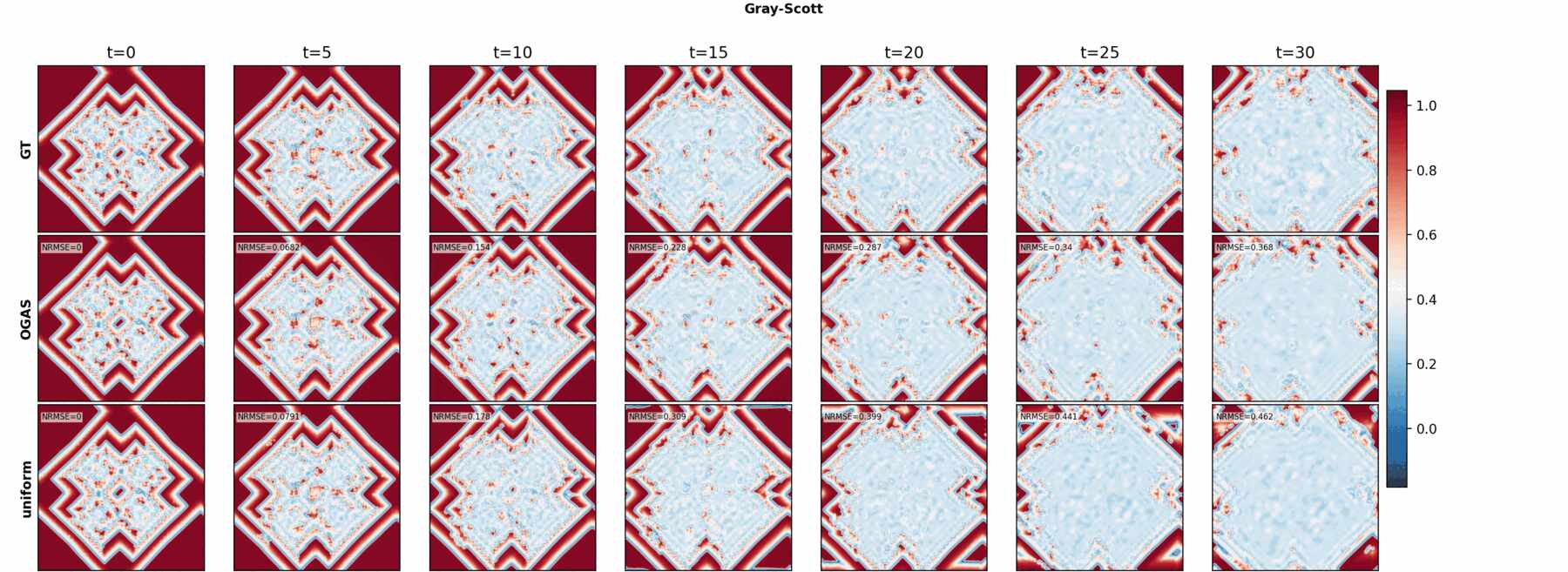}
        \caption{U-Net}
    \end{subfigure}
    \vspace{0.5em}
    \begin{subfigure}[b]{1.0\linewidth}
        \includegraphics[width=\linewidth]{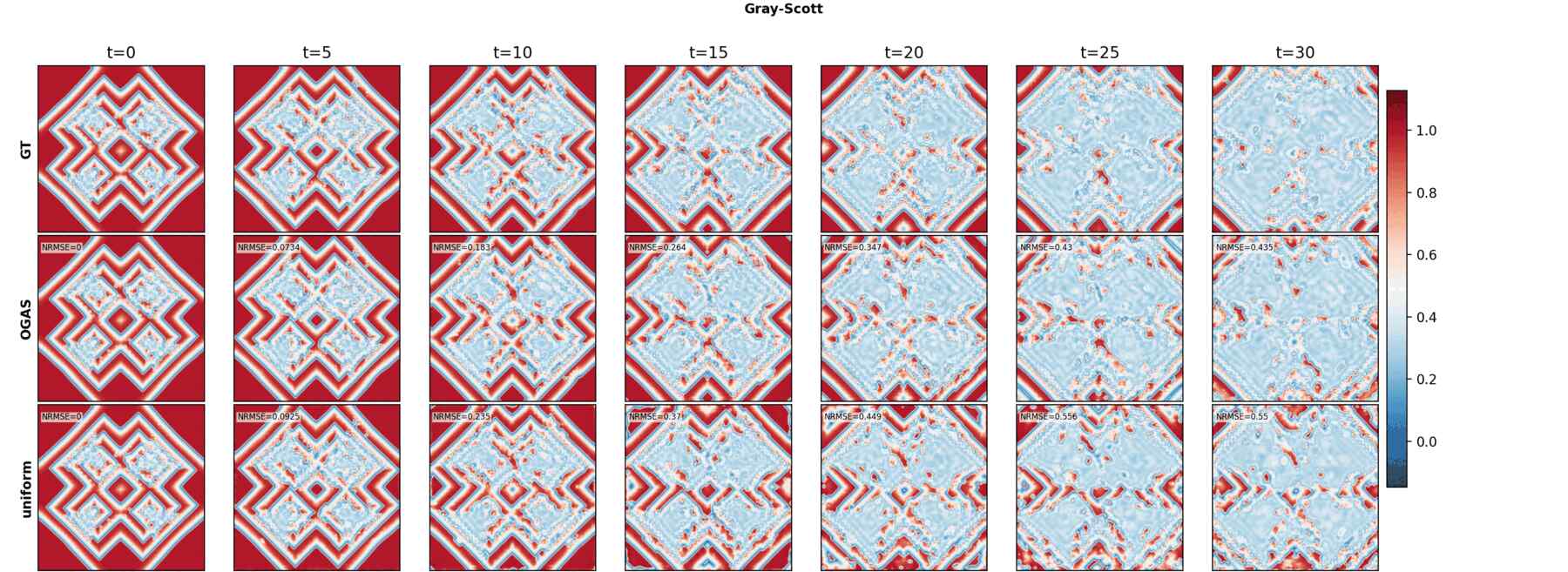}
        \caption{FNO}
    \end{subfigure}
    \vspace{0.5em}
    \begin{subfigure}[b]{1.0\linewidth}
        \includegraphics[width=\linewidth]{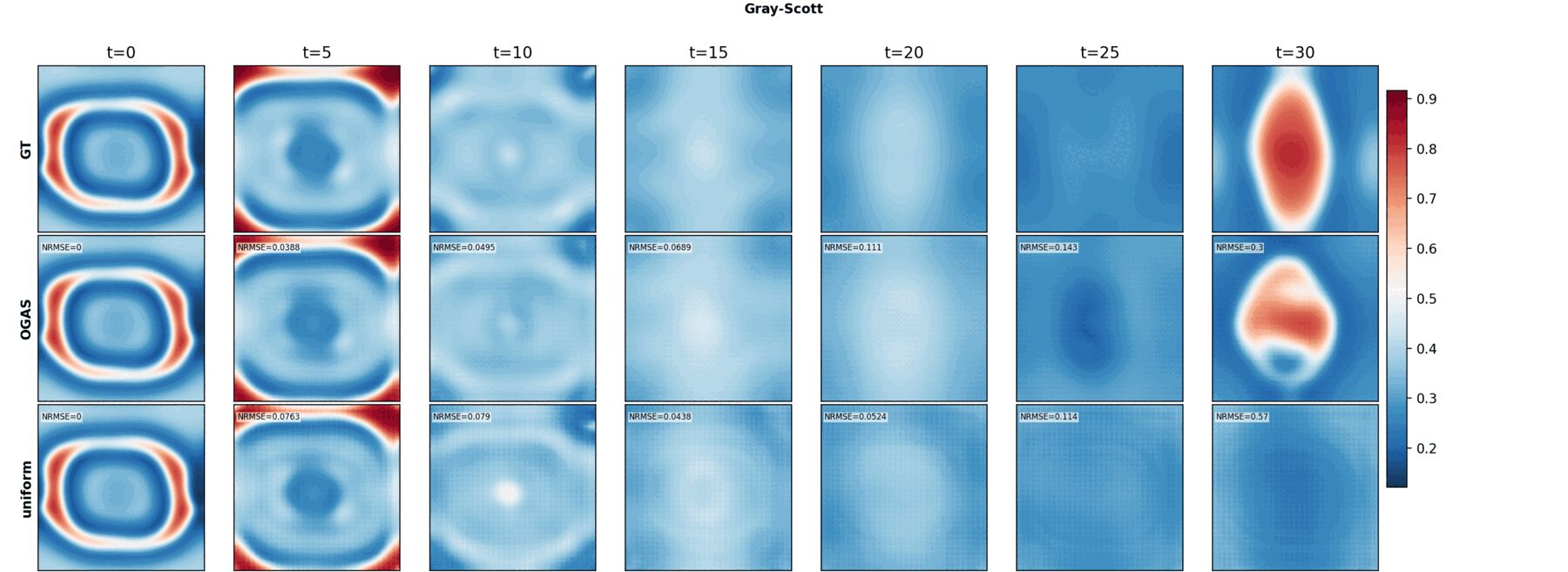}
        \caption{scOT}
    \end{subfigure}
    \Description{Three panels compare Gray--Scott rollout predictions from U-Net, FNO, and scOT against ground truth and uniform- versus OGAS-trained surrogates on difficult trajectories.}
    \caption{\textbf{Gray-Scott Trajectory Samples.} Comparison of rollout predictions on difficult trajectories. We compare the ground truth (GT) against surrogates trained with Uniform sampling and OGAS.}
    \label{fig:gs_samples}
\end{figure}

We visualize example trajectories, comparing the predictions of surrogates trained with OGAS-L versus Uniform sampling against the ground truth. See Figure~\ref{fig:ns_samples}, Figure~\ref{fig:ks_samples}, and Figure~\ref{fig:gs_samples}.

\end{document}